\documentclass[sigconf]{acmart}
\usepackage{balance} 

\usepackage[utf8]{inputenc}
\usepackage{color,soul}
\usepackage{geometry}
\usepackage{multicol}
\usepackage{subfig} 
\usepackage{enumitem}

\usepackage[ruled]{algorithm2e}

\SetAlFnt{\small}
\SetAlCapFnt{\small}
\SetAlCapNameFnt{\small}

\setcopyright{none} 
\renewcommand\footnotetextcopyrightpermission[1]{} 

\acmConference[IPSN'20]{The 19th ACM/IEEE Conference on
Information Processing in Sensor Networks}{April 21--24, 2020}
{Sydney, Australia}

\begin{document}

\title{
MDLdroid: a ChainSGD-reduce Approach to Mobile Deep Learning for Personal Mobile Sensing
}

\author{Yu Zhang}
\affiliation{
  \institution{RMIT University}
}
\email{zac.lhjzyzzoo@gmail.com}

\author{Tao Gu}
\affiliation{
  \institution{RMIT University}
}
\email{tao.gu@rmit.edu.au }

\author{Xi Zhang}
\affiliation{
  \institution{RMIT University}
}
\email{zaibuer@gmail.com}

\begin{abstract}

Personal mobile sensing is fast permeating our daily lives to enable activity monitoring, healthcare and rehabilitation. Combined with deep learning, these applications have achieved significant success in recent years.
Different from conventional cloud-based paradigms, running deep learning on devices offers several advantages including data privacy preservation and low-latency response for both model inference and update. 
Since data collection is costly in reality, Google's Federated Learning offers not only complete data privacy but also better model robustness based on multiple user data. 
However, personal mobile sensing applications are mostly user-specific and highly affected by environment. As a result, continuous local changes may seriously affect the performance of a global model generated by Federated Learning. In addition, deploying Federated Learning on a local server, e.g., edge server, may quickly reach the bottleneck due to resource constraint and serious failure by attacks.
Towards pushing deep learning on devices, we present MDLdroid, a novel decentralized mobile deep learning framework to enable resource-aware on-device collaborative learning for personal mobile sensing applications.
To address resource limitation, we propose a ChainSGD-reduce approach which includes a novel \textit{chain-directed} Synchronous Stochastic Gradient Descent algorithm to effectively reduce overhead among multiple devices. We also design an agent-based \textit{multi-goal} reinforcement learning mechanism to balance resources in a fair and efficient manner. Our evaluations show that our model training 
on off-the-shelf mobile devices achieves 2x to 3.5x faster than single-device training, and 1.5x faster than the master-slave approach.

\end{abstract}



\maketitle

\renewcommand{\shortauthors}{Y.~Zhang et al.}

\section{Introduction}

With the rapid development of mobile and wearable devices, recent years have witnessed an explosion of mobile sensing applications. These applications gain an insight into people's life based on rich personal sensing data, e.g., understanding biological contexts in daily living \mbox{\cite{Murnane:2016:MMA:2935334.2935383}}, recognizing activities in ambient assisted living areas \mbox{\cite{Jeyakumar:19}}, and monitoring personal health in smart home or hospital \mbox{\cite{Miotto:2017:healthcare}}.
Machine learning has been commonly used to process sensing data. However, traditional machine learning techniques require manual and complex feature engineering.
Deep Learning (DL) has gained an increasing popularity due to its higher model accuracy. Besides, its automated feature extraction capability and the ability of scaling with data make it an ideal solution for processing multi-modality sensing data \mbox{\cite{Wang:2018}}. It is advocated that DL will be the next key enabler for advanced personal mobile sensing \mbox{\cite{Lane:2015:DLR:2699343.2699349}}.

\textbf{Personal Sensing Requirements}
A successful DL application requires a huge amount of data to train a model, in which large computation resources will be involved. The commercial solution is to transmit the data from local to cloud, offloading the heavy training workloads to the cloud.
Edge server can also be used to efficiently offload the workloads \mbox{\cite{Li:2018:EIO:3229556.3229562}}. 
However, real-world personal sensing applications pose several requirements in which the server-based approaches may not fit.
Firstly, personal sensing data are significantly privacy-sensitive as the data contain a variety of human motion and biological contexts. Studies \mbox{\cite{Wang:2015:MML:2789168.2790121}} \mbox{\cite{Kroger:2019:PIA:3309074.3309076}} show that leaking motion data can cause a series of violations concerning user location, health condition, emotional state, and identification information. Even governments across the worlds are reinforcing laws to protect personal data privacy, for example the General Data Protection Regulation (GDPR) \mbox{\cite{Yang:19}}. 
Server-based approaches have serious \textit{privacy} concerns. 
Secondly, due to dynamic external nature effects and internal characteristics of each individual, personal sensing data are strongly affected by specific personal activities, social abilities and surrounding environment conditions over time \mbox{\cite{Laport:2019}} \mbox{\cite{Jeyakumar:19}}, and features may be updated at any time. Simply deploying a pre-trained global model on device may not \textit{continually} work well to adapt the \textit{local} features which have been changed. 
Continually training a local model with new data is a fundamental requirement, and applications usually require low-latency responses for both model inference and update.  
In practice, continually transmitting sensing data to the server and downloading model updates for \textit{training} can incur fast battery drain and considerable latency for mobile devices especially when the network connection is unstable or broken.
Different from server-based approaches, mobile DL may be an ideal solution to effectively preserve sensing data privacy without being transmitted over the public network, and enable quick local model inference and update response for continuous on-device training \mbox{\cite{Yunbin:2019}} \mbox{\cite{Wang:2018}}.
Thus, mobile DL is a promising direction for personal sensing applications.

\textbf{Mobile Deep Learning Challenges} 
Existing work \mbox{\cite{Wang:2018}} reveal that mobile DL poses two challenges. 
Firstly, pushing DL with both model training and inference to a resource-constraint mobile device in practice is extremely constricted by its capability.
Secondly, data collection with various conditions is usually costly in reality, and the model performance and robustness can be seriously affected due to insufficient data. 
Collaborative deep learning is proposed to ensure model training efficiency and robustness based on multiple user data \mbox{\cite{Zhang:18:collaborative}}.
Following this direction, Google's Federated Learning \mbox{\cite{Jakub:2016}} has been proposed to provide a mobile collaborative DL framework.
This framework not only preserves data privacy by transferring the gradient parameters of a DL model without exposing local data, but also improves the global model performance and robustness by a large number of user models. 
However, Federated Learning relies on a central server for intensive global model aggregations, which may not work well for personal sensing scenarios due to the low-latency requirement for continuous local model update. 
Scaling down the framework to a central edge server \mbox{\cite{Wang:19}} or even a master device may efficiently reduce the latency. However, Federated Learning deploys a master-slave structure which is less attack-resistant \mbox{\cite{Zyskind:15}}. A recent study \mbox{\cite{pmlr-v97-bhagoji19a}} demonstrates that the model poisoning attacks can cause a serious negative impact to the central training process. 
In addition, the central edge server or master device may quickly become the bottleneck due to substantial communication and intensive model aggregation, resulting in huge resource overhead \mbox{\cite{Lian:2017:DAO:3295222.3295285}}. 
Moreover, the system fault-tolerant may be limited in practice if the central node goes down.

In contrast, applying a decentralized structure for mobile DL can theoretically offer high attack-resistant and reliable fault-tolerant \mbox{\cite{Zyskind:15}}.
Existing approaches propose a theoretical decentralized structure based on a \textit{fixed directed} graph network such as Ring-Allreduce \mbox{\cite{Li:2014:SDM:2685048.2685095}}. These approaches have been applied in high performance cloud environments where resources are rich and stable. However, when applied to mobile devices where their resources are scarce and dynamic, the training process can easily suspend or crash due to low level of resources available. 
Furthermore, studies \mbox{\cite{Hoseini:13}} \mbox{\cite{Jeyakumar:19}} report that using opportunistic user context from \textit{multiple} people can enrich the sensing data distribution to make model robust to the \textit{local} variations.
Therefore, moving towards a \textit{local resource-aware decentralized collaborative} mobile DL approach without central server support for personal sensing applications is strongly motivated in order to mitigate resource overhead and reduce latency for training.

\textbf{Decentralized Mobile Scheduling}
Deploying a decentralized DL framework on multiple devices can be extremely difficult due to the design of resource-efficiency task scheduling for resource-constrained devices. A generic optimization algorithm \mbox{\cite{Yang:2009}} can be applied to perform task scheduling. 
However, the recent \textit{Multi-agents Reinforcement Learning} (MARL) approach \mbox{\cite{Thanh:2018}} may work better in such a \textit{non-stationarity} and multi-device scenario. 
Agent-based approaches have a significant advantage of estimating the future scheduling order because agents inherently learn features from the past experiences based on Reinforcement Learning (RL), while the generic optimization approaches cannot estimate a future possible state and will be limited to control a complex environment due to lacking of such learning mechanism \mbox{\cite{ben:2017:RL}}.
Thus, using agent-based approaches can theoretically make scheduling more reliable for complex environments.

\textbf{Our Approach}
Aiming to push DL to mobile devices, we present MDLdroid, a novel decentralized \textbf{M}obile \textbf{D}eep \textbf{L}earning framework to enable resource-aware on-device collaborative learning for personal mobile sensing applications. MDLdroid targets to fully operate on multiple off-the-shelf An\textbf{droid} smartphones connected in a mesh network, and achieve high training accuracy and reliable execution of the state-of-the-art DL models.

Our challenge is two-fold. To address the first challenge of resource constraint on device, we propose a ChainSGD-reduce approach which essentially uses a novel \textit{chain-directed} Synchronous Stochastic Gradient Descent (S-SGD) algorithm to effectively reduce resource overhead among devices for training. The key idea is to decentralize the S-SGD algorithm \cite{Dean:2012:LSD:2999134.2999271} running on a single device to multiple devices with dynamic \textit{chain-directed} model aggregation. Specifically, each device runs a descendant model for training in which model aggregation task is managed by any two devices at a time to achieve \textit{minimal-peak} (i.e., minimum communication and memory) of resource overhead for each device. 
Different from the existing Ring-Allreduce approach \cite{Li:2014:SDM:2685048.2685095}, each pair of devices is dynamically scheduled to perform model aggregation based on their resource condition for latency reduction. 
In practice, we leverage on a mesh graph to aggregate multiple descendant model parameters into global model parameters to complete each training iteration. Once the \textit{Epoch} reaches the given iterations, the entire training process is completed. The global model generated can then be deployed on each device for inference.

Secondly, designing a resource-efficiency task scheduler in a decentralized DL framework is challenging because resources on device are constrained strictly and changed dynamically (i.e., the \textit{Non-stationarity} problem). A device may be dynamically switched to \textit{busy} state due to other high priority tasks, leading to training being suspended. In addition, the model aggregation task needs to be distributed to multiple devices in a fair manner to battery consumption across the network. To address this challenge, we propose a single agent-based scheduler based on Reinforcement Learning to self organize scheduling tasks using dynamic information of on-device resource, named \textit{Chain-scheduler}.
Chain-scheduler includes an effective reward function design to map each perceived scheduling action to a reward, aiming to optimize the given constraints.
Through our analysis in Section (\S\ref{section:chain-balance}), Chain-scheduler can achieve the best latency-reduced scheduling close to \textit{Tree-scheduler} in the Tree-Allreduce approach \cite{Shaohuai:2019}, and it can also achieve the best energy balance close to \textit{Ring-scheduler} in the Ring-Allreduce approach \cite{Li:2014:SDM:2685048.2685095}. 
To tackle the \textit{Non-stationarity} problem, we apply a continuous environment learning strategy to repeat learning if the scheduling environment is changed. In this way, scheduling can be more adaptive to on-device resource changes.  
In addition, we further enhance the performance of the reward function by designing a Threshold-based Decaying Greedy-Exploration (TDGE) strategy to save on-device resources used for scheduling tasks.

We fully implement MDLdroid on off-the-shelf smartphones based on modified DL4J libraries. To evaluate MDLdroid, we use two standard Convolutional Neural Network (CNN) models (LeNet \mbox{\cite{Lecun:1998}} and AlexNet \mbox{\cite{Krizhevsky:2012:ICD:2999134.2999257})} since CNN can be used to effectively process multi-channel sensing data \mbox{\cite{Yang2015DeepCN}}.
We evaluate the training performance using 6 public datasets in Table \ref{table:dataset}, containing diverse personal mobile sensing data. Results show that MDLdroid achieves high training accuracy which is comparable to the state-of-the-art accuracy reported in \ref{tab:data_accuracy}, speeds up training by 2x to 3.5x compared to the single-device baseline and 1.5x compared to Federated Learning. 
In addition, MDLdroid reduces latency overhead due to busy condition by 23\% and 53\% compared to \textit{Tree-scheduler} and \textit{Ring-scheduler}, respectively. Moreover, MDLdroid reduces the variance in battery consumption among devices by 40\% compared to \textit{Tree-scheduler}.



\textbf{Summary of Contributions}
Key contributions of this paper are summarized as follows:

\begin{itemize}[leftmargin=*,noitemsep,topsep=0pt]
\item We present MDLdroid, a novel decentralized mobile DL framework based on a mesh network to enable resource-aware on-device collaborative learning for personal mobile sensing applications (\S\ref{section:system}). 

\item We propose a ChainSGD-reduce approach, in particular a \textit{chain-directed} S-SGD algorithm, to minimize the resource overhead of model aggregation tasks in a decentralized framework (\S\ref{section:chain-reduce}).

\item We design an agent-based \textit{multi-goal} reinforcement learning mechanism, Chain-scheduler, with an \textit{accelerated} reward function to manage and balance resources in a fair and efficient manner (\S\ref{section:chain-balance}).

\item We evaluate MDLdroid on off-the-shelf Android smartphones with two standard DL models using 6 public personal mobile sensing datasets (\S\ref{section:evaluation}). Results indicate that MDLdroid accelerates training effectively, outperforming the state-of-the-arts. 
\end{itemize}

To the best of our knowledge, this is the first time that collaborative DL is fully implemented and functioned on off-the-shelf mobile devices with both model training and inference without central server support. MDLdroid takes mobile DL to the next step, and opens up new possibilities for building secure and adaptive mobile sensing applications.

\textbf{Implication}
MDLdroid defines two implication models. 
The \textit{individual} model is used for an individual who has sufficient personal data and multiple mobile devices to offer shared resources. The personal sensing data can only be safely distributed to the given mobile devices verified by the same identity (e.g., Google account). MDLdroid can accelerate on-device learning, and alleviate the resource burden from a single device.

The \textit{non-individual} model is mainly applied for a group of people to explore specific local sensing features. The data will be strictly kept on device to preserve data privacy, and only the model gradient parameters of each individual will be exchanged to improve model robustness. With the nature benefits of a decentralized approach, model poisoning check \mbox{\cite{pmlr-v97-bhagoji19a}} can be easily deployed in each device to prevent attacks. Even if a few of devices are down due to attacks, MDLdroid can isolate them and keep others working safely to avoid catastrophic failure. Moreover, MDLdroid can be potentially used in many multi-user sensing scenarios \mbox{\cite{Abkenar:19}}, such as specific family behaviour recognition in a smart home, patient-specific health condition monitoring in a hospital, and healthcare tracking for older adults in different aged care facilities.

\section{Motivation}

\begin{figure}
 \centering
     \subfloat[]{
    \includegraphics[trim=0 0 25 5, clip,width=0.5\columnwidth]{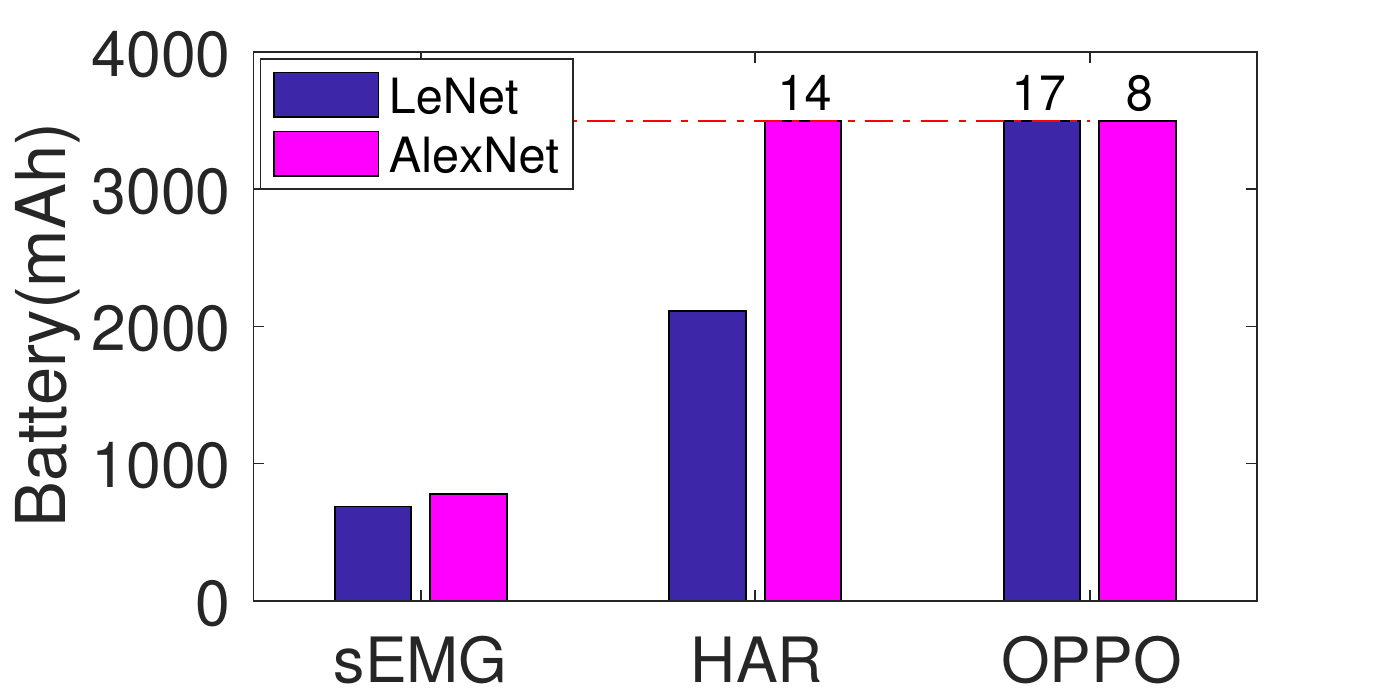}
    \label{fig:intro_sensing_a}
    }
    \subfloat[]{
    \includegraphics[trim=0 0 20 5, clip,width=0.5\columnwidth]{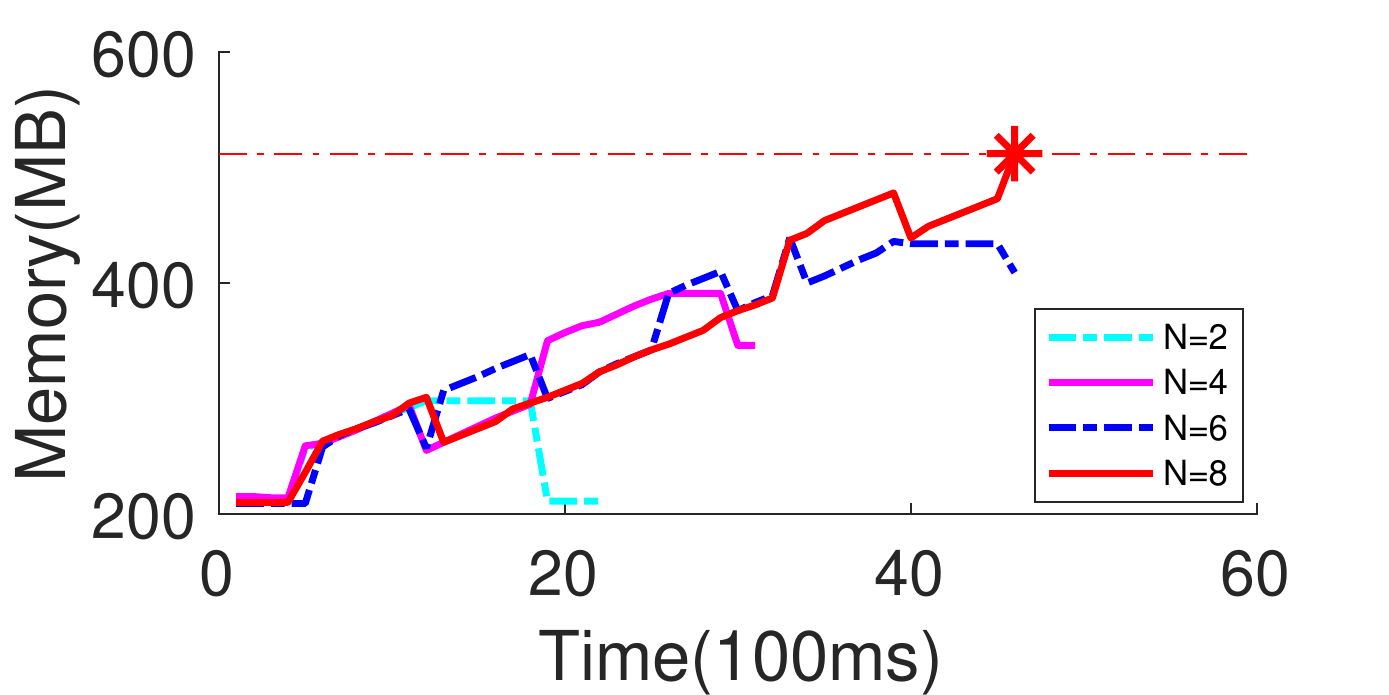}
    \label{fig:intro_memory_a}
    } \quad
    \subfloat[]{
    \includegraphics[trim=0 0 25 5, clip,width=0.5\columnwidth]{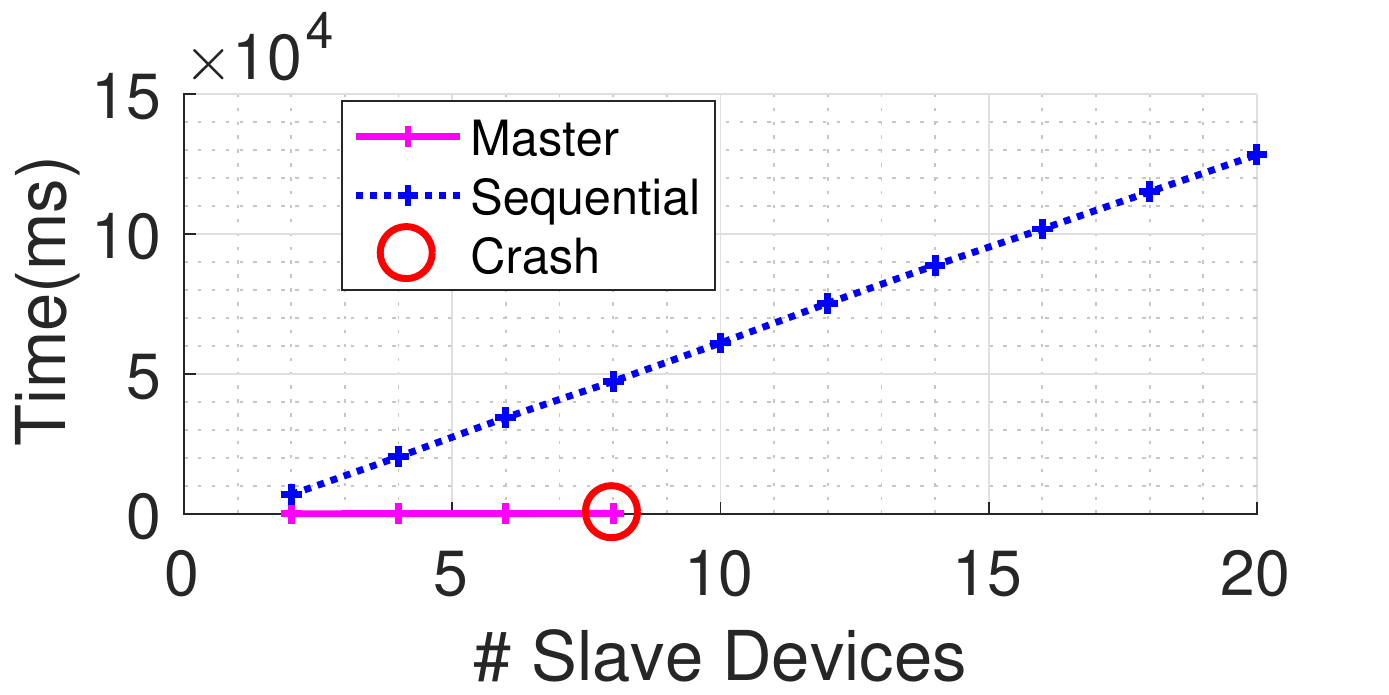}
    \label{fig:intro_sensing_b}
    }
    \subfloat[]{
    \includegraphics[trim=0 0 25 5, clip,width=0.5\columnwidth]{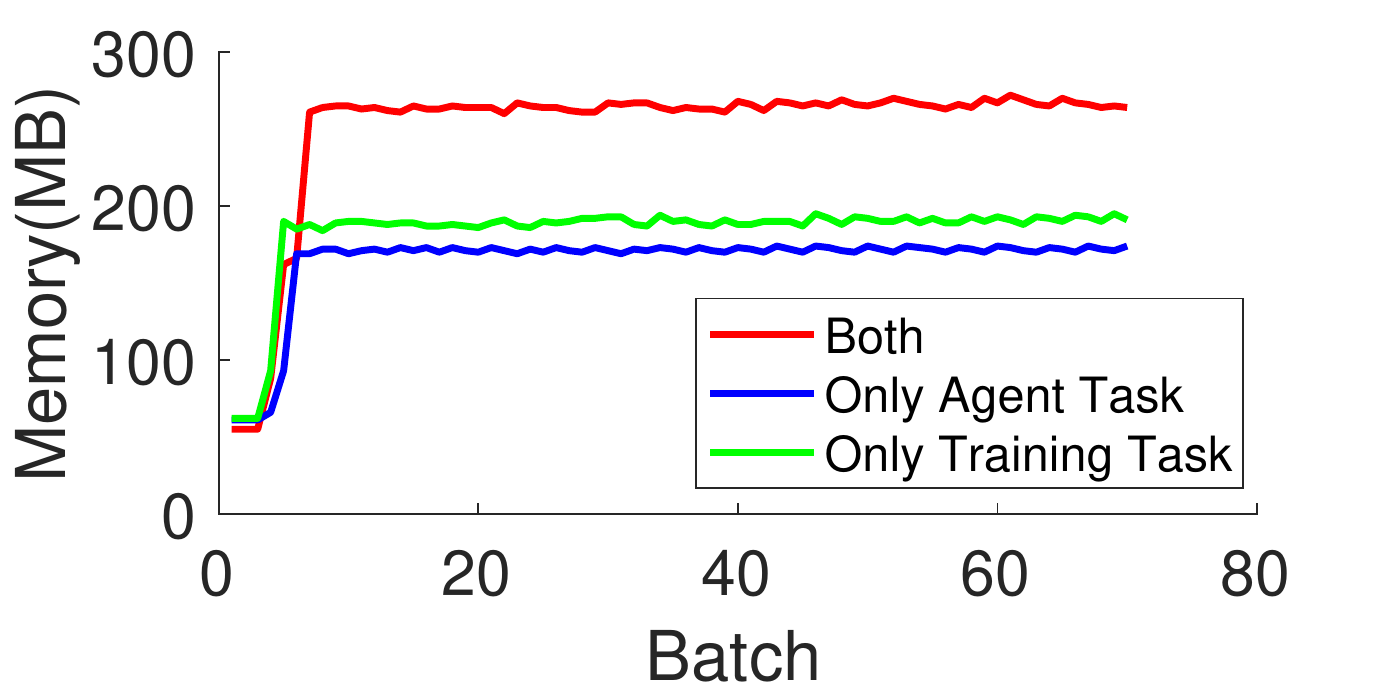}
    \label{fig:intro_memory_b}
    }
     \caption{Preliminary results. \protect\subref{fig:intro_sensing_a} Battery consumption on single-device training;  
     \protect\subref{fig:intro_memory_a} Memory crash on CNN-LSTM;
     \protect\subref{fig:intro_sensing_b} Time cost of Master vs. Sequential model aggregation;
      \protect\subref{fig:intro_memory_b} Memory conflict when running multi-agents.}
 \label{fig:intro_memory}
\end{figure}

To investigate the limitations of deploying mobile DL on device in current solutions, we conduct preliminary studies to discover the bottleneck of on-device training.

\textbf{Setup}
We select three well-known public datasets with different scales, i.e., sEMG (25MB) \mbox{\cite{Lobov:2018:EMG}}, HAR (160MB) \mbox{\cite{Anguita:2013}}, and OPPO (500MB) \mbox{\cite{Anguita:2013}} in Table \mbox{\ref{table:dataset}}. 
We use an off-the-shelf Android smartphone, i.e., Google Pixel 2XL running Android 8.1. 
In each experiment, the smartphone runs complete training with a dataset using DL4J libraries \mbox{\cite{dl4j:nd4j}}.

\textbf{Bottleneck of Single-device Training}
The encouraging result is that single-device training with LeNet for both sEMG and HAR achieves a fair accuracy with no failure.
As we raise the memory threshold, the run-time memory usage of training is moderately increased with no major issues observed.  
However, from Figure \ref{fig:intro_sensing_a}, we observe that battery overhead dramatically rises with the growth of data size and model complexity. In particular, with a maximum battery capacity of 3500 mAh on the smartphone, the training of HAR with AlexNet and the training of OPPO with both LeNet and AlexNet fail due to massive battery drain. Thus, single-device training on off-the-shelf smartphone is not currently feasible.

\textbf{Memory Overhead of Master Model Aggregation} \label{section:memory_issue}
To have quantitative measurements, we implement Federated Learning \mbox{\cite{DBLP:journals/corr/abs-1902-01046}} on device without central server support, and run the master model aggregation process (i.e., memory usage is O(N)) on a master device to evaluate its resource utilization. Each experiment runs on the same smartphone receiving $N$ copies of model gradient parameters, where $N$ denotes the number of other slave devices. 
We select PAMA2 dataset from Table \mbox{\ref{table:dataset}} to train a CNN-LSTM hybrid model \mbox{\cite{Xiang:2018}} (i.e., 46.4MB model file and 11.61M model parameters).
Figure \mbox{\ref{fig:intro_memory_a}} illustrates its memory usage since memory is more critical than battery due to model aggregation. 
We observe that memory usage is rapidly increased with $N$. When $N=8$, it ends up with \textit{OutOfMemory} (i.e., maximum 512MB). Therefore, multi-device training in a centralized framework has severe memory limitations for model aggregation.

\textbf{Cost of Sequential Model Aggregation}
To solve the memory limitations, a naive sequential model aggregation approach (i.e., memory usage is O(2)) is implemented by sequentially loading model files into memory on the master device. We use the same CNN-LSTM hybrid model to evaluate the time cost. We simulate $N$ is up to 20 to increase the scalability of model aggregation. 
Figure \mbox{\ref{fig:intro_sensing_b}} indicates the time usage of sequential aggregation is way higher than using master aggregation (e.g., 47307ms vs. 160ms respectively when $N=8$). Besides, the time usage of sequential aggregation presents a linear increase with $N$, which is also critical in practice.
Moreover, since two copies of model files are aggregated in memory space for each loading, the NeighborSGD in Eq. (\mbox{\ref{eq:sgd_de}}) is used in this implementation.
However, a severe accuracy degradation issue is revealed as shown in Figures \mbox{\ref{fig:sgd_error_6}} and \mbox{\ref{fig:sgd_error_6}} (\S\mbox{\ref{section:sgd_intro}}).
Thus, although sequential model aggregation simply minimizes the memory usage, it can still cause considerable extra training time cost and accuracy degradation.

\textbf{Resource Conflict in MARL} \label{section:scheduling_issue}
Our preliminary experiment reveals that severe resource conflict exists in MARL. We implement a simplified MARL \mbox{\cite{ben:2017:RL}} on the same smartphone using DL4J libraries. Figure \mbox{\ref{fig:intro_memory_b}} indicates that running training and agent tasks concurrently on-device can dominate the main memory of the device, resulting in 0.6x higher than that of running any single task. We thus turn our intention to a single agent-based RL scheduling approach, which motivates our proposal.




\section{MDLdroid FRAMEWORK} \label{section:system}

In this section, we detail the system architecture of MDLdroid.
We also present the proposed ChainSGD-reduce approach and Chain-scheduler.

\subsection{System Architecture}

\begin{figure}
  \centering
    \includegraphics[width=\linewidth]{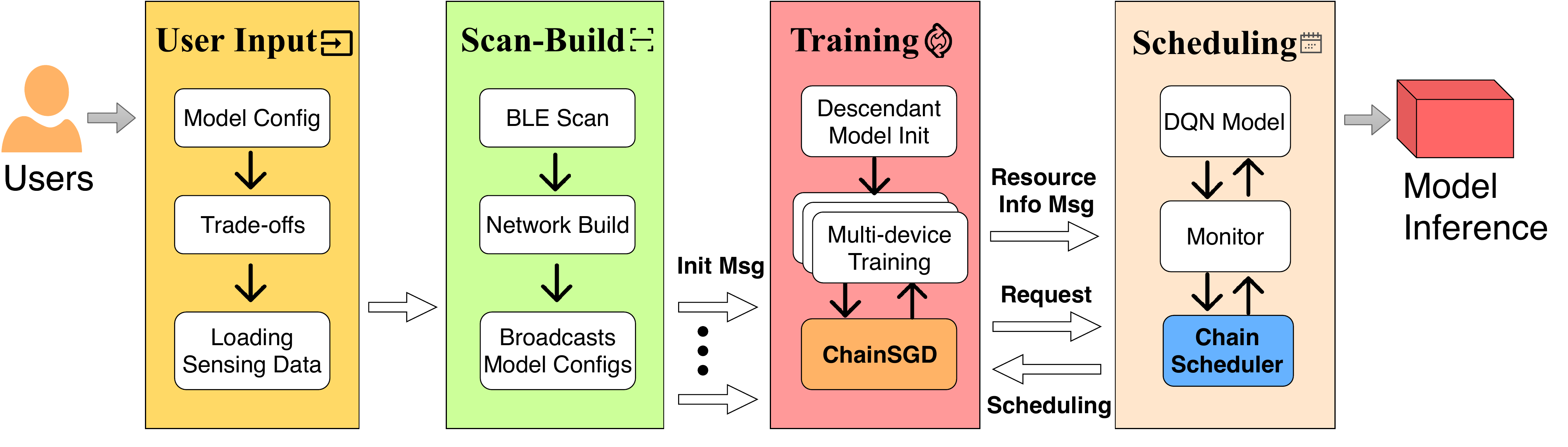}
  \caption{MDLdroid Architecture consists of three-stage processes: 1) user model configuration input; 2) network scan \& build; 3) model training \& task scheduling. 
  }
  \label{fig:sys_overview_a}
\end{figure}

We first give an architecture of MDLdroid in Figure \ref{fig:sys_overview_a}. 
Since MDLdroid is designed to operate full-scale DL on Android based on a mesh network, we employ both Bluetooth Low Energy (BLE) and Bluetooth Socket (BS) to build the mesh network 
due to accessibility and low energy consumption. 
In principle, any on-device mesh-based protocol can be applied (\S\ref{section:chain-reduce}).

In the \textbf{first} stage, users input different model configurations based on their demand to train models.
MDLdroid uses two combined network topology in the \textbf{second} stage.
The BS-based mesh topology is applied to perform the decentralized model aggregation between devices in the training stage, while the BLE-based tree topology is used for the centralized resource condition monitoring in the task scheduling stage. 
In the \textbf{third} stage, each device is required to continually report resource condition to a mobile agent (MA).
Once all training tasks request the model aggregation for each iteration,
the Chain-scheduler on the MA can manage the resource-efficiency scheduling paths as a \textit{chain-directed} graph.
When model aggregation of each iteration are completed, a copy of the aggregated model parameters will be resource-aware broadcasted (\S\ref{section: broadcast}) to all devices for the next iteration. 
\textbf{Finally}, once training is completed, a global model will be distributed to all devices via broadcasting for model inference 
Especially, MDLdroid can also reload a pre-trained model to continually train with new local sensing data.



\subsection{ChainSGD-reduce Approach} \label{section:chain-reduce}

Collaborative learning relies on the distributed \textit{stochastic gradient descent} (SGD) algorithms \cite{Jakub:2016}. To achieve reliable training accuracy, a central server typically gathers local gradient parameters $\Delta w_{i}^{k}$ from all machines and aggregates them to be global gradient parameters $\Delta w^{(k)}(t)$ after each training iteration. 
The central server then updates the local parameters $w^{(k)}(t+1)$ for all machines via a one-to-many broadcasts \cite{Gupta:2016}: 
\begin{equation}\label{eq:sgd1}
\begin{split}
\Delta w^{(k)}(t) & = \frac{1}{N}\sum_{i=1}^{N}\Delta w_{i}^{k} \\
w^{(k)}(t+1) & =w^{k}(t)-\eta \Delta w^{k}(t)
\end{split}
\end{equation}
where $w^{(k)}(t)$ denotes the $k^{(th)}$ parameters at each training iteration $t$, $\Delta w_{i}$ denotes a local gradient parameter from the $i$th machine, $N$ is the number of machines and $\eta$ is the learning rate.

\subsubsection{\textbf{Asynchronous SGD vs. Synchronous SGD}}\hfill
The distributed SGD algorithms are mainly classified into two categories---Asynchronous SGD (A-SGD) and S-SGD \mbox{\cite{Dean:2012:LSD:2999134.2999271}}. 
A-SGD can be more communication efficient and runs with no strong dependency among machines. However, study \mbox{\cite{Kwangmin:2019}}
points out that A-SGD suffers from an uncertain training accuracy degradation issue due to its delay model updating mechanism. 
S-SGD representing in Eq. (\mbox{\ref{eq:sgd1}}), on the other hand, runs quite stable without this issue, but the drawback of S-SGD is that some machines may be slowed down in run-time due to dynamic low resources, and hence the overall training time depends on the slowest machine. 
By given limited training condition on mobile device, S-SGD can effectively achieve a higher accuracy and more reliable training performance than A-SGD, which motivates S-SGD in our approach.

\subsubsection{\textbf{Chain-directed Synchronous SGD}} \hfill \label{section:sgd_intro}

In a decentralized DL framework, the relationship between training nodes and the central node is decoupled. Therefore, the centralized model aggregation can be separated into multiple descendant aggregations by structural transformation. 
The existing work
\cite{Jiang:2017:CDL:3295222.3295340} 
present decentralized SGD algorithms based on a \textit{fixed directed} network. 
We define a decentralized topology with a \textit{fixed directed} graph as $(V,  E)$, where $V$ denotes a set of devices and $E$ represents a set of edges. When we have $N$ training devices, $V=\left \{ 1, 2,...,N \right \}$ and $E\in \mathbb{R}^{V\times V}$. We define \textit{directed} edges as $\left ( i,j \right )\in E$, which means device $i$ can send its gradient parameter to device $j$ for model aggregation. The number of device $j$'s neighbors is $m$, and the number of device $j$ is $n$, where $m, n\in N$. With these settings, we can transform the centralized model aggregation (CentralSGD) in Eq. (\ref{eq:sgd1}) to the decentralized neighbor aggregation (NeighborSGD) as follows.
\begin{equation}\label{eq:sgd_de}
\begin{split}
 \Delta w^{(k)}(t) & = \frac{1}{n}\sum_{j=1}^{n}  \left ( \frac{\sum_{i=1}^{m}  \Delta w_{i}^{k} + m\Delta w_{j}^{k}}{2m}  \right )
\end{split}
\end{equation}
However, although the model aggregation is divided by $j$ descendant neighbor aggregations, the single device $j$ still requires to concurrently perform $m$ model aggregations, which may cause significant memory overhead revealed by our preliminary study. 
On the other hand, considering the dynamicity of resources on device is uncertain during training, the neighbor model aggregation based on the \textit{fixed directed} decentralized topology may cause distinct latency if device $j$ pauses the process due to low-resource condition. 

To reduce memory overhead and latency, we propose a ChainSGD-reduce approach with a mesh-based decentralized topology. 
In this approach, $m$ is constantly managed as one for every neighbor aggregation to achieve a \textit{minimal-peak} in memory and communication overhead for both devices $i$ and $j$. Our approach also includes an agent-based RL Chain-scheduler to schedule the neighbor aggregation task as a dynamic \textit{chain-directed} graph in a resource efficiency way. 

\begin{figure}
  \centering
    \includegraphics[width=\linewidth]{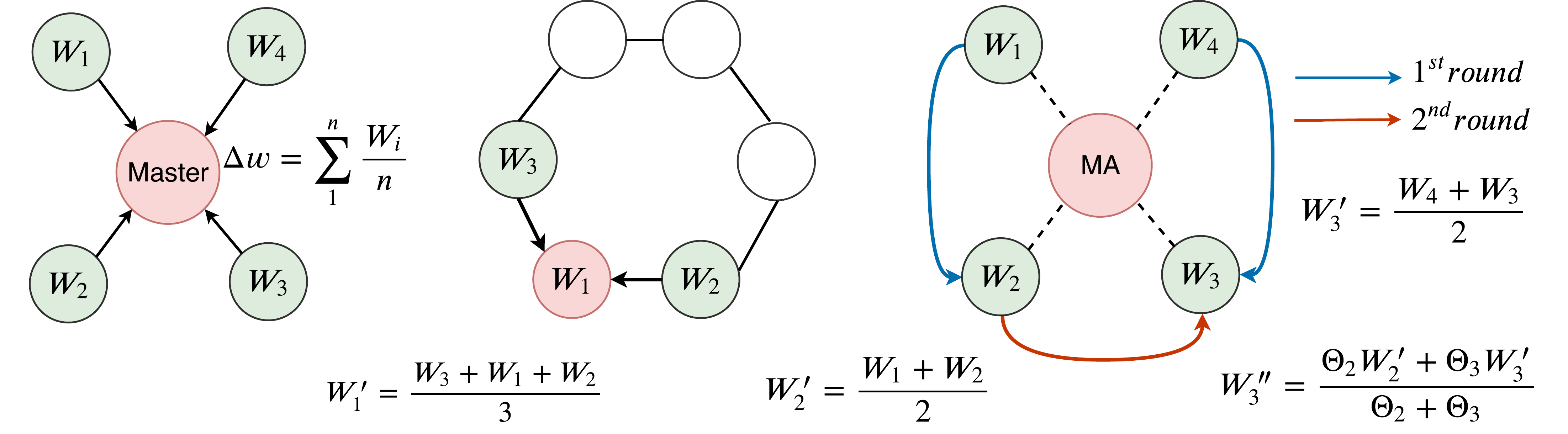}
  \caption{Model aggregation structure comparison: CentralSGD vs. NeighborSGD vs. ChainSGD}
  \label{fig:aggregation}
\end{figure}

Compared with centralized graph and decentralized neighbor graph, Figure \ref{fig:aggregation} demonstrates that the major differences of ChainSGD-reduce are twofold: 
1) the model aggregation is managed only with one of neighbors at a time;
2) the order of the aggregation tasks is dynamically scheduled depending on the real-time resource condition of device.

\begin{figure}
 \centering
    \subfloat[]{
    \includegraphics[width=0.5\linewidth]{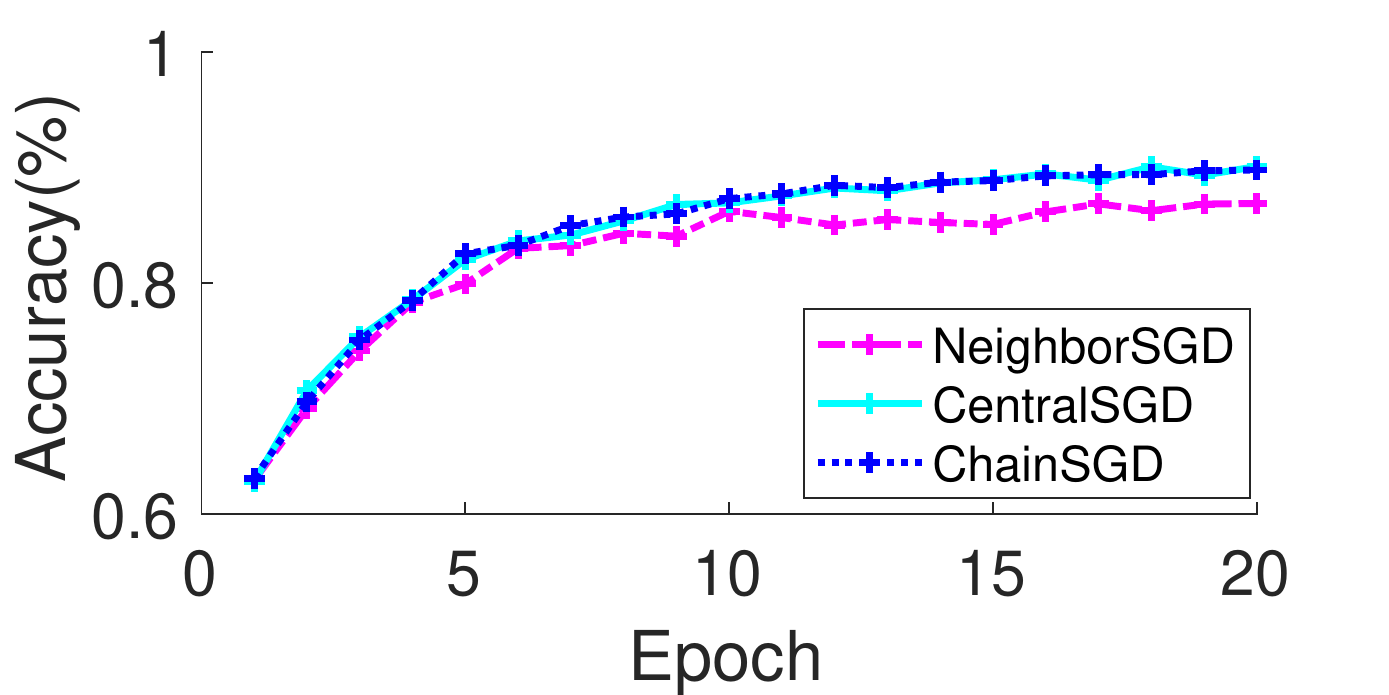}
    \label{fig:sgd_error_6}
    }
    \subfloat[]{
    \includegraphics[width=0.5\linewidth]{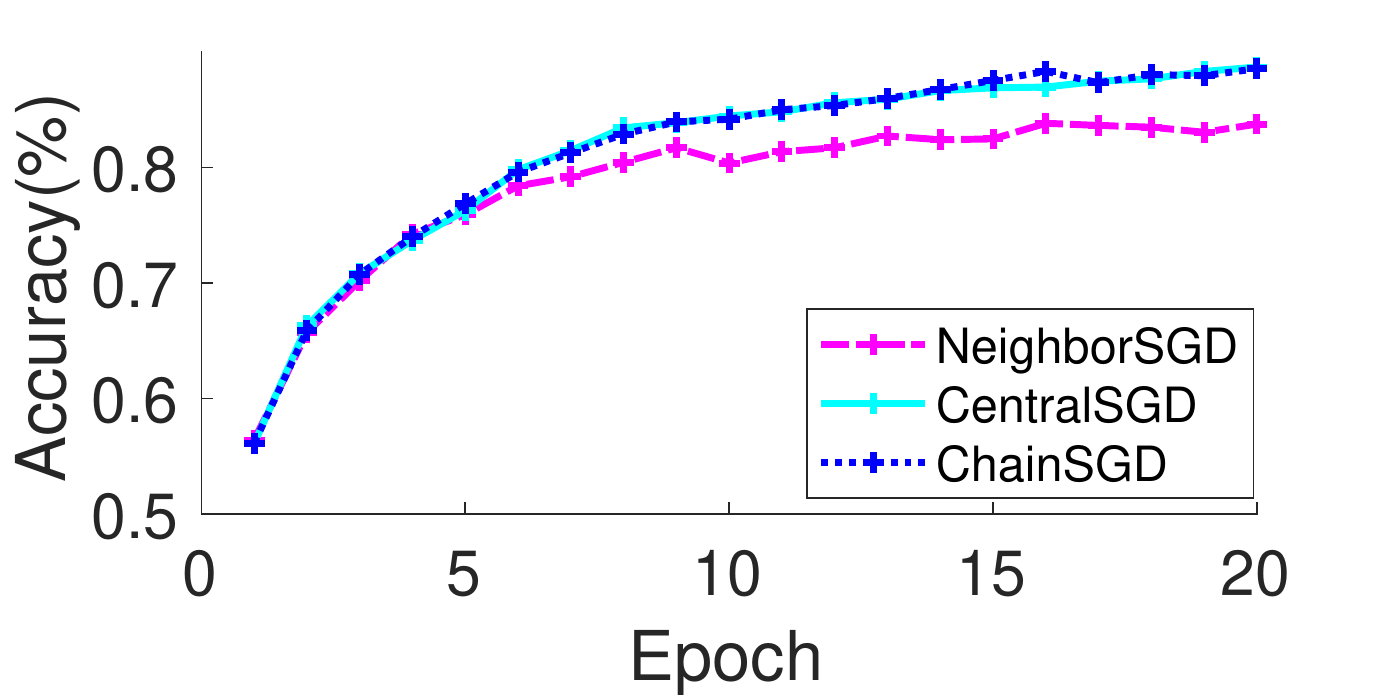}
    \label{fig:sgd_error_9}
    }
     \caption{Training accuracy degradation and restore when the number of device increases \protect\subref{fig:sgd_error_6} $N=6$; \protect\subref{fig:sgd_error_9} $N=9$.}
 \label{fig:sgd_error}
\end{figure}

When ChainSGD-reduce is applied to NeighborSGD in Eq. \ref{eq:sgd_de}, we reveal an accuracy degradation issue.
Figure \ref{fig:sgd_error_6} shows that the training accuracy of NeighborSGD is lower than that of CentralSGD by 3\% ($N=6$). 
Besides, Figure \ref{fig:sgd_error_9} shows that with $N$ increases, the training accuracy gap between NeighborSGD and CentralSGD becomes larger by roughly 5\% less ($N=9$). 
It is because that the global gradient parameters $\Delta w^{(k)}(t)$ decompose through the model aggregation process. 
We thus redefine ChainSGD as the following \textit{pair aggregation} function $W(j,i)$ aiming to restore training accuracy. 
\begin{equation}\label{eq:sgd_error}
W(j,i) =\left\{\begin{matrix}
 \frac{\theta _{i}\Delta w_{i}^{k} + \theta _{j}\Delta w_{j}^{k} }{\theta _{i} + \theta _{j}} \\ 
\theta_{i}^{'}=\theta _{i} + \theta _{j}
\end{matrix}\right.
\end{equation}
where $W(j,i)$ represents the aggregated gradient parameters sending from a remote device $j$ to device $i$ as a \textit{pair aggregation}.  
The global gradient parameters of each iteration are denoted as $\Delta w^{(k)}(t) = (N-1)W(j,i)$. Since ChainSGD offers a dynamic model aggregation process, multiple \textit{pair aggregation} can be performed simultaneously based on resource condition of devices within one round to reduce latency showing in Figure \ref{fig:aggregation}.
Especially, $\theta$ is a reversal parameter to enable that the gradient parameters of $W(j,i)$ can be restored close to CentralSGD, and $\theta$ starts from 1. Once a \textit{pair aggregation} is done, the local $\theta_{i}$ will be updated with a remote $\theta_{j}$ as $\theta_{i}^{'}$ for sending to next round. 
Our approach requires that the BS message for model aggregation contains $\Delta w^{k}$ and $\theta$ as $(\Delta w^{k}, \theta)$. 
Both Figures \ref{fig:sgd_error_6} and \ref{fig:sgd_error_6} show that the performance of ChainSGD is consistent with that of CentralSGD with no accuracy degradation.

In summary, the ChainSGD-reduce algorithm is shown in Algorithm \ref{al:chain} below.
\begin{algorithm}
Initialize parameters $w_{0}$, number of iteration $t$, number of devices $N$, training dataset $D_{train}$;\\ 
\For{All devices $i\in N$}{
\For{$t$ Epochs}{
Train model on local $d_{train}^{t}\in D_{train}$\;
\While{true}{
\If{Get an incoming BS message $(W(j,i), \theta)$}{
\textit{Pair aggregation} $\Delta w_{i}^{k}$ in Eq. (\ref{eq:sgd_error})\;
Update $\theta$\;
}
\Else{
Get a remote neighbor $j$ from MA\;
Send $(\Delta w_{i}^{k}, \theta)$ to neighbor $j$, Break\;
}
}
Update local $w_{i}^{(k)}(t+1) = w_{i}^{k}(t)-\eta \Delta w^{k}(t)$\;
}
}
\caption{ChainSGD-reduce Algorithm}\label{al:chain}
\end{algorithm}



\subsubsection{\textbf{Resource-aware Broadcasting}} \hfill \label{section: broadcast}

In ChainSGD-reduce, the MA device handles the global model parameters broadcasting after each training iteration. If the MA device sequentially broadcasts by BS messages, it may suffer from heavy communication overhead when the number of devices $N$ is large. 
We employ a binomial-tree broadcast algorithm \cite{Hoefler:2007} with a message forwarding function to address this issue. Basically, our implementation enables the broadcast list sorted by devices' resource condition, and the devices with more resources will be assigned with a list of forwarding tasks to mitigate the overhead on MA. Thus, the number of broadcast round by MA can be reduced from $N$ to $\lceil\log_{2}{N}\rceil$.

\subsubsection{\textbf{Fault-tolerant}} \hfill
 
MDLdroid offers a fault-tolerant strategy to ensure the stability for on-device training: 
1) \textit{File cache}: model parameters and necessary records can be saved as files backup in run-time for training recovery;
2) \textit{Training device fault}: once a training device is lost after the re-connection attempts, \textit{File cache} will be first preformed. Then MA will skip the device for scheduling until reconnected. If re-connection is successful, MA will require the device to send over the last model parameters and iteration records. After that, MA will send back the current iteration records for synchronization, and the device will receive the global aggregated model parameters until next iteration.

\subsection{Resource-aware Chain-scheduler} \label{section:chain-balance}

We propose Chain-scheduler using an agent-based RL model to dynamically schedule model aggregation tasks. 
To mitigate the memory overhead analyzed in our preliminary experiment (\S\ref{section:memory_issue}), we design a single agent-based RL mechanism. The agent task is separated into an individual device as the MA which is responsible of resource-aware scheduling, and other devices are mainly responsible of running training tasks. With a mesh network, MA can globally monitor all training devices' resource condition in real-time with low-energy cost. 
In MDLdroid, two optimal goals are defined for Chain-scheduler to achieve resource-efficiency. 
One is to reduce training latency due to the resource condition of some devices dynamically turning \textit{busy} in real-time. Another is to balance communication overhead and battery consumption across the network. 

Mathematically, Chain-scheduler deals with the following constrained optimization function:
\begin{equation}\label{eq:optimal}
\begin{split}
\arg\min & \: N(T(t))+N(E(t)) \\
\text{s.t.} & \:  M\leq M_{max}, \:  B\leq B_{max}
\end{split}
\end{equation}
where $T$ and $E$ denote \textit{training time} and \textit{energy balance} across the network, respectively. 
$N(x)=(x - x_{min}) / (x_{max} - x_{min})$ is a standard normalization function to transform \textit{training time} and \textit{energy balance} to be at the same scale. $t$ represents the current training iteration.
We denote $M_{max}$ and $B_{max}$ as the maximum memory and maximum battery offered by the target device, respectively.

\textbf{Training Time $T$}
The training time for each iteration $t$ is defined as:
\begin{equation}\label{eq:train_time}
\begin{split}
T(t)=T_{tr}+f(T_{a},T_{b})
\end{split}
\end{equation}
where $T_{tr}$ denotes the training time of each individual device, $T_{a}$ denotes the overall model aggregation time for iteration $t$, and $T_{b}$ represents the overall latency time caused by the \textit{busy} state of devices. $f(x)$ is the schedule function. Especially, due to concurrent processes, $f(x)$ can schedule $T_{a}$ and $T_{b}$ to reduce iteration latency. 
The worse case is $T_{a}+T_{b}$ in which the scheduled sequence is linear and no concurrent overlap, while the ideal case is $\max(T_{a},T_{b})$ in which training latency is minimized as the concurrent overlap.

\textbf{Energy Balance $E$}
The energy balance for each iteration $t$ can be modeled as:
\begin{equation}\label{eq:energy_cost}
\begin{split}
E(t)=\frac{1}{N}\sum\limits_{i=1}^N (E_c^i -\mu)^2
\end{split}
\end{equation}
where $\mu$ is the mean of $\{E_{1}, E_{2}...E_{N}\}$, and $E_c^i$ denotes the energy consumption of device $i$. 
We employ a variance function to represent the balance performance of Chain-scheduler.

The best estimated scheduling sequence represents as a reverse tree structure of \textit{Tree-scheduler} \cite{Shaohuai:2019} 
to reduce latency, which requires $\lceil\log_{2}{N}\rceil$ communication rounds. While the worse case represents as a ring structure of \textit{Ring-scheduler} \cite{Li:2014:SDM:2685048.2685095} with $N$ communication rounds. 
However, \textit{Ring-scheduler} can perform the best energy balance as each device only requires to aggregate model once.  
Thus, Chain-scheduler is aiming to achieve optimal trade-off between \textit{Tree-scheduler} and \textit{Ring-scheduler}.

\begin{figure}
 \centering
    \subfloat[]{
    \includegraphics[width=0.48\linewidth]{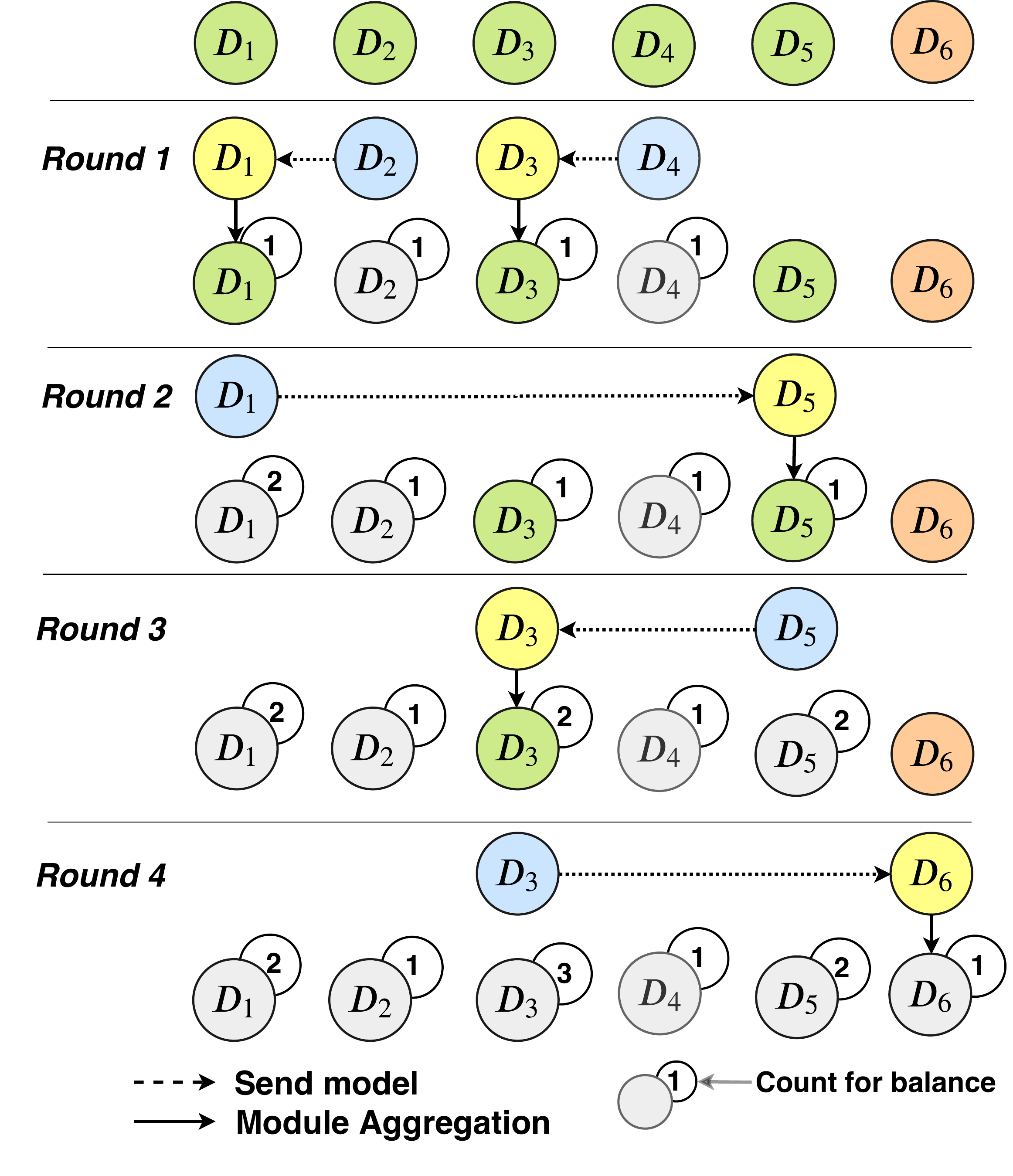}
    \label{fig:state_tree_a}
    }
    \subfloat[]{
    \includegraphics[width=0.52\linewidth]{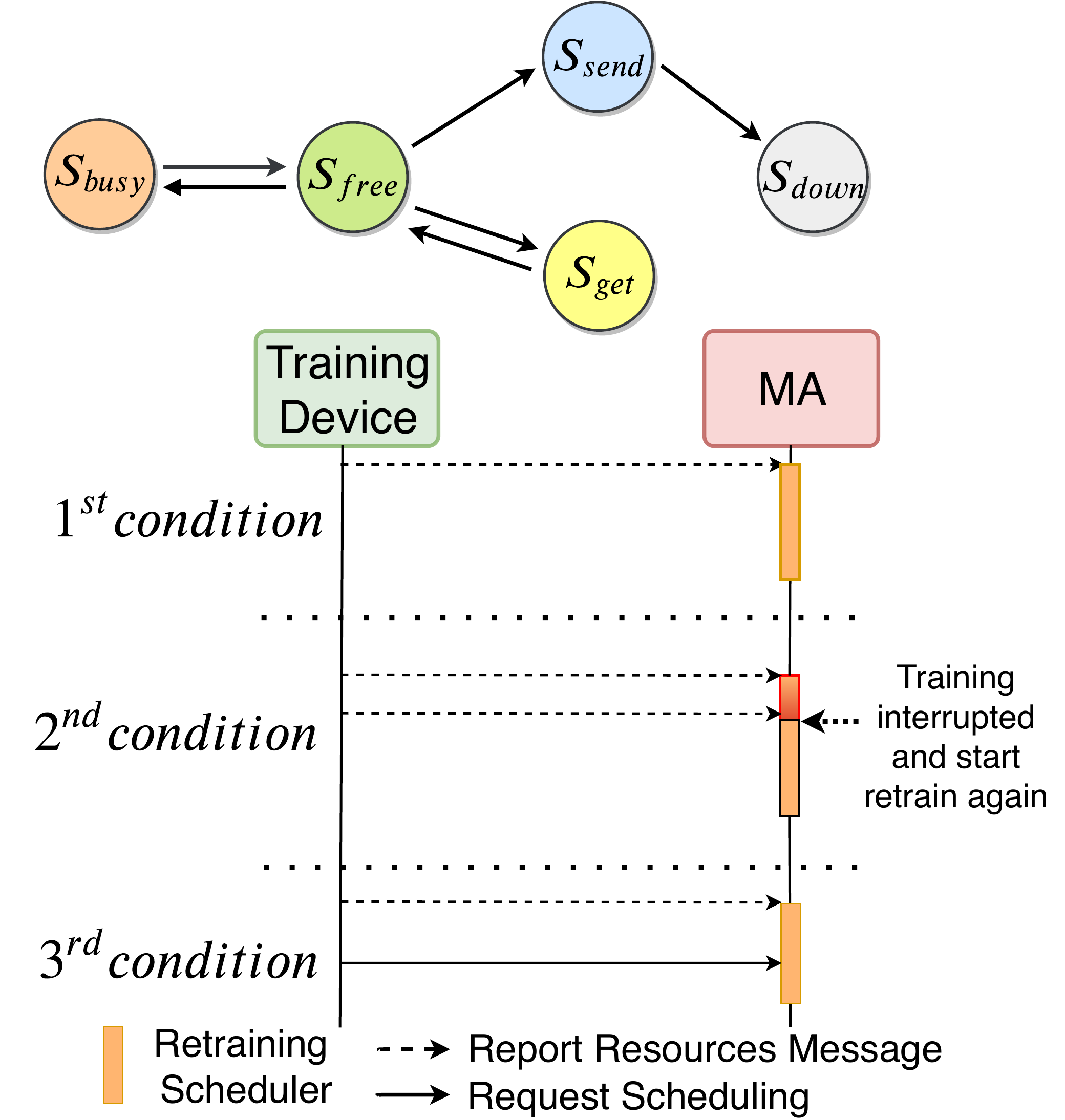}
    \label{fig:state_tree_b}
    }
     \caption{\protect\subref{fig:state_tree_a} The process of scheduling by busy condition; \protect\subref{fig:state_tree_b} State diagram and Re-training mechanism}
 \label{fig:state_tree}
\end{figure}

\subsubsection{\textbf{Multi-goal Reward Function}} \hfill

We employ a DQN model 
\cite{Thanh:2018} in MA to learn the scheduling environment based on the optimization function in Eq. (\ref{eq:optimal}). 
In our protocol, all train devices are required to continually report their resource condition to MA. 
We select five essential resource parameters including: \textit{free memory} (i.e., remaining memory), \textit{battery} (i.e., remaining battery), \textit{in-use} (i.e., whether if under intensive use condition), \textit{charge} (i.e., whether if charging), \textit{cpu} (i.e., current CPU), to identify whether the device is in \textit{busy} ($S_{busy}$) or \textit{free} ($S_{free}$). Next, we present the design of the Chain-scheduler structure:

\begin{itemize}[leftmargin=*,noitemsep,topsep=0pt]
\item \textbf{State}: We design five states for Chain-scheduler to make crucial scheduling decisions based on our protocol. $s=\{s_{t}, i=1, 2, ..., T \}$ represents as a set of devices' states, where $t$ denotes the learning step. 
$States = (S_{free}, S_{busy}, S_{send}, S_{get}, S_{done})$, where $S_{send}$ denotes that the device is sending model parameters, and $S_{get}$ represents the device is getting model parameters. Specifically, the state of devices can be only defined as one of the five states at any learning step. Figure \ref{fig:state_tree_b} illustrates the relationship of these states. 

\item \textbf{Action}: We define $a=\{a_{i}, i=1, 2, ..., N \}$, where $N$ denotes the number of devices. It represents which device is selected by the scheduler to act in the environment. 

\item \textbf{Reward}: $r=\{r_{t}, t=1, 2, ..., T\}$ is defined by the reward function $r(s_{t},a, s_{t+1})$ in Eq. (\ref{eq:reward}).

\end{itemize}

\textbf{Reward Model}
To solve the optimization function in Eq. (\ref{eq:optimal}), we design a three-stage mechanism as follows to offer a fast learning process. 
\textbf{Firstly}, to reduce training latency, we design a decaying penalty function $\rho + t{\rho}/{2(N-1)}$ aiming to set $S_{busy}$ as the lowest priority. Since our protocol requires that a pair ($i, j$) should be selected by each scheduling, it takes two steps in the learning process. Then, $2(N-1)$ represents the \textit{best learning steps} at one \textit{Epoch}. With $N$ increases, the penalty reward of $S_{busy}$ decreases. Thus, the later $S_{busy}$ chooses, the smaller penalty reward the model gets. 
From concurrent perspective, the training latency can be reduced as the process of model aggregation can be done with $S_{free}$ in advance. 
\textbf{Secondly}, to balance energy consumption, if the count of ${S_{free}} \rightarrow{S_{get}}$ for each device is close to the average, the energy variance of all devices in Eq. (\ref{eq:energy_cost}) can reach the minimal. Hence, we design a penalty function $\alpha-\beta n_{agg}$, where $n_{agg}$ records the count of ${S_{free}} \rightarrow{S_{get}}$. With $n_{agg}$ increases, the penalty reward also increases. 
In addition, we design an incentive function $\alpha+\beta n_{agg}$ to speed up the model learning. With $n_{agg}$ increases, the model can get more rewards if ${S_{free}} \rightarrow{S_{send}}$ is done. 
\textbf{Thirdly}, to achieve fast learning, we design an efficient \textit{termination} function. If the selected action is a repeat as an \textit{invalid} action or the learning step is larger than the \textit{best learning steps}, the model gets a severe penalty reward and the current learning step is terminated. In summary, we present our \textbf{reward function} as follows.
\begin{equation}
             r(t) =\left\{
                \begin{array}{lcl}
                \alpha-\beta n_{agg} & & {S_{free}} \rightarrow{S_{get}}\\
                 \alpha+\beta n_{agg} & & {S_{free}} \rightarrow{S_{send}}\\
                \rho + t{\rho}/{2(N-1)}& & {S_{busy}}\\
                 -1 & & {Exceeds \: limits}\\
                 -1 & & {Select \: invalid \: action}\\
                 1 & & {Completed}
                \end{array}
                \right.
    \label{eq:reward}
\end{equation}

where $\alpha$, $\beta$ and $\rho$ denote initial reward, initial busy penalty and battery cost per aggregation, respectively. $n_{agg}$ denotes the number of model aggregations. By default, we set $\alpha$ to $-0.04$, $\beta$ to $0.1$ and $\rho$ to $-0.8$.

\subsubsection{\textbf{Accelerated Reward Function}} \hfill \label{section:explore}

\begin{figure}
  \centering
    \includegraphics[width=\linewidth]{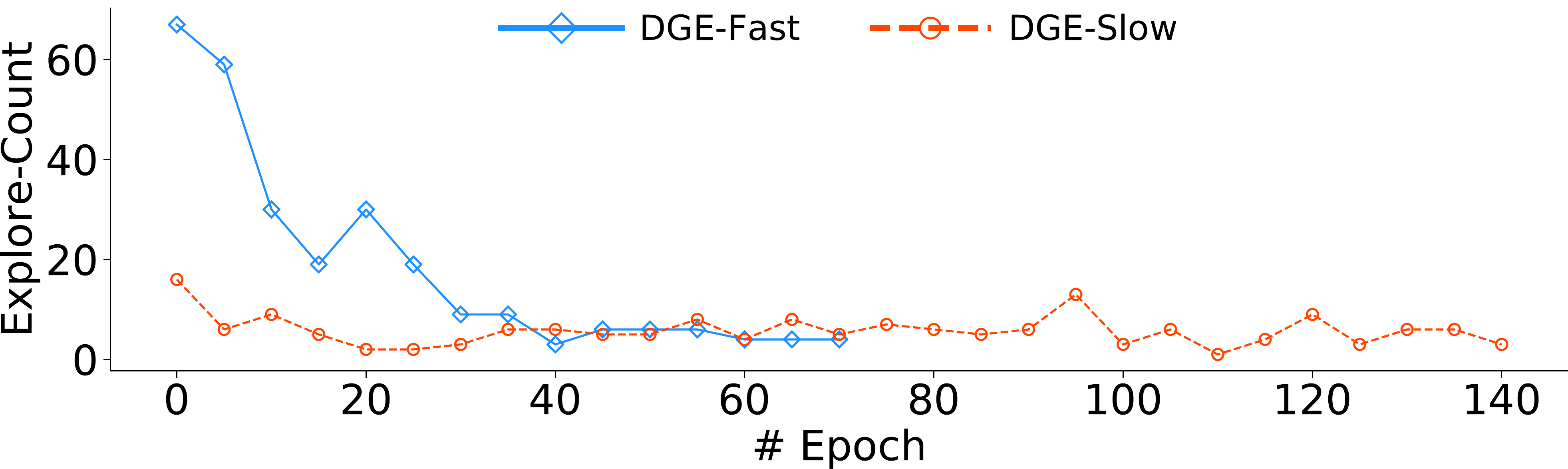}
  \caption{RL training speed comparison by using different exploration rate}
  \label{fig:RL_exploration}
\end{figure}

To further accelerate the RL learning process, we propose a threshold-based decaying greedy-exploration (TDGE) strategy which extends the existing decaying greedy-exploration (DGE) strategy \cite{Tokic:2010}.
The decaying greedy-exploration (DGE) strategy has been used as a default option with a fair performance.
However, Figure \ref{fig:RL_exploration} shows DGE-Fast (using 2x exploration rate) completes earlier than DGE-Slow, i.e., each stops at 73 \textit{Epoch} and 143 \textit{Epoch}, respectively.  
A key observation from our empirical study is that more sufficient exploration achieves more efficient learning time reduction.
Therefore, in the proposed TDGE strategy, we aim to ensure sufficient exploration in the beginning to accelerate the learning process. The interpretation is to ensure the model to fully explore until the reward reaches the given threshold, we then switch the exploration to start epsilon decaying from the predefined value. Specifically, the threshold is approximately defined by achieving optimal latency and battery balance at the same time in Eq.(\ref{eq_threshold}) since the reward function is designed to optimize both features by Eq. (\ref{eq:optimal}). The estimation of optimal latency is defined as reward value by Eq. (\ref{eq_RLlatency}), where the estimation of optimal battery balance is defined in Eq. (\ref{eq_RLbattery}).

We define threshold $T(N)$ by separating the calculation of the optimal reward into two parts in Eq. (\ref{eq_threshold}), which imitate architecture in defining optimization function of Chain-scheduler. The first part represents the maximal reward value of battery balance by Eq. (\ref{eq_RLbattery}). The second part denotes the optimal reward value of latency by Eq. (\ref{eq_RLlatency}).

\begin{equation} \label{eq_threshold}
T(N) = f(N) + g(N) - \Phi \\ 
\end{equation}
where $\Phi$ denotes an offset value obtained from our experiment. 

\begin{equation} \label{eq_RLbattery}
f(N) =\left\{\begin{matrix} 
\beta (N \: mod \: 2)  + f(\lfloor N/2 \rfloor) &  N > 1\\ 
0    & N=0\And{1}
\end{matrix}\right.
\end{equation}

\begin{equation} \label{eq_RLlatency}
 g(N) = \beta \log_{2} N 
\end{equation}

In summary, the Chain-scheduler algorithm is shown in Algorithm \ref{al:dqn}.


\begin{algorithm}
\SetKwFunction{validAction}{validAction}
\SetKwFunction{random}{random}
    Initialize target network weights, epsilon $\epsilon = 1.0$, threshold $TR$ in Eq. (\ref{eq_threshold}), state $s$;\\
    \While{$ Epoch < maxEpoch $}{
    \For{$t \leftarrow 1$ \KwTo $maxStep$}{
    \eIf{\random{0,1} $< \epsilon$}{
        randomly explore an action from \validAction{};
        }{
        take action a with $\epsilon$-greedy policy based on Q-value function $Q(s_{t}, a_{t})$;
        }
    receive  $s_{t+1}$ and  $r_{t}$ in  Eq. (\ref{eq:reward}); \\
    $s_{t} \leftarrow s_{t+1}$;\\
    \If{$s_{t+1}$ is terminated }{
        \If{reward $> TR$}{
        $\epsilon \leftarrow \epsilon_{new}$
        }
        \bf break;
        }
    }
    update $\epsilon$ by decaying;
 }
\caption{Chain-scheduler Algorithm}\label{al:dqn}
\end{algorithm}

\subsubsection{\textbf{Continuous Environment Learning}} \hfill

As analyzed in Section (\S\ref{section:scheduling_issue}), \textit{Non-Stationarity} scenario has a negative impact on the performance of the RL model. 
However, since MA can continuously monitor device resource condition, this impact can be mitigated by a repeating environment learning mechanism \cite{JMLR:v17:14-037}. If the current resource condition is not equal to the previous record, MA can restart the RL learning process to update the model. Figure \ref{fig:state_tree_b} indicates three \textit{re-learning} conditions. In the first two conditions, the RL learning process can be restarted with a small learning slot by the resource condition changes of environment, and it finishes before the scheduling actions. For the third condition, if scheduling actions require to perform during the slot of \textit{re-learning}, Chain-scheduler can output a fair scheduling based on the history experience of the RL model.


\section{EVALUATION} \label{section:evaluation}

\subsection{MDLdroid Implementation}

\begin{figure}
 \centering
    \subfloat[]{
    \includegraphics[width=0.33\columnwidth, height=\columnwidth,keepaspectratio]{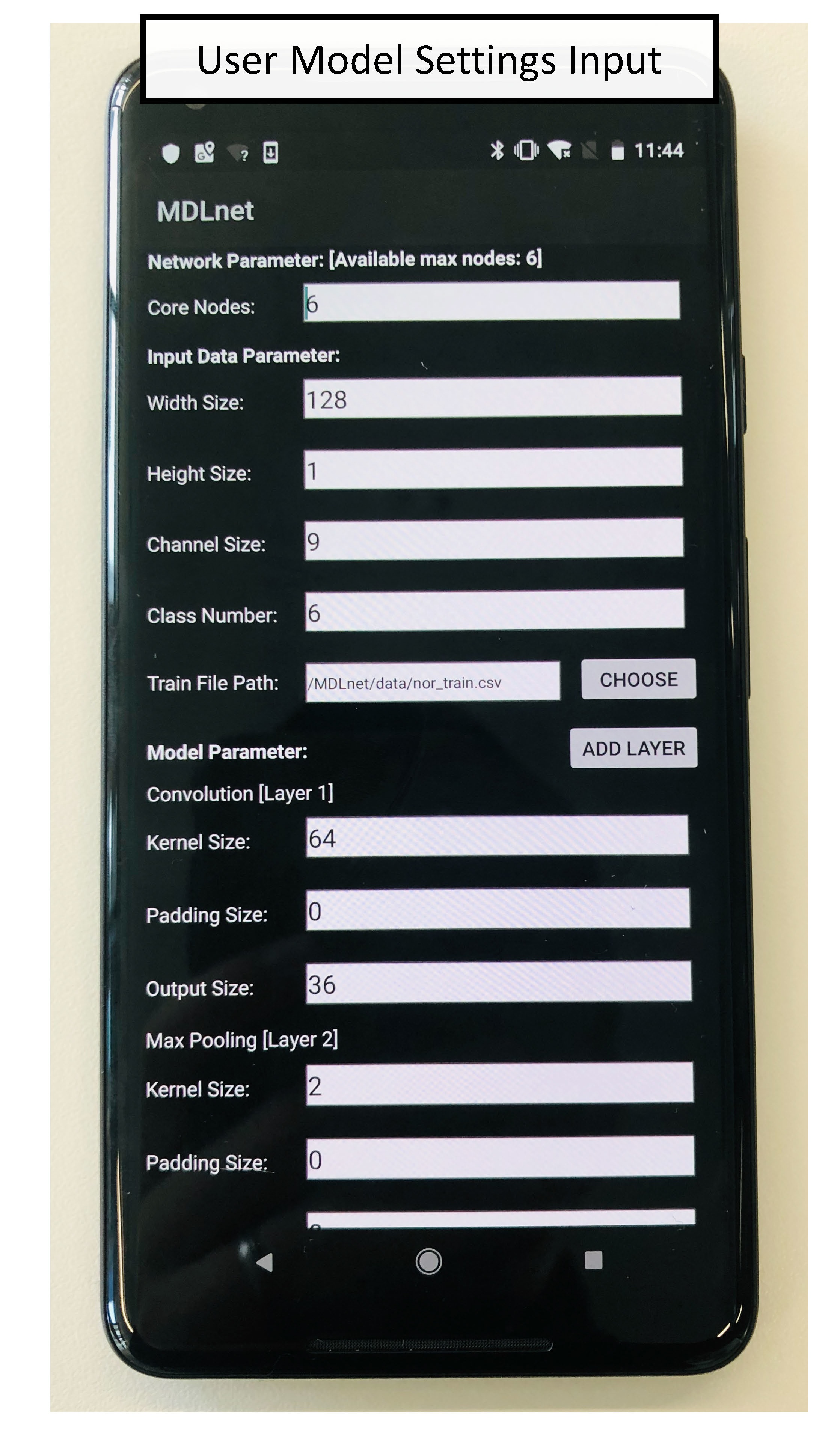}
    \label{fig:realcase_a}
    }
    \subfloat[]{
    \includegraphics[width=0.67\columnwidth]{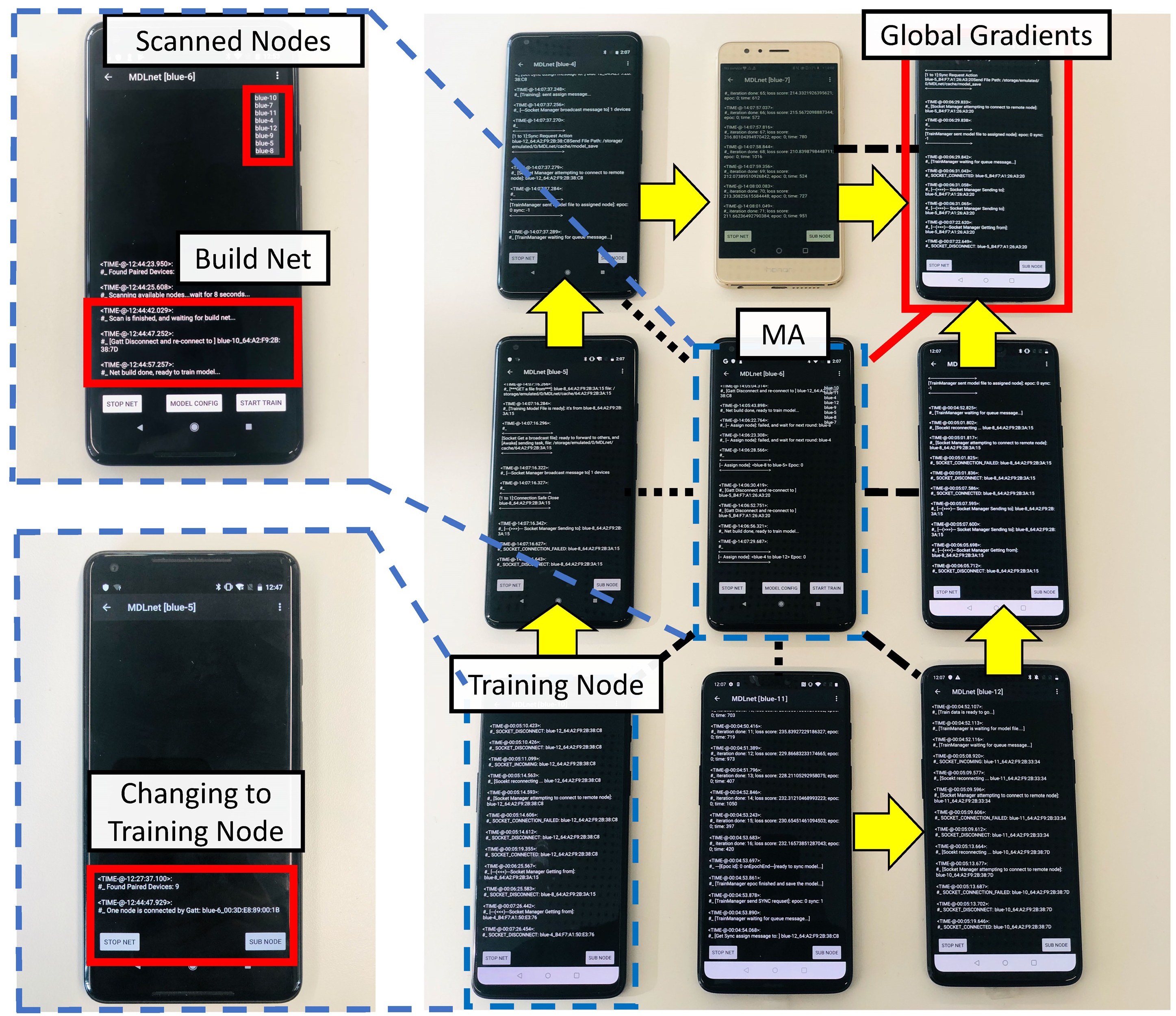}
    \label{fig:realcase_b}
    } \quad
     \caption{Experiment screenshot. \protect\subref{fig:realcase_a} User model input; \protect\subref{fig:realcase_b} Training execution screenshot}
 \label{fig:realcase}
\end{figure}

We implement MDLdroid based on an open-source DL library (i.e., DL4J).
In particular, we essentially modify DL4J to enable the proposed ChainSGD-reduce approach on device. We also tailor our implementation for execution on Android smartphone. With minor model configurations, MDLdroid is fully compatible with a range of DL models without scaling down the model. 
We employ 9 off-the-shelf Android smartphones 
and Table \ref{tab:phone_specs} gives their specifications. 
Figure \ref{fig:realcase_a} plots a screenshot in which user customizes model configuration such as the parameters for certain datasets, customized hidden layer structures, and the required number of training devices.
Figure \ref{fig:realcase_b} plots a screenshot during an execution of training on 9 smartphones using MDLdroid. The MA device scans all nearby devices, and build a BLE mesh network. 
The black dash lines represent the BLE connections between MA and training devices for resource condition monitoring. The yellow lines indicates a particular \textit{chain-directed} model aggregation process via BS. 

\subsection{Datasets}


\begin{table}
\caption{Dataset Specifications}
\label{table:dataset}
\resizebox{\columnwidth}{!} {
\begin{tabular}{llllllllll}
\hline
\textbf{Datasets}        & \textbf{Type} & \textbf{Task} & \textbf{\#-S} & \textbf{\#-C} & \textbf{\#-SA}  & \textbf{\#-R} & \textbf{\#-H} & \textbf{\#-TR} & \textbf{\#-TE} \\ \hline
\textbf{sEMG}    \cite{Lobov:2018:EMG} & EMG           & GR            & 37            & 6             & 14695            & 50   &    8   & 25     &       6  \\
\textbf{MHEALTH}   \cite{Banos:2014}      & IMU           & HBM            & 10            & 9             & 3255             & 50  &  23   & 100     &    30     \\
\textbf{UniMiB}   \cite{micucci2017SHAR}  & IMU           & FDR            & 30            & 8             & 8430             & 50  &   1  & 116 &        15     \\
\textbf{HAR}       \cite{Anguita:2013}      & IMU           & ADLs            & 30            & 6             & 10299        & 50 &     9  & 160      &     60   \\
\textbf{OPPORTUNITY}   \cite{Roggen:2010}  & IMU           & ADLs            & 12            & 11            & 16837            & 50  &   77   & 500     &     50    \\ 
\textbf{PAMAP2}   \cite{Reiss:2012:ISWC}       & IMU           & ADLs            & 9             & 12            & 12397          & 100   &  9    & 900   &    90        \\ \hline
\end{tabular}
}
\end{table}

To evaluate MDLdroid, we select 6 public personal mobile sensing datasets with a training scale ranged from 25M to 900M. These datasets are typically used for building a variety of personal mobile applications, e.g., gesture recognition (GR), recognition of activity for daily living (ADLs), fall detection recognition (FDR), and health behavior monitoring (HBM).
The specification of each dataset is listed in Table \ref{table:dataset}. 
\#-S, \#-C, \#-SA, \#-R, \#-H, \#-TR, \#-TE separately denote the number of subjects, the number of classes, the number of samples, sampling rate (Hz), the number of channels, the size of training data (MB), and the size of testing data (MB),


\subsection{Evaluation Methodology}

In our evaluation, all the participating smartphones are placed in proximity with a range from 1m to 5m for any twos.
To evaluate actual battery consumption, we set all smartphones as discharging. Before each experiment, we reset the battery of smartphones to be full. 
For static dataset allocation, we divide each dataset equally among all the training devices since they have similar resource capacity and pre-load a sub-dataset to each device in advance to simplify our evaluation. 
We first evaluate the performance of Chain-scheduler. We then compare MDLdroid to Federated Learning. 
Finally, we conduct a series of experiments to discover optimized parameters to trade-off between resource used, training accuracy and scalability.

\begin{table}
\caption{Smartphone specifications}
\label{tab:phone_specs}
\resizebox{\columnwidth}{!} {
\begin{tabular}{lllll}
 \hline
\textbf{Device}       & \textbf{HD}    & \textbf{RAM} & \textbf{CPU}     & \textbf{Battery(mAh)} \\ \hline
OnePlus 6       & 128GB & 8GB & Snapdragon 845    & 3300          \\
Pixel 2 XL      & 64GB  & 4GB & Snapdragon 835  & 3520          \\
Huawei Honor 8  & 32GB  & 4GB & HiSilicon Kirin 950   & 3000  \\       
 \hline
\end{tabular}
}
\end{table}

\subsection{Performance of Chain-scheduler} \hfill

To evaluate Chain-scheduler, we select two resource-agnostic schedulers (i.e., \textit{Tree-scheduler} and \textit{Ring-scheduler}) as the baseline. We also use both DGE and the threshold-based greedy-exploration (TGE) (i.e., the exploration only relies on the given threshold without decaying) approaches (\S\ref{section:explore}) as the baseline to compare with our TDGE approach and evaluate the performance of the exploration strategy when training Chain-scheduler. 

\subsubsection{\textbf{Experimental Setup}} \hfill

For a fair benchmark comparison, we simulate the resource dynamicity scenarios in reality for each device in MDLdroid. Specifically, we randomly allocate resources for each training device while assuring a maximum of 50\% of the devices being in \textit{busy} state. 
To evaluate our re-learning mechanism, we randomly modify the resource state of some devices being \textit{busy} or \textit{free} to emulate the conditions mentioned in Figure \ref{fig:state_tree_b}. 
We run each experiment 50 times and report the performance and resource usage presented in the following sections.

\subsubsection{\textbf{Exploration Strategy of Chain-scheduler}} \hfill

\begin{figure}
 \centering
    \subfloat[]{
    \includegraphics[width=0.5\linewidth]{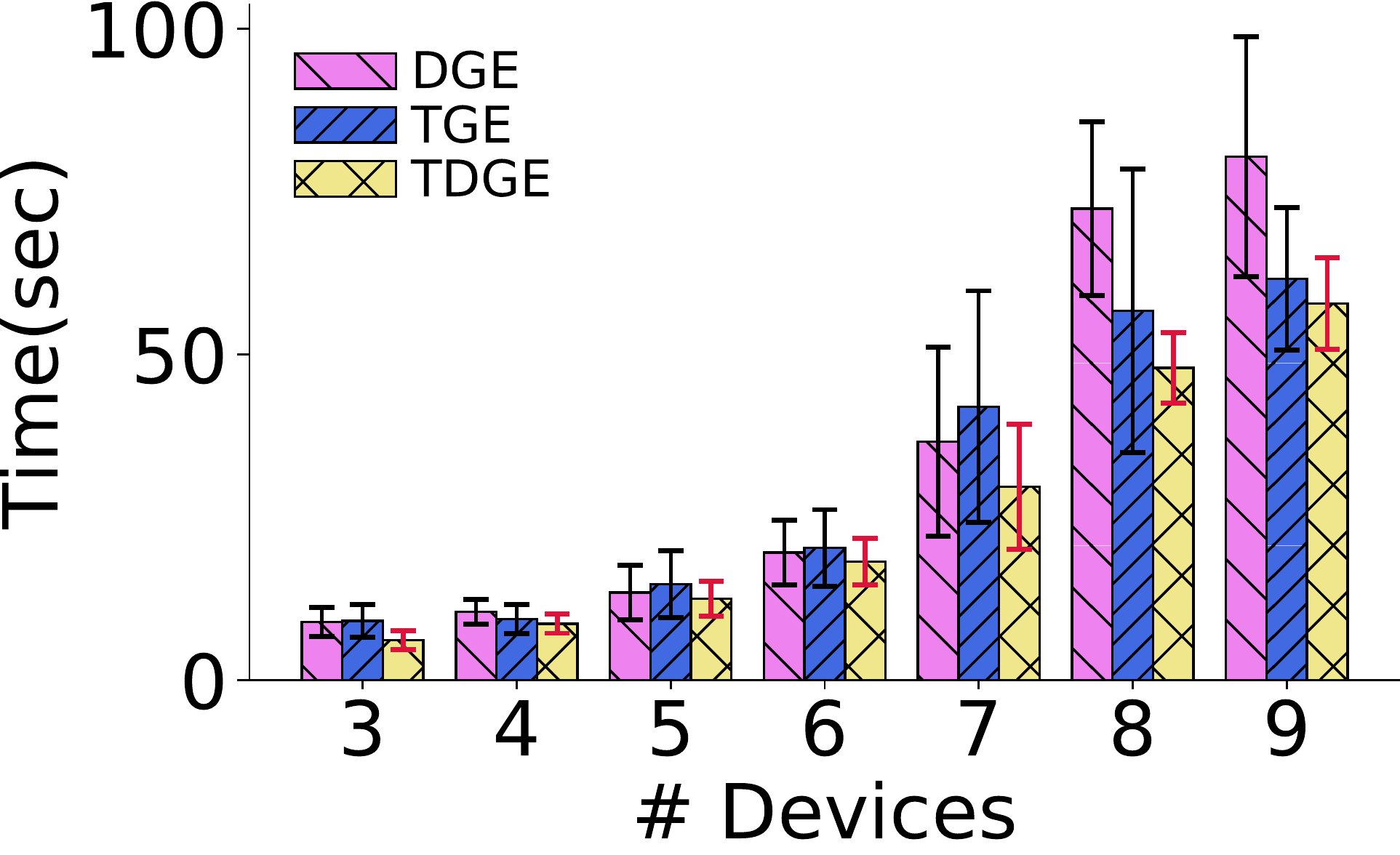}
    \label{fig:rl_train_speed}
    }
     \subfloat[]{
    \includegraphics[width=0.5\linewidth]{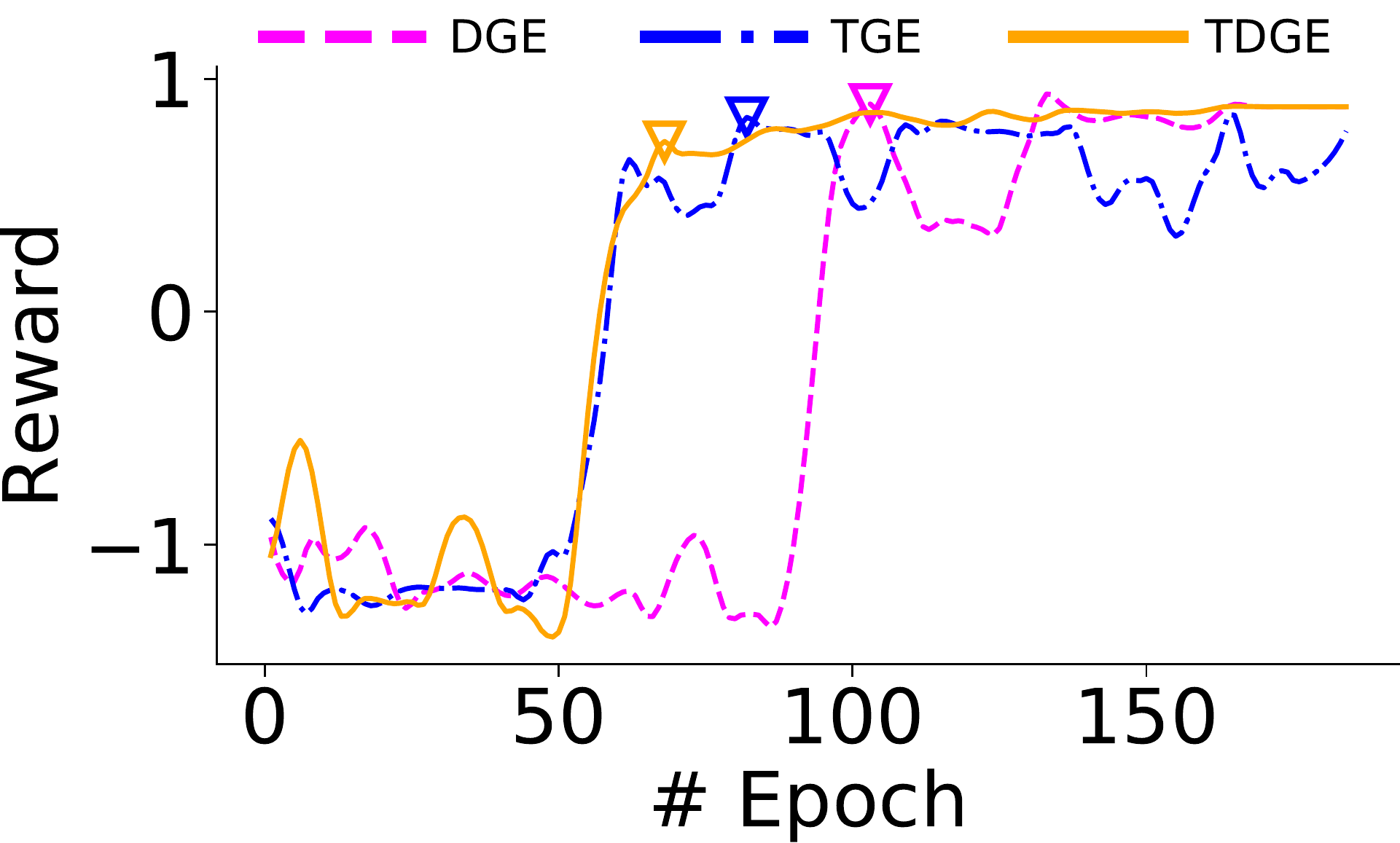}
    \label{fig:rl_reward_compare}
    }
     \caption{Chain-scheduler training comparison by different exploration strategies. \protect\subref{fig:rl_train_speed} Training time; \protect\subref{fig:rl_reward_compare} Convergence of cumulative reward}
 \label{fig:rl_exploration}
\end{figure}

In this experiment, we first compare TDGE with both DGE and TGE in term of training time. Figure \ref{fig:rl_train_speed} shows the average training time for each exploration strategy in a network size ranged from 3 to 9 devices. Result shows that TDGE outperforms TGE and DGE in training time by 1.2x and 1.3x less, with a standard deviation of 1.9x and 2.1x less, respectively. In addition, Figure \ref{fig:rl_reward_compare} shows the convergence of cumulative reward in each strategy in a network size of 8 devices. The triangle markers indicate convergence points. We observe that TDGE converges 1.2x and 1.4x faster than TGD and DGE, respectively. As a result, TDGE can accelerate the process of Chain-scheduler training. 

\subsubsection{\textbf{Performance in Re-learning}}\hfill

\begin{figure}
 \centering
    \subfloat[]{
    \includegraphics[width=0.5\linewidth]{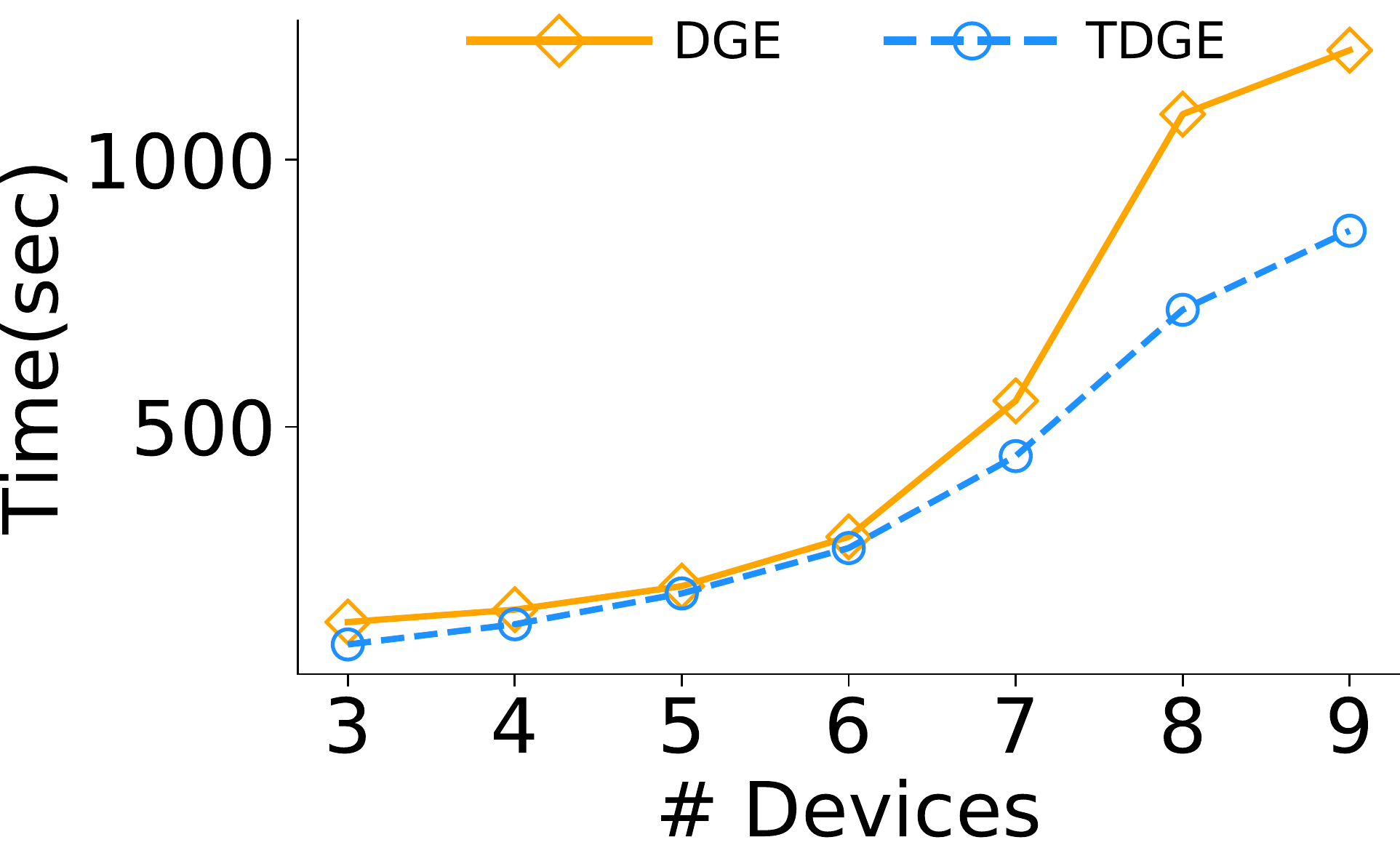}
    \label{fig:rl_retrain_time}
    }
     \subfloat[]{
    \includegraphics[width=0.5\linewidth]{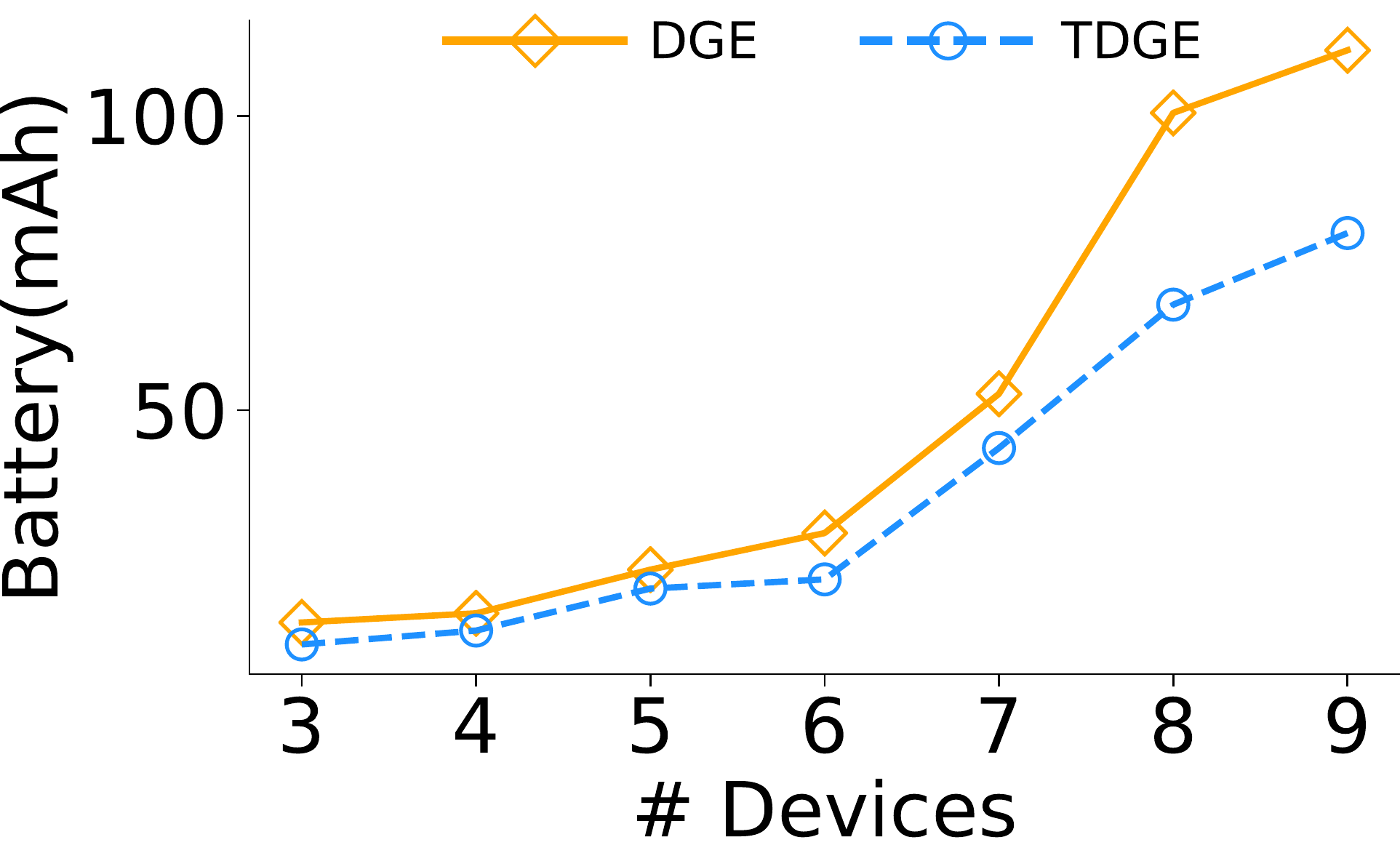}
    \label{fig:rl_retrain_battery}
    }
     \caption{Resources reduction in re-learning conditions. \protect\subref{fig:rl_retrain_time}\:Time and \protect\subref{fig:rl_retrain_battery}\:Battery}
 \label{fig:rl_retrain}
\end{figure}

In this experiment, we evaluate TDGE under a re-learning scenario in terms of training time and battery consumption. Figure \ref{fig:rl_retrain_time} shows that TDGE outperforms DGE with less training time, especially in more-device scenarios (e.g., 1.5x less when the number of devices is larger than 7). Figure \ref{fig:rl_retrain_battery} shows that on average TDGE consumes 1.3x less battery consumption than DGE. Similarly, more battery savings are observed in TDGE with more devices. 

\subsubsection{\textbf{Scheduling Performance}}\hfill

\begin{figure}
 \centering
    \subfloat[]{
    \includegraphics[width=0.5\linewidth]{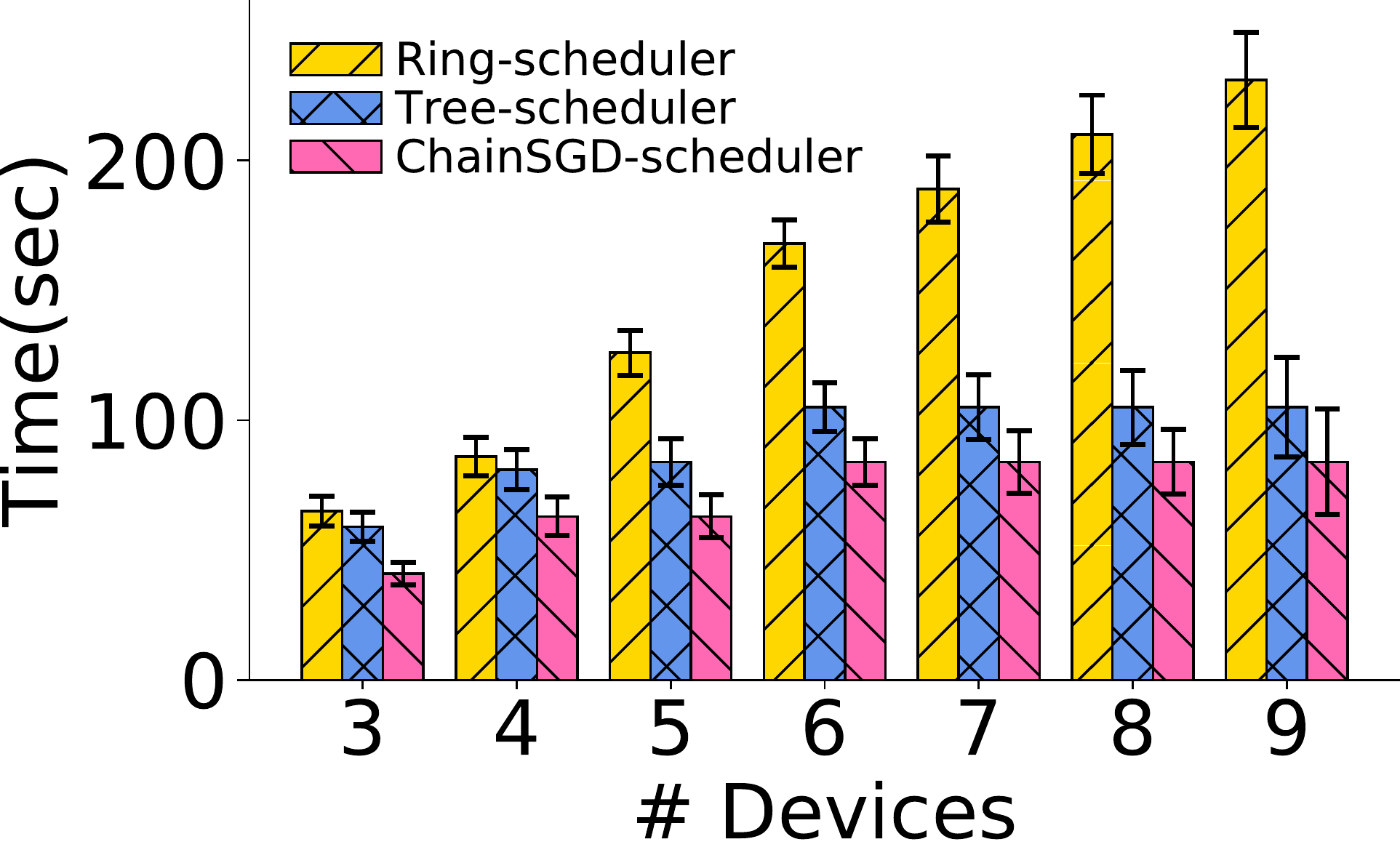}
    \label{fig:rl_tree_compare}
    }
    \subfloat[]{
    \includegraphics[width=0.5\linewidth]{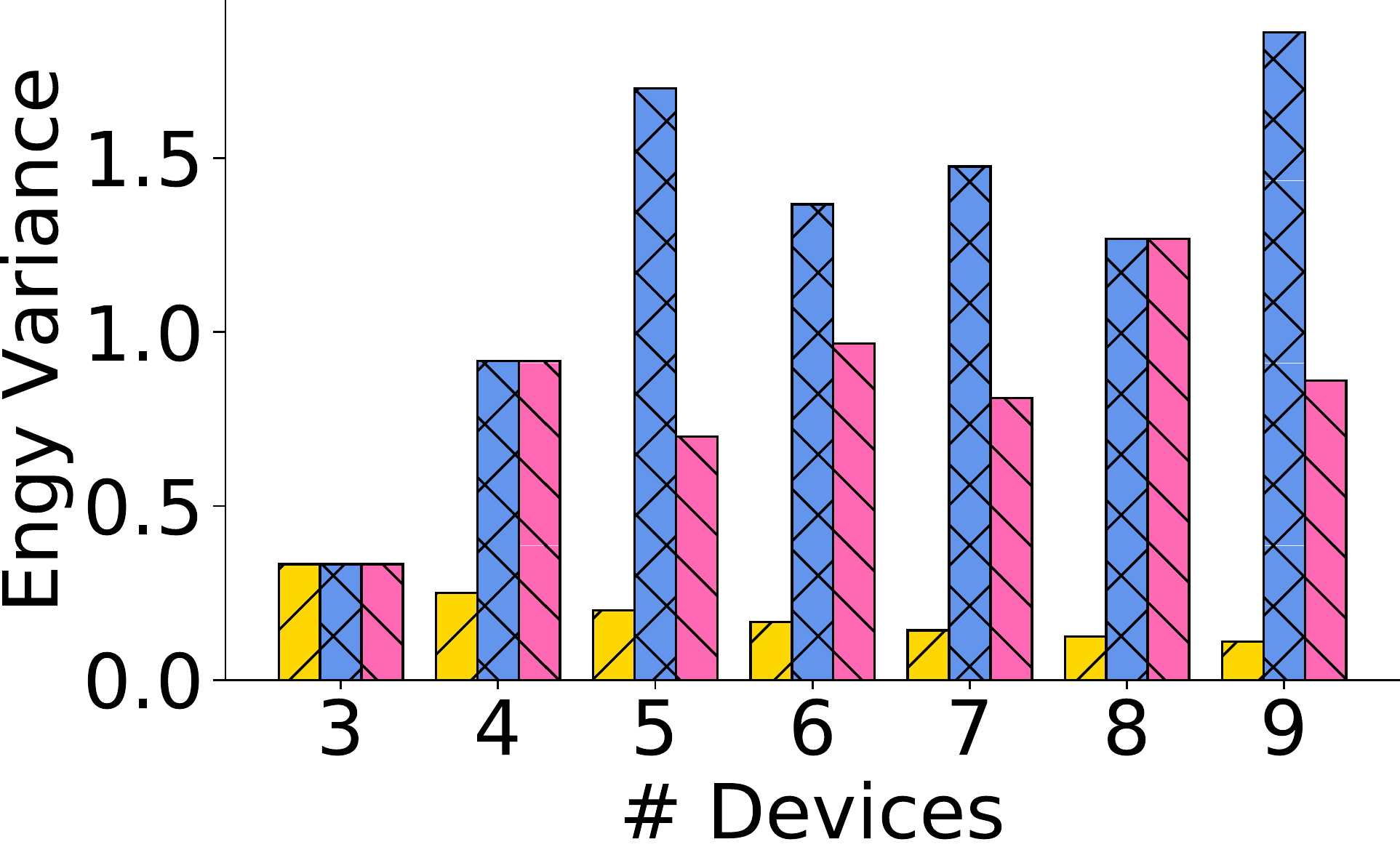}
    \label{fig:rl_ring_compare}
    }
     \caption{Scheduling performance comparison. \protect\subref{fig:rl_tree_compare} Latency;  \protect\subref{fig:rl_ring_compare} Energy balance}
 \label{fig:rl_scheduling}
\end{figure}

In this experiment, we compare Chain-scheduler with Tree-scheduler and Ring-scheduler in terms of training time and energy balance using the HAR dataset in Table \ref{table:dataset}. Figure \ref{fig:rl_tree_compare} 
shows that on average ChainSGD-scheduler outperforms Tree-scheduler and Ring-scheduler by 23\% and 53\% less training time, respectively. In addition, more time savings are observed in ChainSGD-scheduler when the number of devices increases. Figure \ref{fig:rl_ring_compare} shows that Ring-scheduler achieves the minimal energy variance among three, and ChainSGD-scheduler reduces energy variance by 40\% on average compared to Tree-scheduler. This experiment demonstrates that ChainSGD-scheduler achieves the best trade-off between training time and energy variance compared to Tree-scheduler and Ring-scheduler.

\subsection{Performance of MDLdroid} 

To give a comprehensive evaluation for the performance of MDLdroid, we compare MDLdroid with the Federated Learning (FL) \cite{DBLP:journals/corr/abs-1902-01046} from training accuracy and resource used perspectives. Besides, we choose a server-based approach to further ensure the training accuracy to be reliable. we next explore the optimized resource-accuracy trade-off options and limitations of MDLdroid.

\subsubsection{\textbf{Experimental Setup}}\hfill

For performance benchmarking, we use FL as our baseline, which is implemented based on a master-slave structure, i.e., the same system architecture, 
to keep training computational and communication costs identical for fair comparison. In addition, we also compare the training accuracy of MDLdroid vs. the server-based approach by running training with the same model configurations on a desktop computer using all the 6 datasets. To measure battery consumption, we use Java refection to access the system instance of \textit{BatteryStatsImpl} after rooting the smartphones. To monitor real-time CPUs and memory consumption on smartphone, we use Android Debug Bridge (ADB) commands.

\subsubsection{\textbf{MDLdroid vs. FL}} \hfill

We compare the overall performance of MDLdroid vs. FL from three resource perspectives---peak-memory overhead, training time, and network energy balance.

\begin{figure}
 \centering
     \subfloat[]{
    \includegraphics[trim=0 0 25 5, clip,width=0.5\columnwidth]{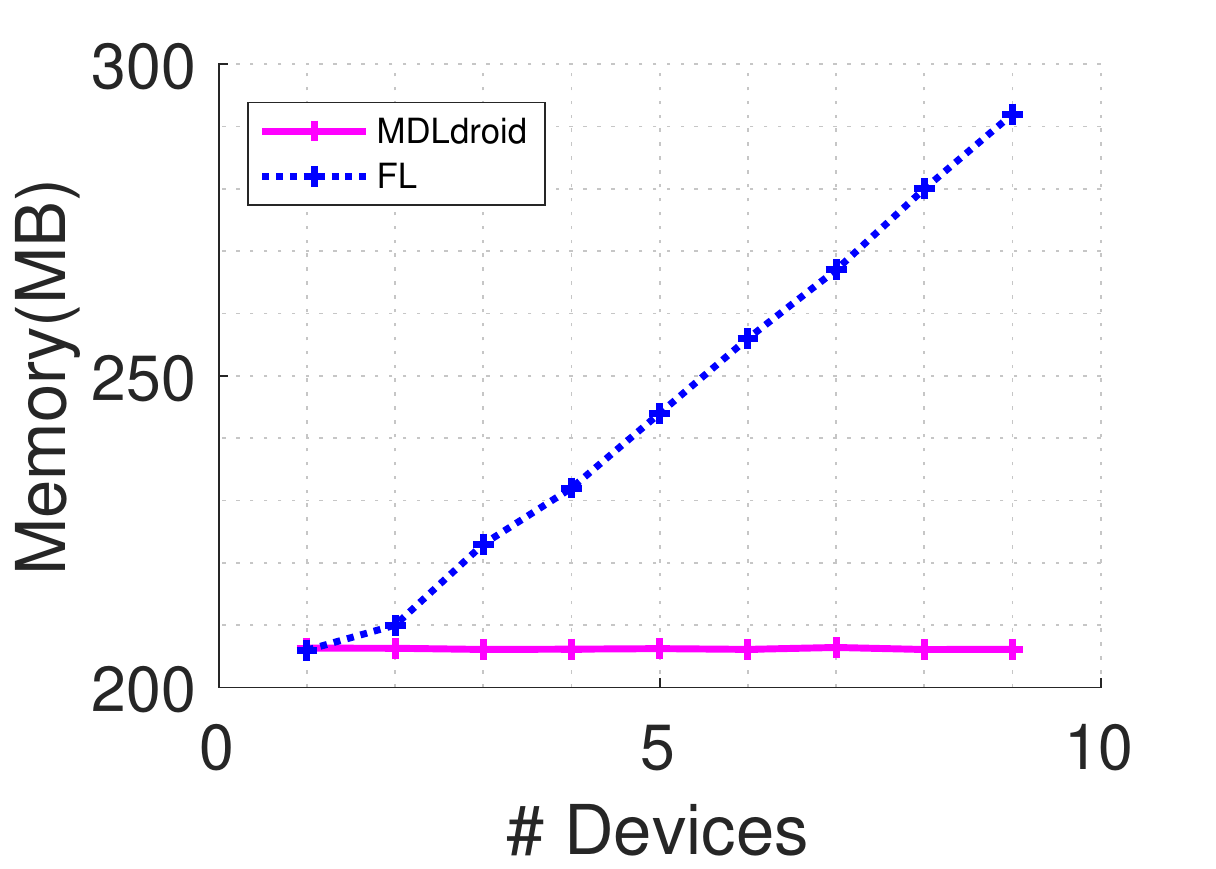}
    \label{fig:master_chain_memory}
    }
    \subfloat[]{
    \includegraphics[trim=0 0 20 5, clip,width=0.5\columnwidth]{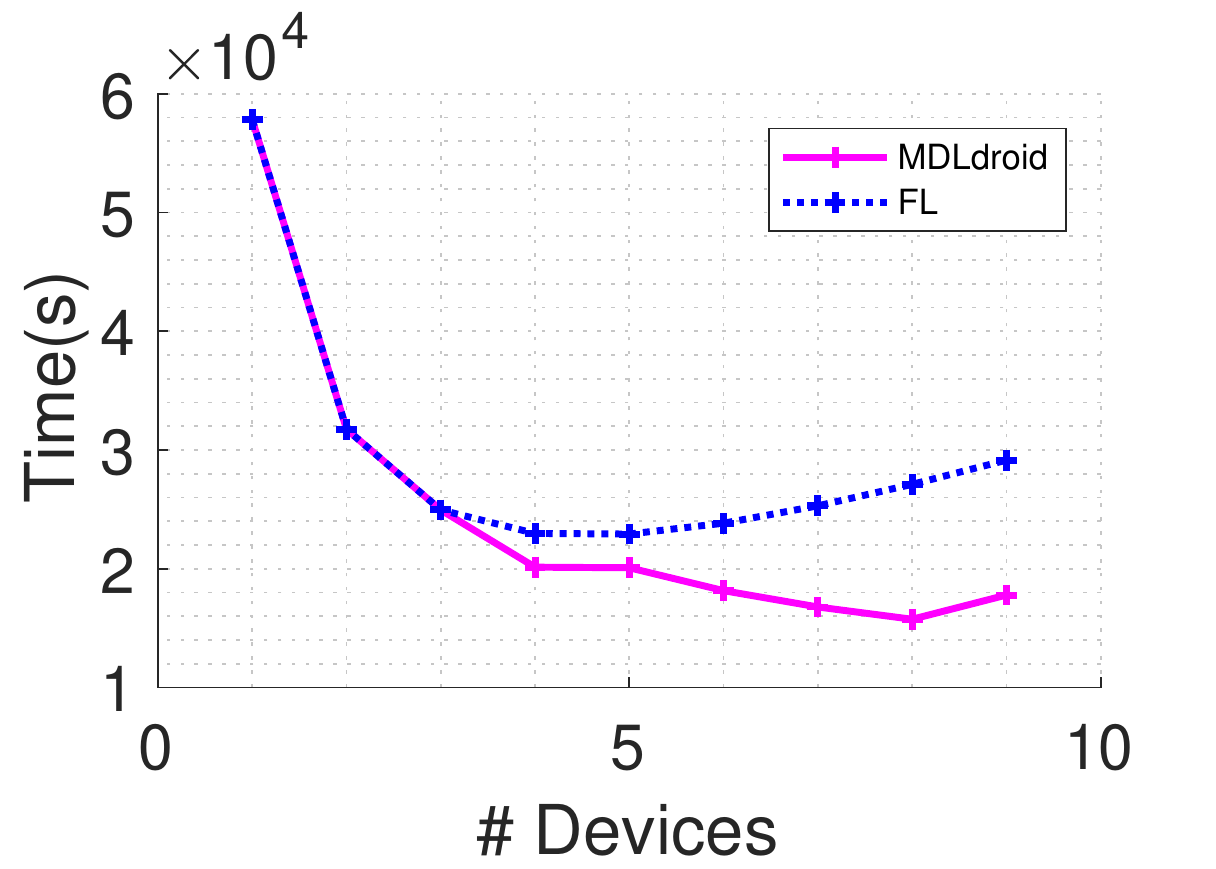}
    \label{fig:master_chain_time}
    } \quad
    \subfloat[]{
    \includegraphics[trim=0 0 25 5, clip,width=0.5\columnwidth]{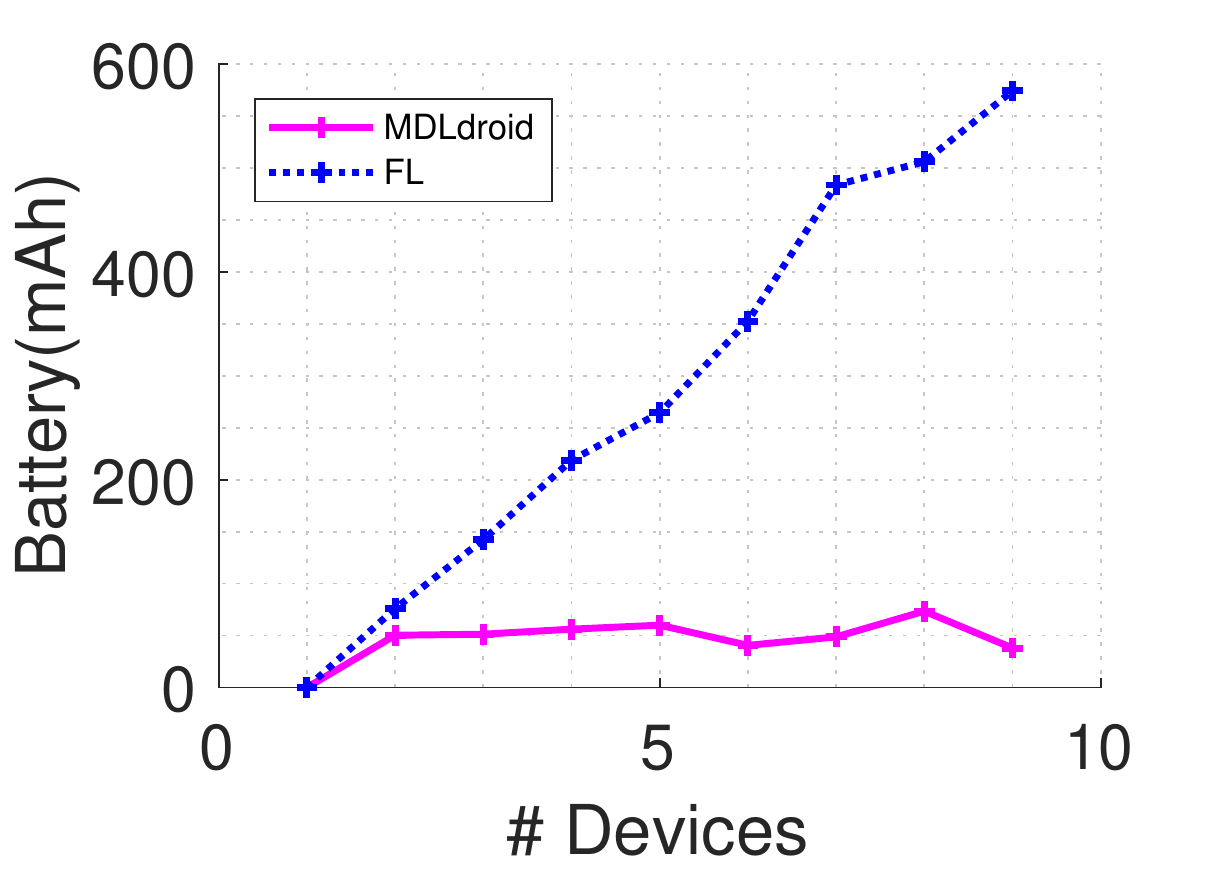}
    \label{fig:master_chain_battery}
    }
    \subfloat[]{
    \includegraphics[trim=0 0 25 5, clip,width=0.5\columnwidth]{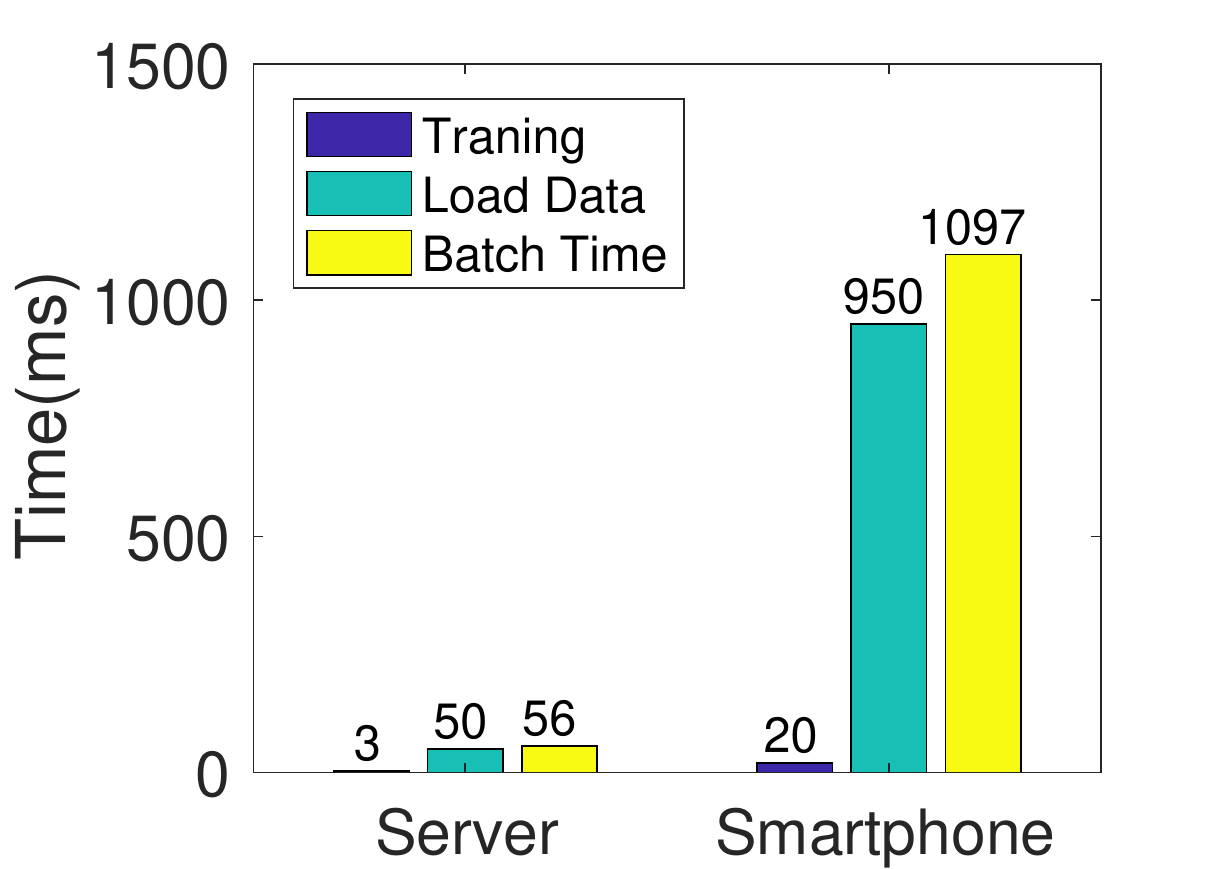}
    \label{fig:read_speed}
    }
     \caption{MDLdroid vs. FL performance comparison. \protect\subref{fig:master_chain_memory} Memory; \protect\subref{fig:master_chain_time} Time; \protect\subref{fig:master_chain_battery} Battery consumption; \protect\subref{fig:read_speed} One batch training time breakdown comparison}
 \label{fig:intro_memory}
\end{figure}

\textbf{Peak-Memory}
Figure \ref{fig:master_chain_memory} plots the peak-memory value for each approach with a network size ranged from 1 to 9 using the PAMA2 dataset based on LeNet. Result shows that the peak-memory overhead of the master device in FL increases linearly with the number of devices, while MDLdroid remains stable at low overhead due to 
only a pair of devices is involved during each model aggregation task. 

\textbf{Training Time}
Figure \ref{fig:master_chain_time} shows that the training time in MDLdroid is effectively saved by 1.5x on average compared to FL. The training time in MDLdroid decreases with the network size increased. While the training time in FL presents a U-shaped curve as the communication time for model aggregations gradually increases.
This is due to the efficient design of both model aggregation scheduling (\S\ref{section:chain-balance}) and resource-aware broadcasts (\S\ref{section: broadcast}) in MDLdroid.

\textbf{Network Energy Balance}
Figure \ref{fig:master_chain_battery} shows that the energy consumption of each device for model aggregation in MDLdriod is much less than that of the master device in FL, especially in more device scenarios.
The result shows a training device in MDLdriod achieves 5.8x energy consumption reduction for communication on average compared to the master device in FL. 
In addition, another key observation is that MDLdroid achieves better energy balance among devices as MDLdroid evenly distribute model aggregation tasks to each device during scheduling. 
Since MDLdroid requires a training device to periodically report its resource condition via a tiny BLE message, the actual battery consumption to send model parameters is much smaller than that using BS, e.g., fully training PAMA2 by 20 \textit{Epoch} on a single device with roughly 15 mAh out of 3600 mAh cost.

\subsubsection{\textbf{Trade-off between Resource and Accuracy}}\hfill

Both model aggregation frequency and training iteration \textit{Epoch} can significantly impact the balance between resource and accuracy.
In this experiment, we evaluate the trade-off between resource and accuracy in three aspects---training accuracy by a given threshold, battery consumption by maximum battery, and training time. 
In addition, we design 5 trade-off parameters, i.e., 1-E20, 2-E20, 10-E20, 1-E30, and 1-E40, where each of them represents the number of model aggregation rounds per \textit{Epoch}---the number of training iteration \textit{Epoch}.

\begin{figure*}
 \centering
    \subfloat[HAR]{
    \includegraphics[width=0.16\linewidth]{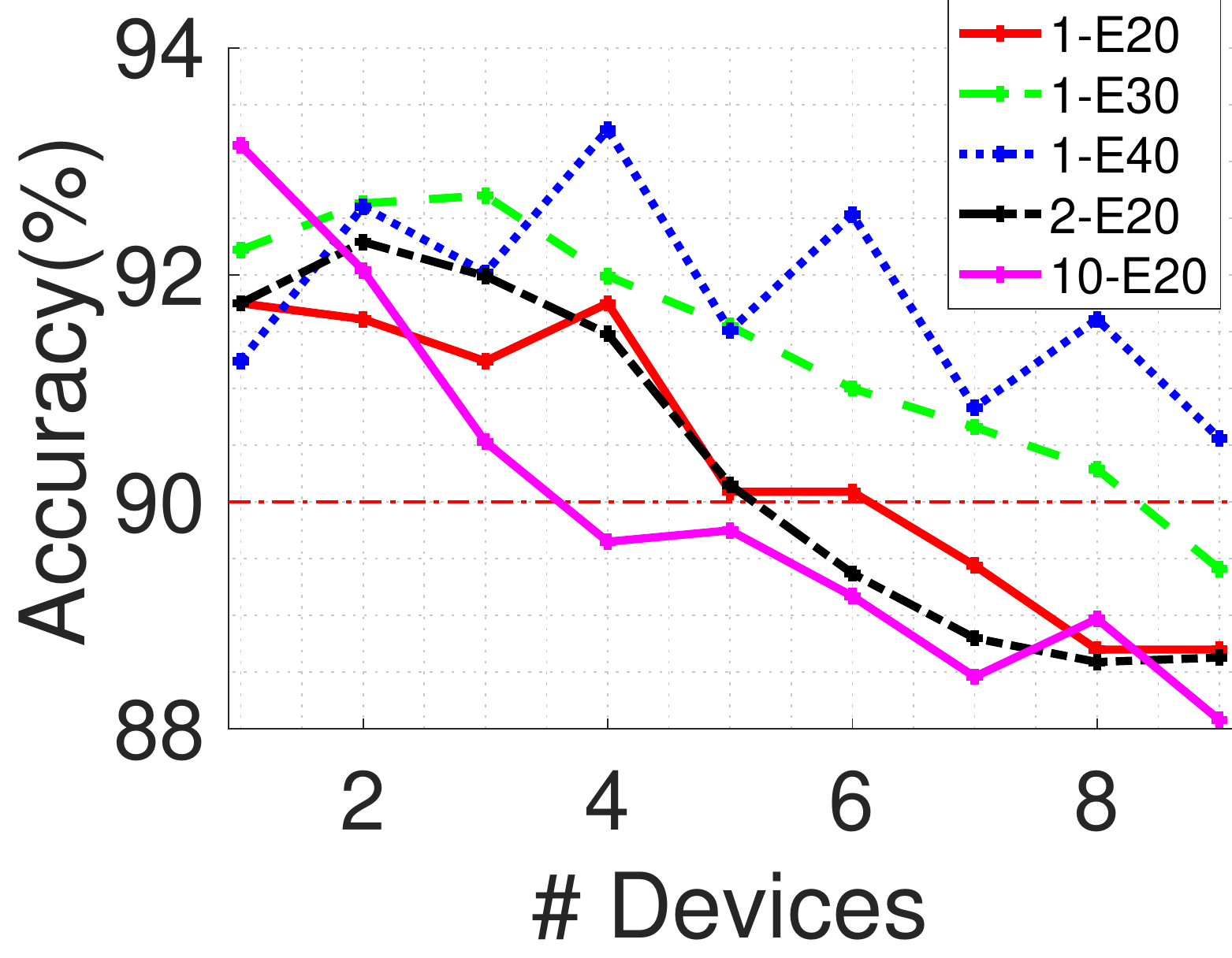}
    \label{fig:tradeoff_ac_har}
    }
    \subfloat[sEMG]{
    \includegraphics[width=0.16\linewidth]{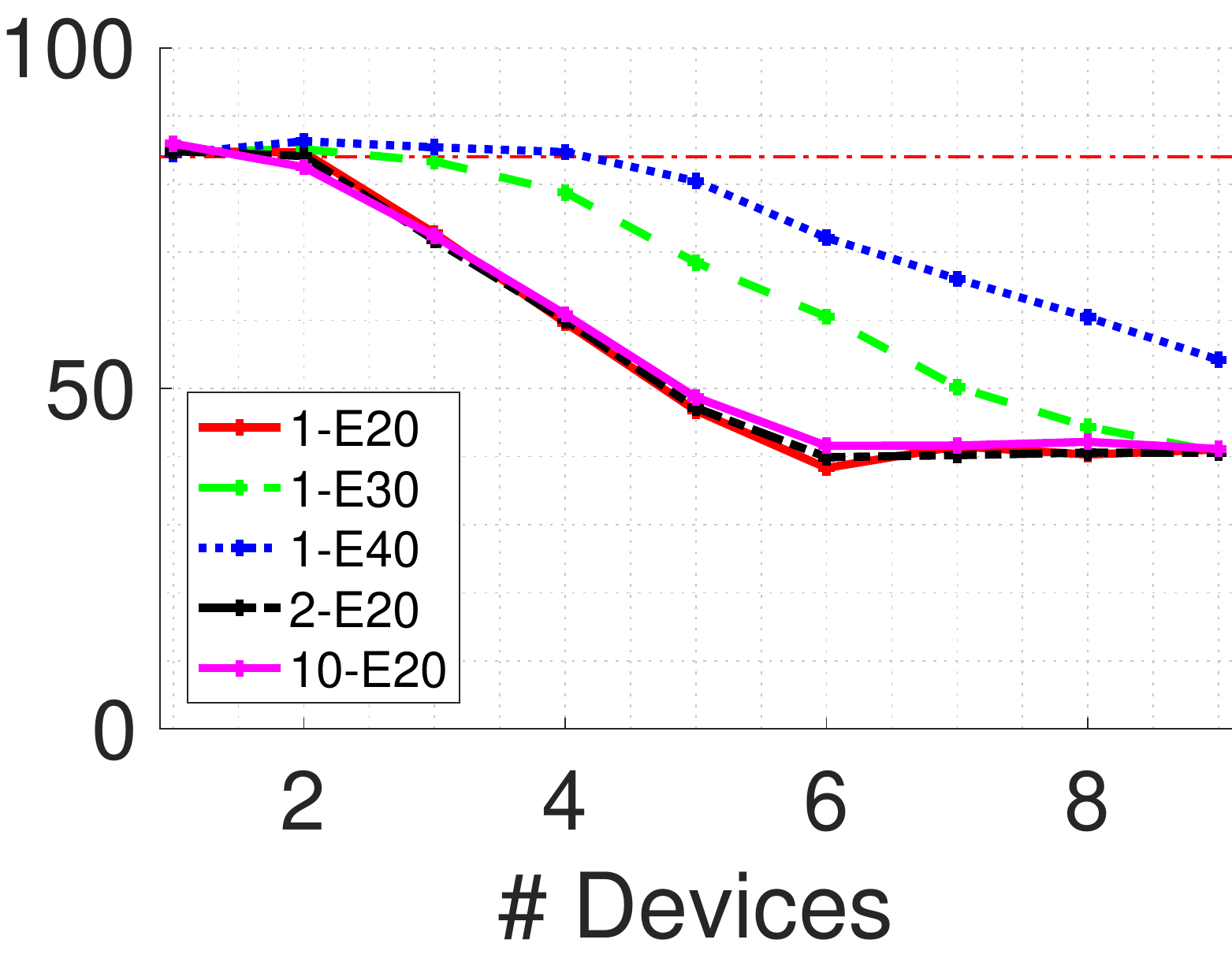}
    \label{fig:tradeoff_ac_emg}
    } 
      \subfloat[UniMiB]{
    \includegraphics[width=0.16\linewidth]{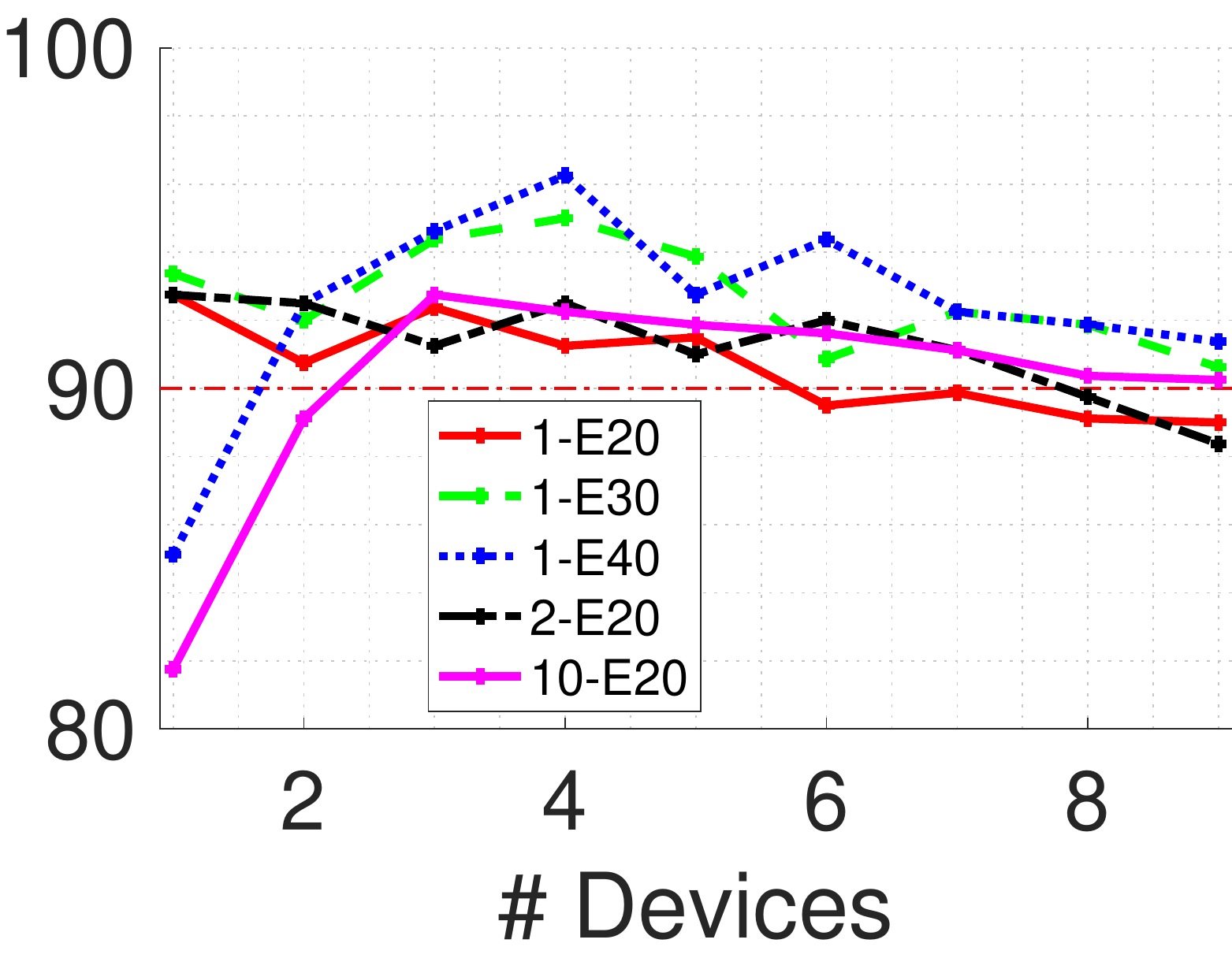}
    \label{fig:tradeoff_ac_fall}
    } 
      \subfloat[PAMAP2]{
    \includegraphics[width=0.16\linewidth]{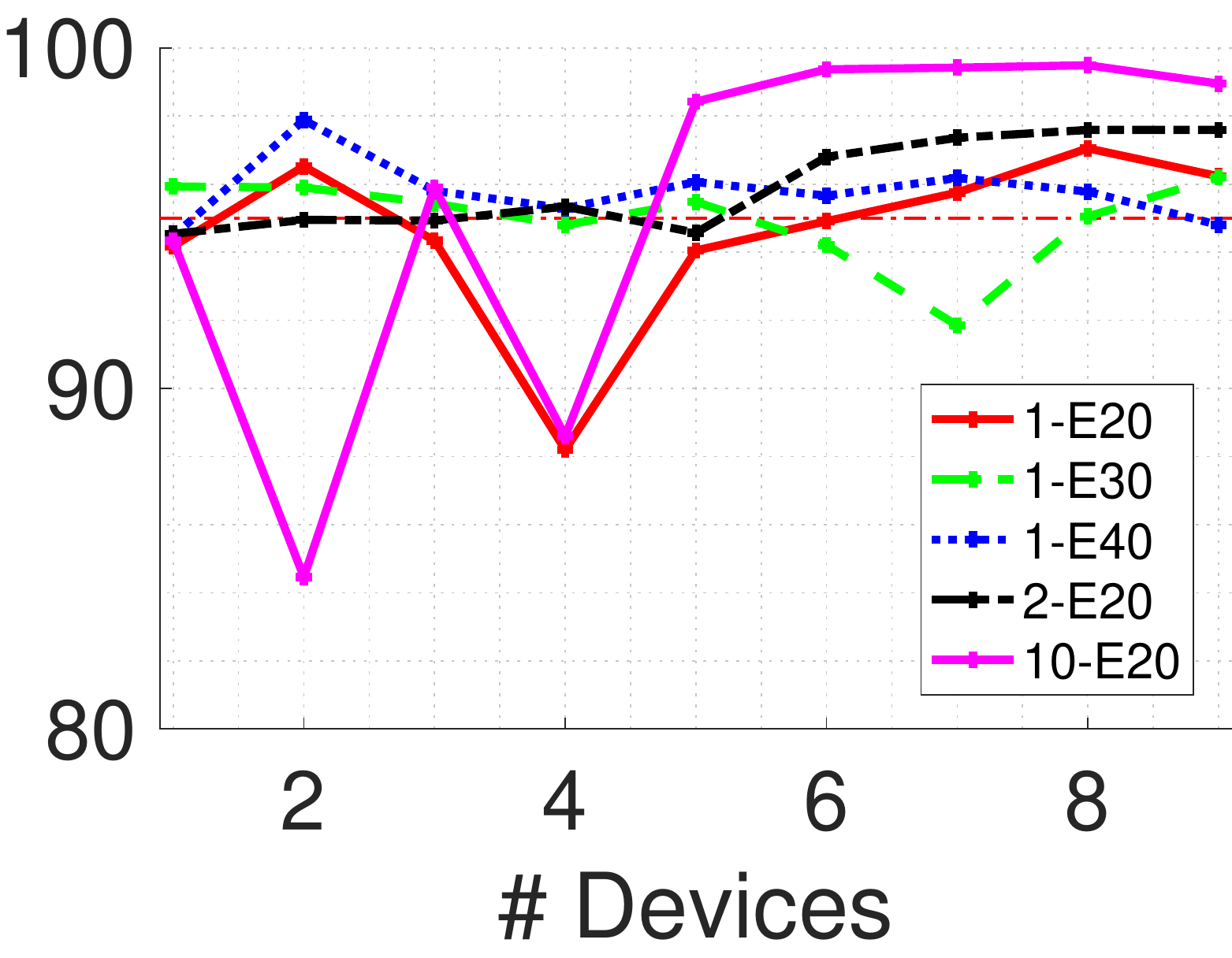}
    \label{fig:tradeoff_ac_pama}
    } 
      \subfloat[MHEALTH]{
    \includegraphics[width=0.16\linewidth]{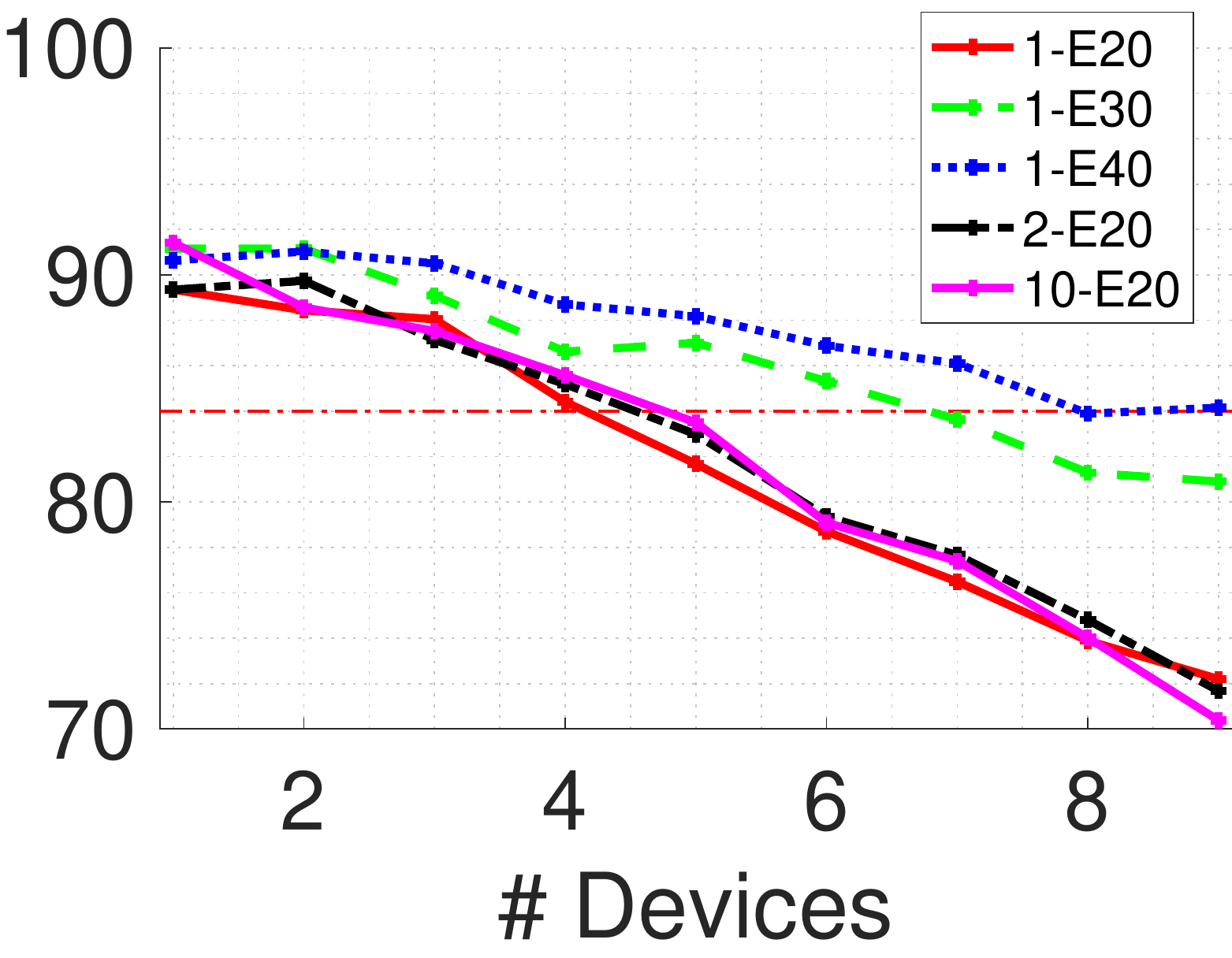}
    \label{fig:tradeoff_ac_mhe}
    } 
      \subfloat[OPPORTUNITY]{
    \includegraphics[width=0.16\linewidth]{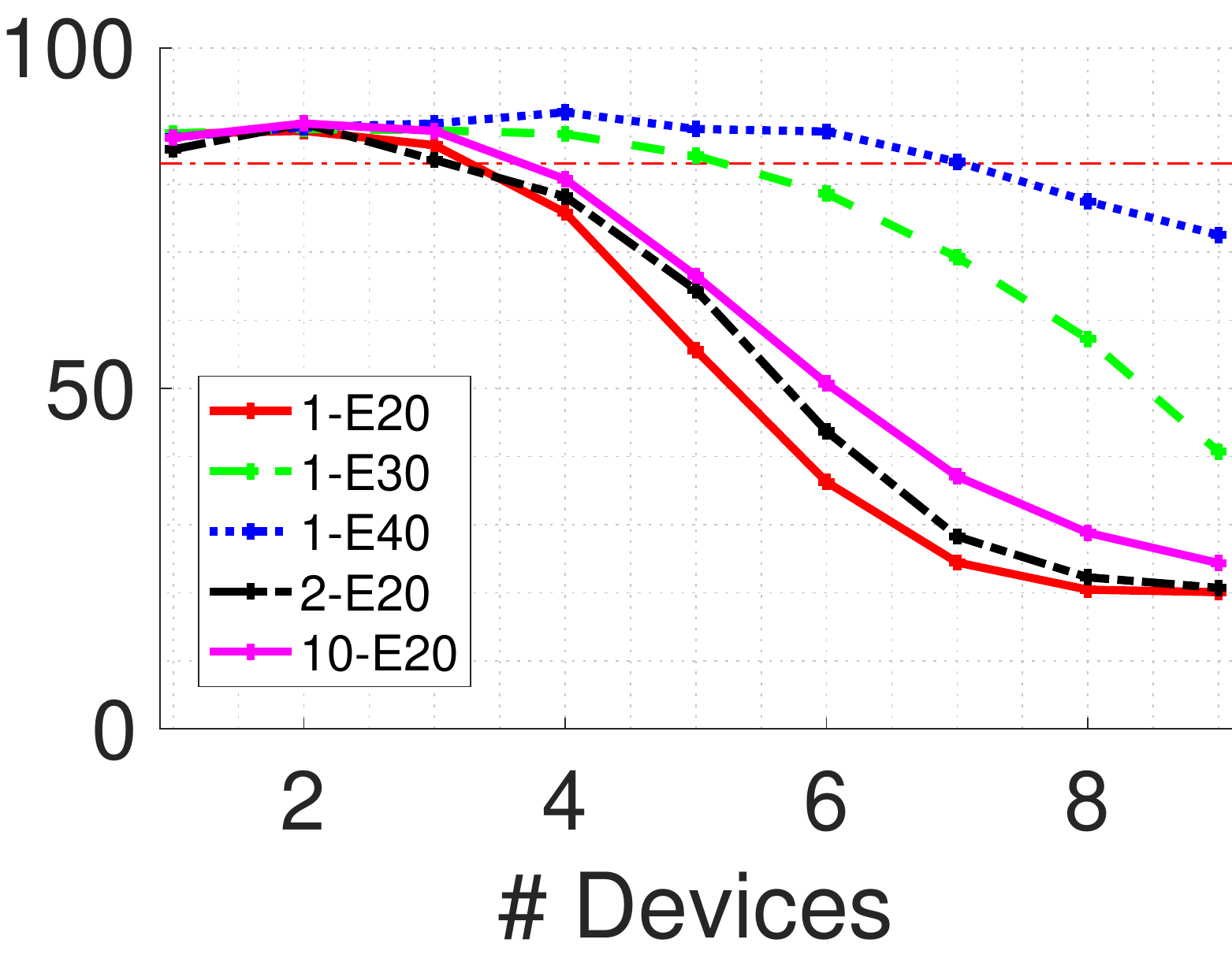}
    \label{fig:tradeoff_ac_op}
    } 
     \caption{Training accuracy by trade-off options}
 \label{fig:tradeoff_ac}
\end{figure*}

\begin{figure*}
 \centering
    \subfloat[HAR]{
    \includegraphics[width=0.16\linewidth]{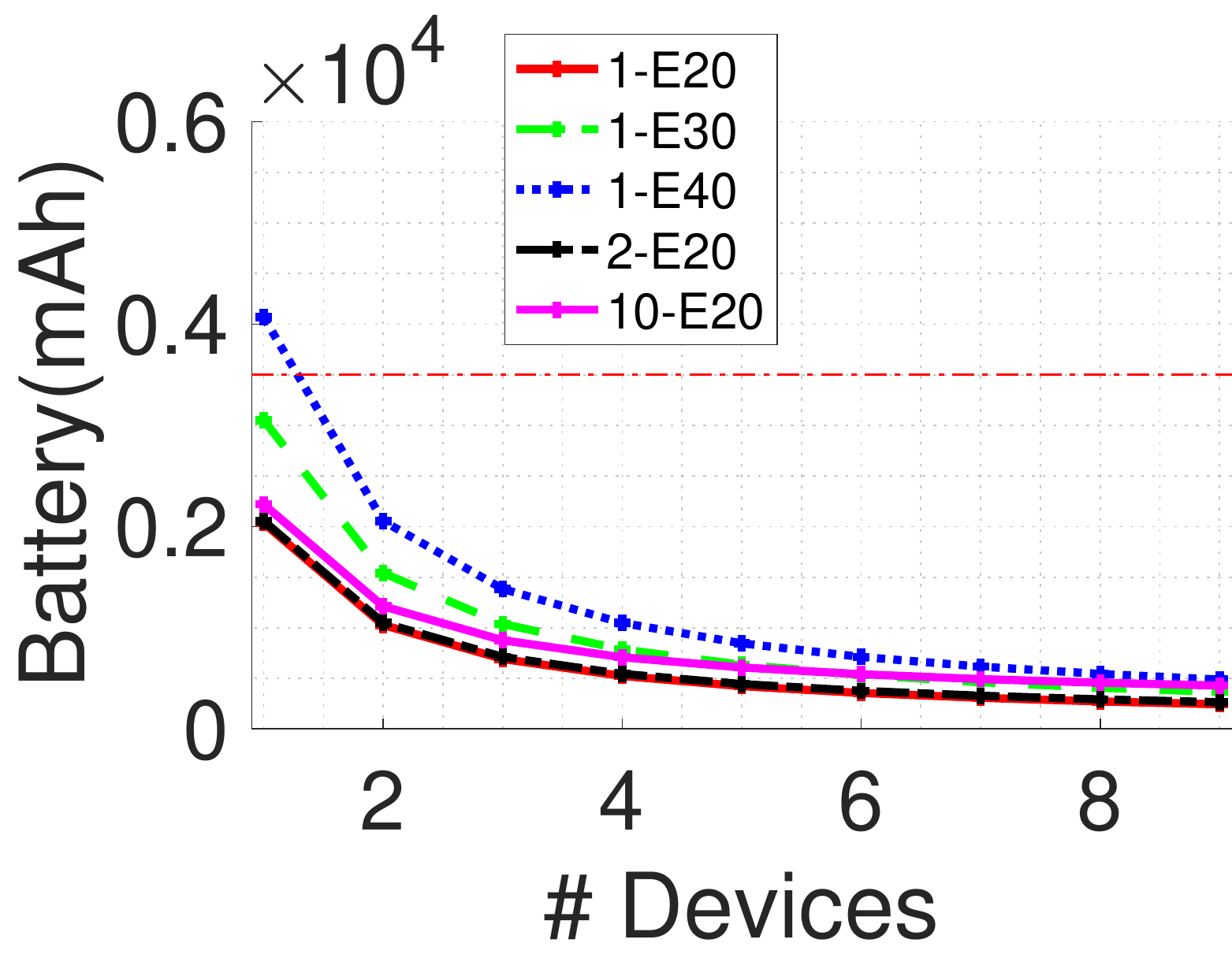}
    \label{fig:tradeoff_battery_har}
    }
    \subfloat[sEMG]{
    \includegraphics[width=0.16\linewidth]{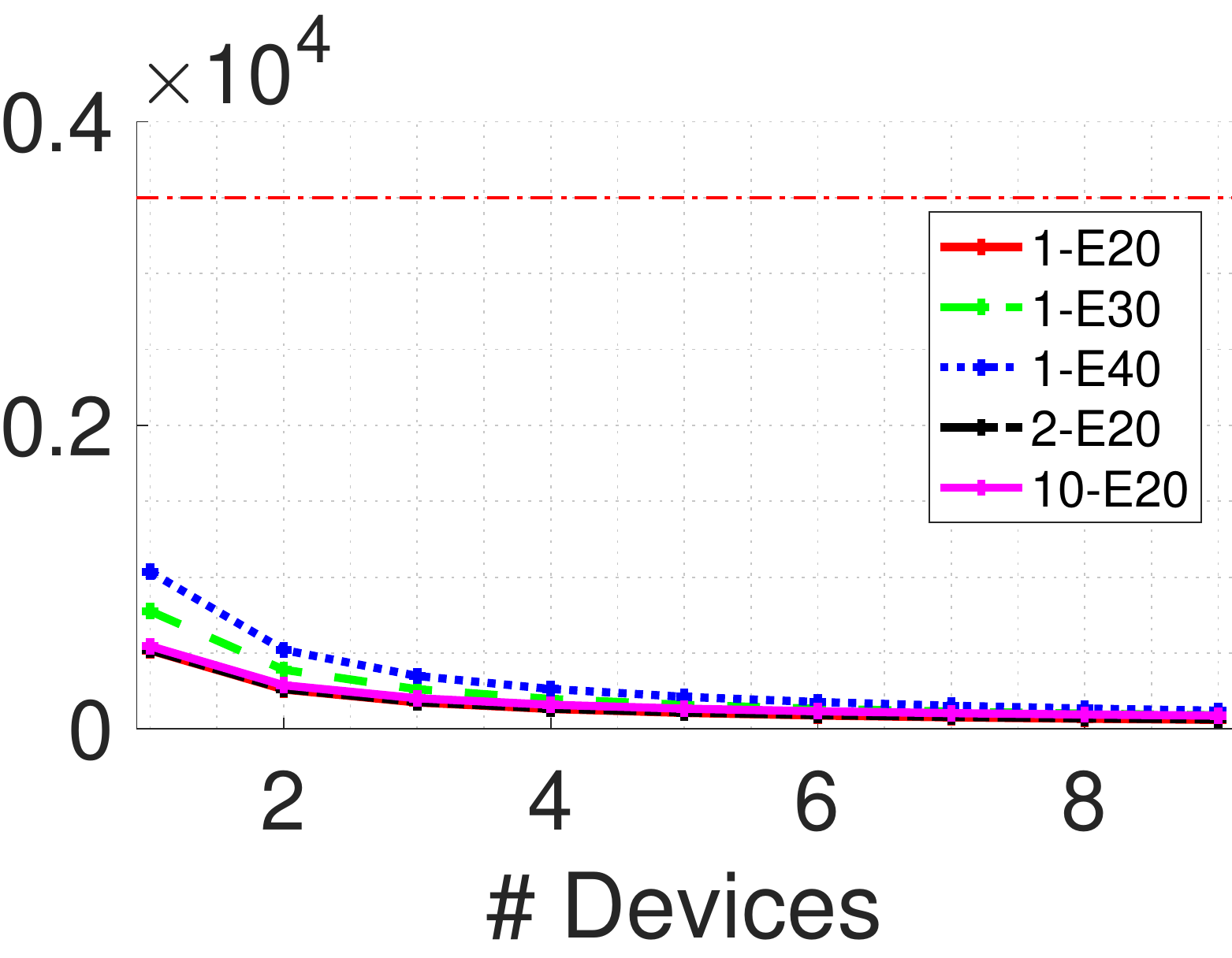}
    \label{fig:tradeoff_battery_emg}
    } 
      \subfloat[UniMiB]{
    \includegraphics[width=0.16\linewidth]{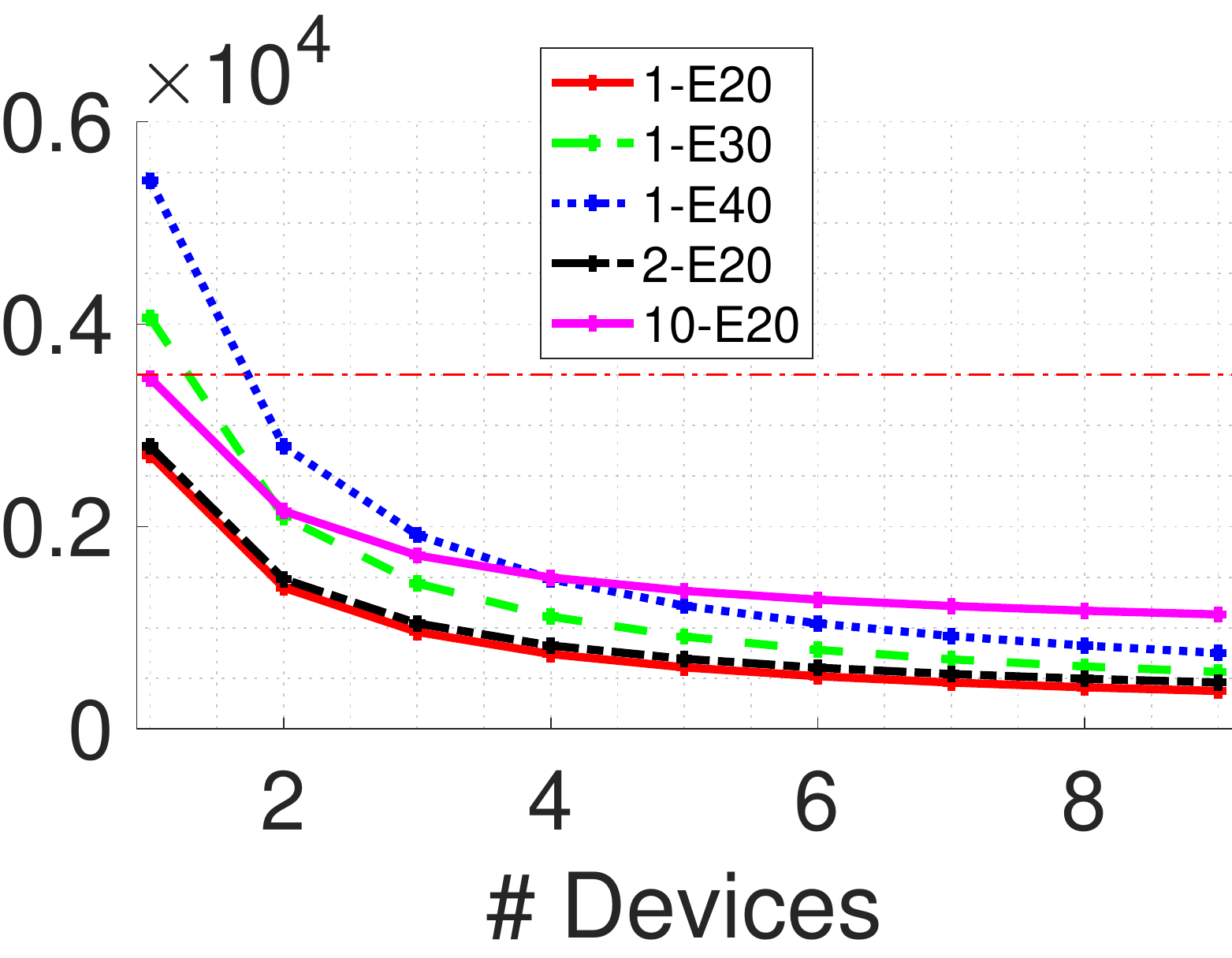}
    \label{fig:tradeoff_battery_fall}
    } 
      \subfloat[PAMAP2]{
    \includegraphics[width=0.16\linewidth]{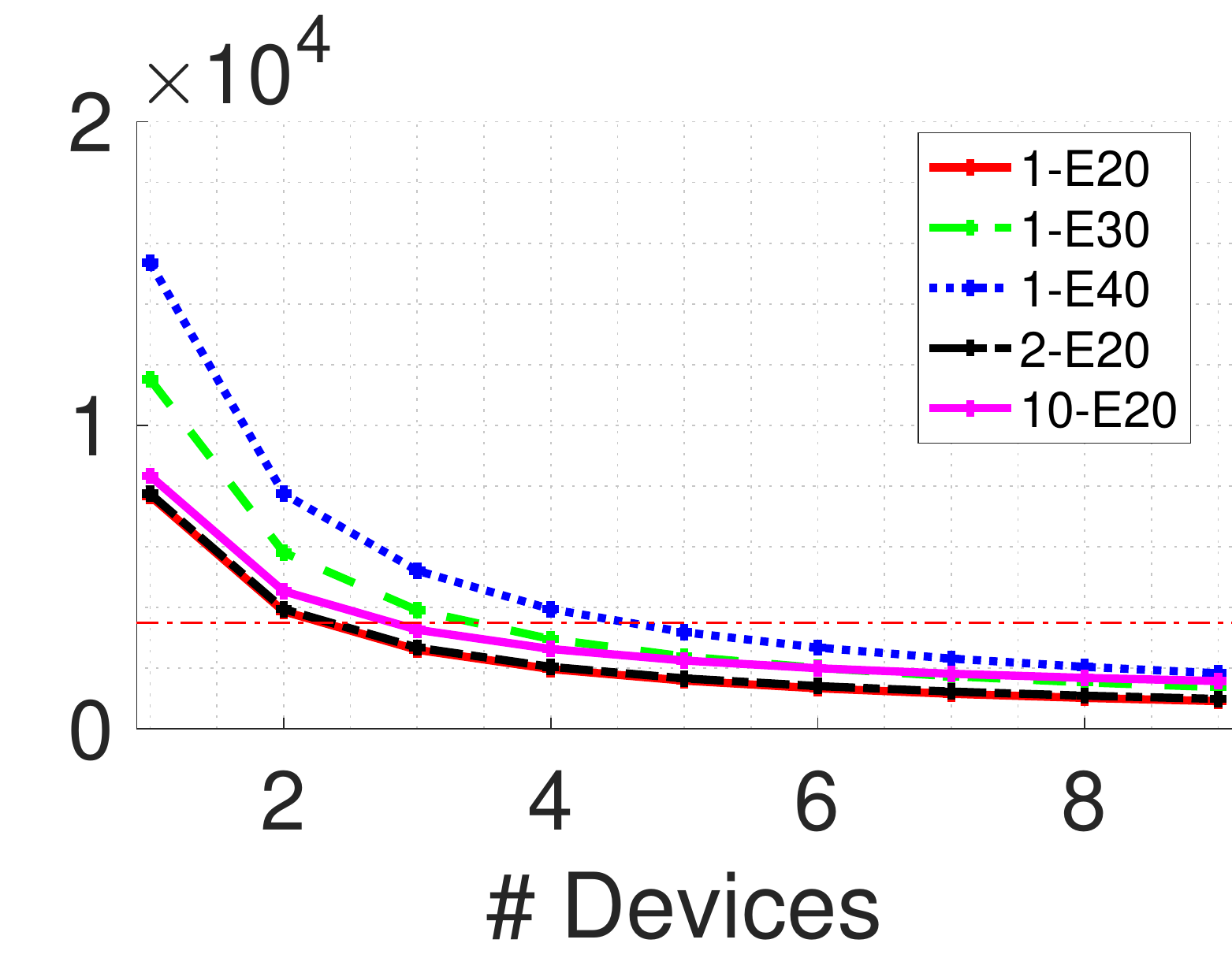}
    \label{fig:tradeoff_battery_pama}
    } 
      \subfloat[MHEALTH]{
    \includegraphics[width=0.16\linewidth]{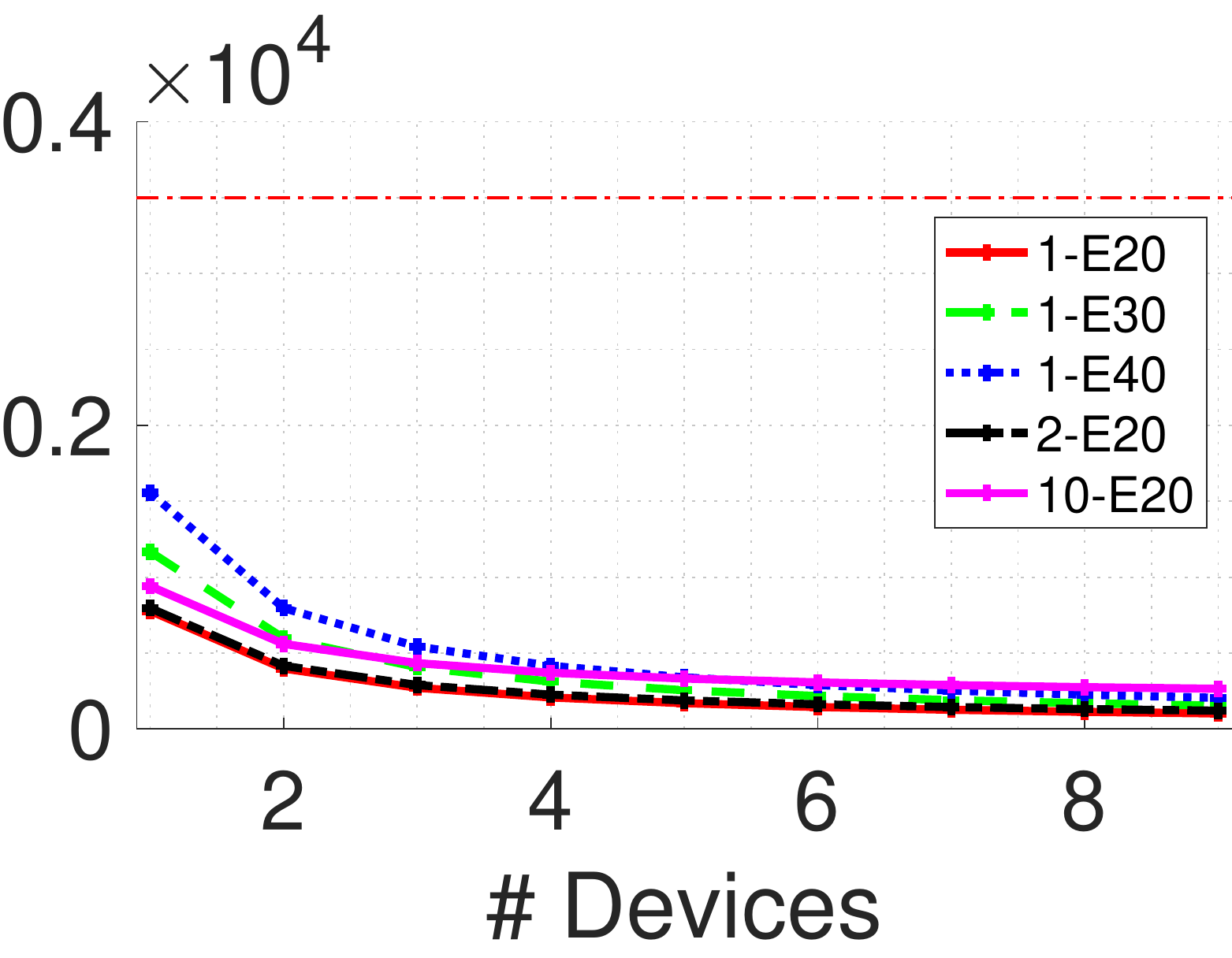}
    \label{fig:tradeoff_battery_mhe}
    } 
      \subfloat[OPPORTUNITY]{
    \includegraphics[width=0.16\linewidth]{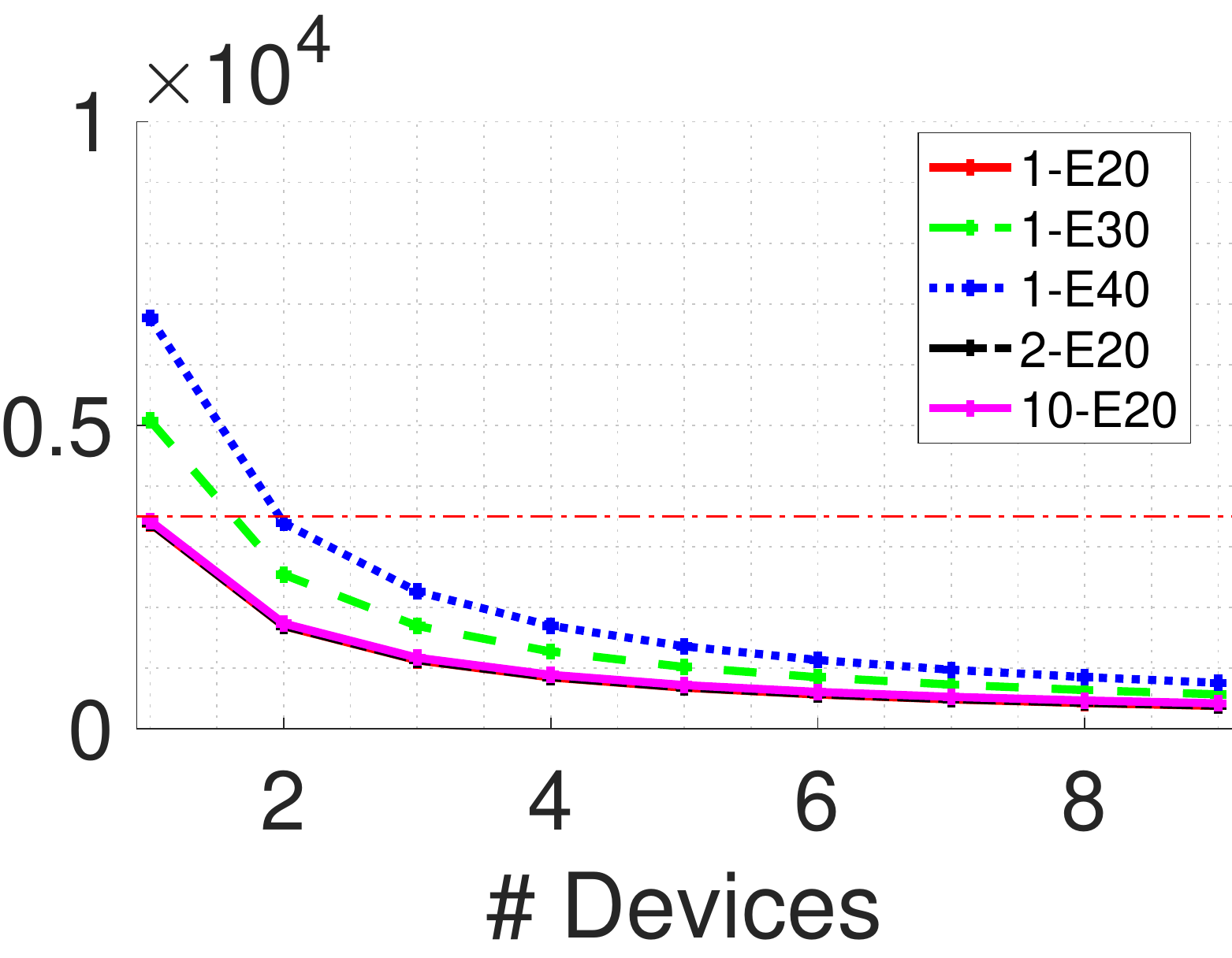}
    \label{fig:tradeoff_battery_op}
    } 
     \caption{Training battery consumption by trade-off options}
 \label{fig:tradeoff_battery}
\end{figure*}

\begin{figure*}
 \centering
    \subfloat[HAR]{
    \includegraphics[width=0.16\linewidth]{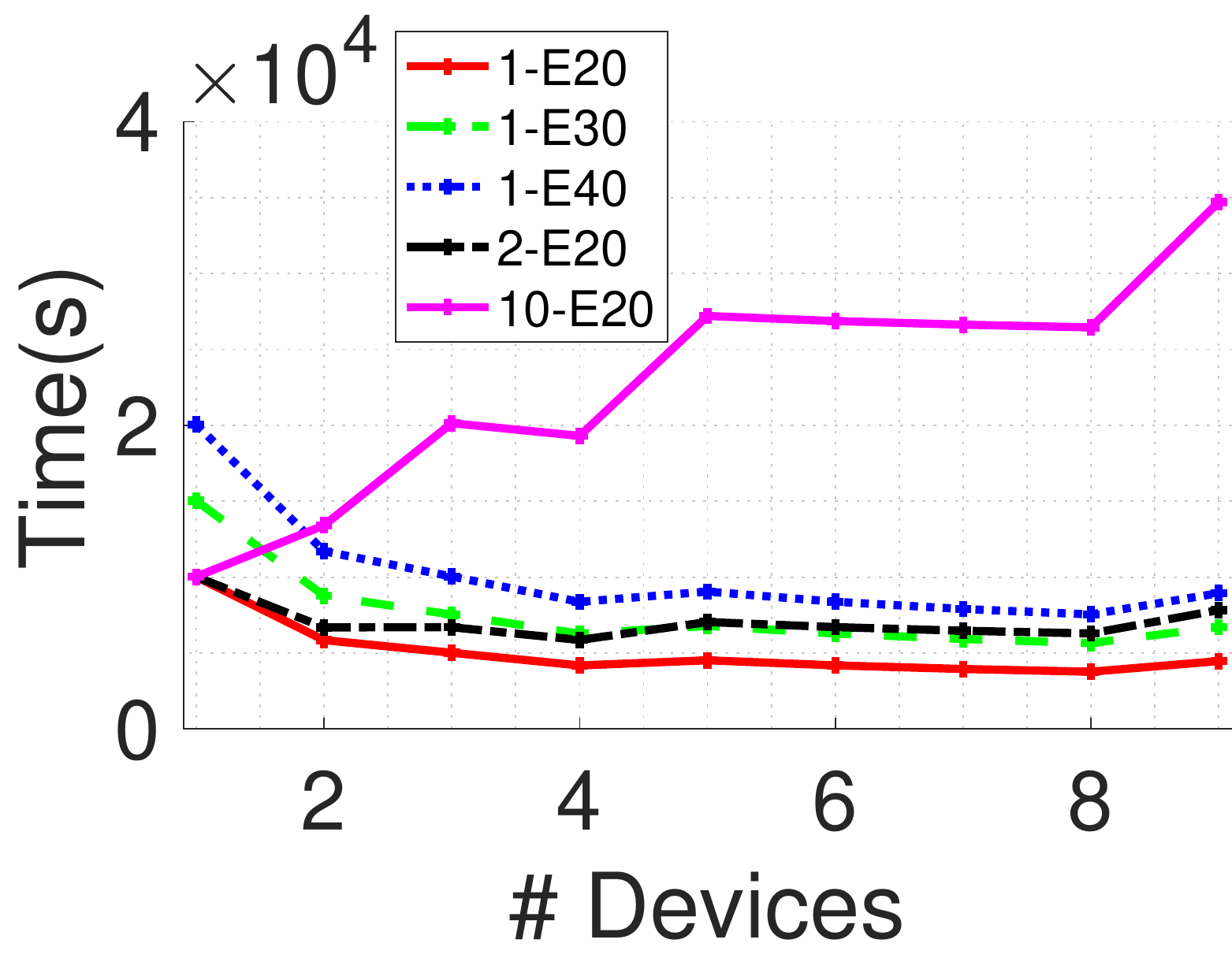}
    \label{fig:tradeoff_time_har}
    }
    \subfloat[sEMG]{
    \includegraphics[width=0.16\linewidth]{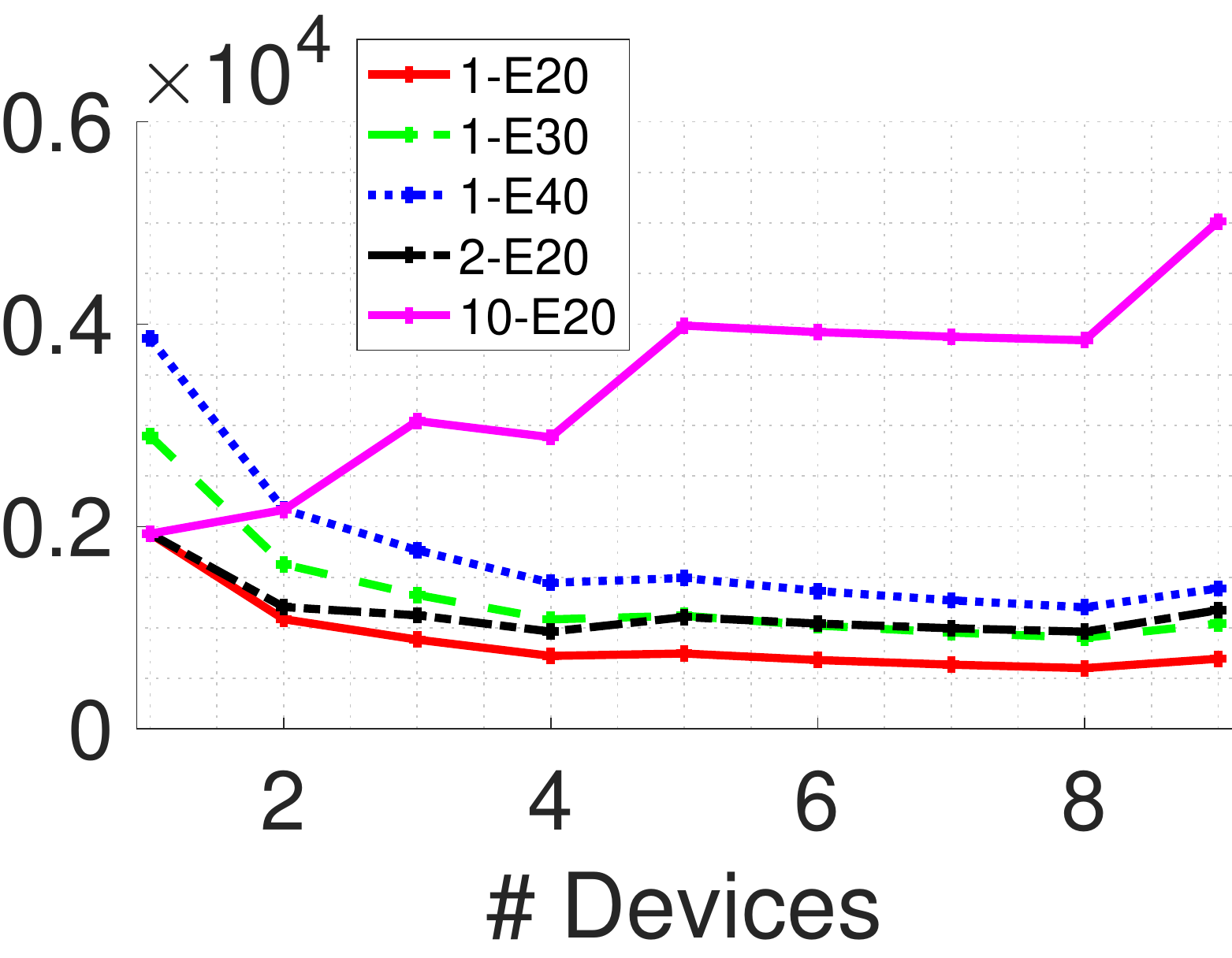}
    \label{fig:tradeoff_time_emg}
    } 
      \subfloat[UniMiB]{
    \includegraphics[width=0.16\linewidth]{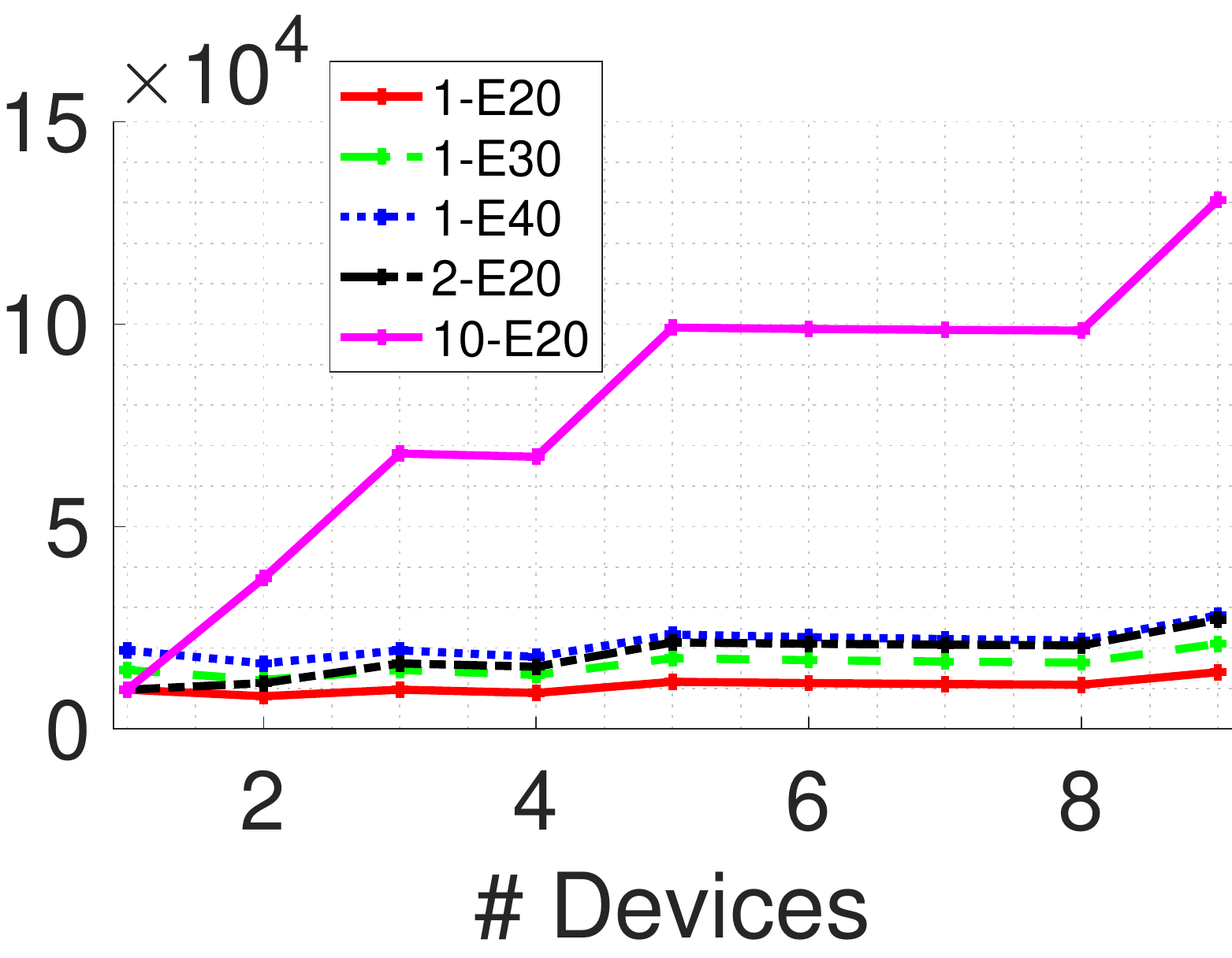}
    \label{fig:tradeoff_time_fall}
    } 
      \subfloat[PAMAP2]{
    \includegraphics[width=0.16\linewidth]{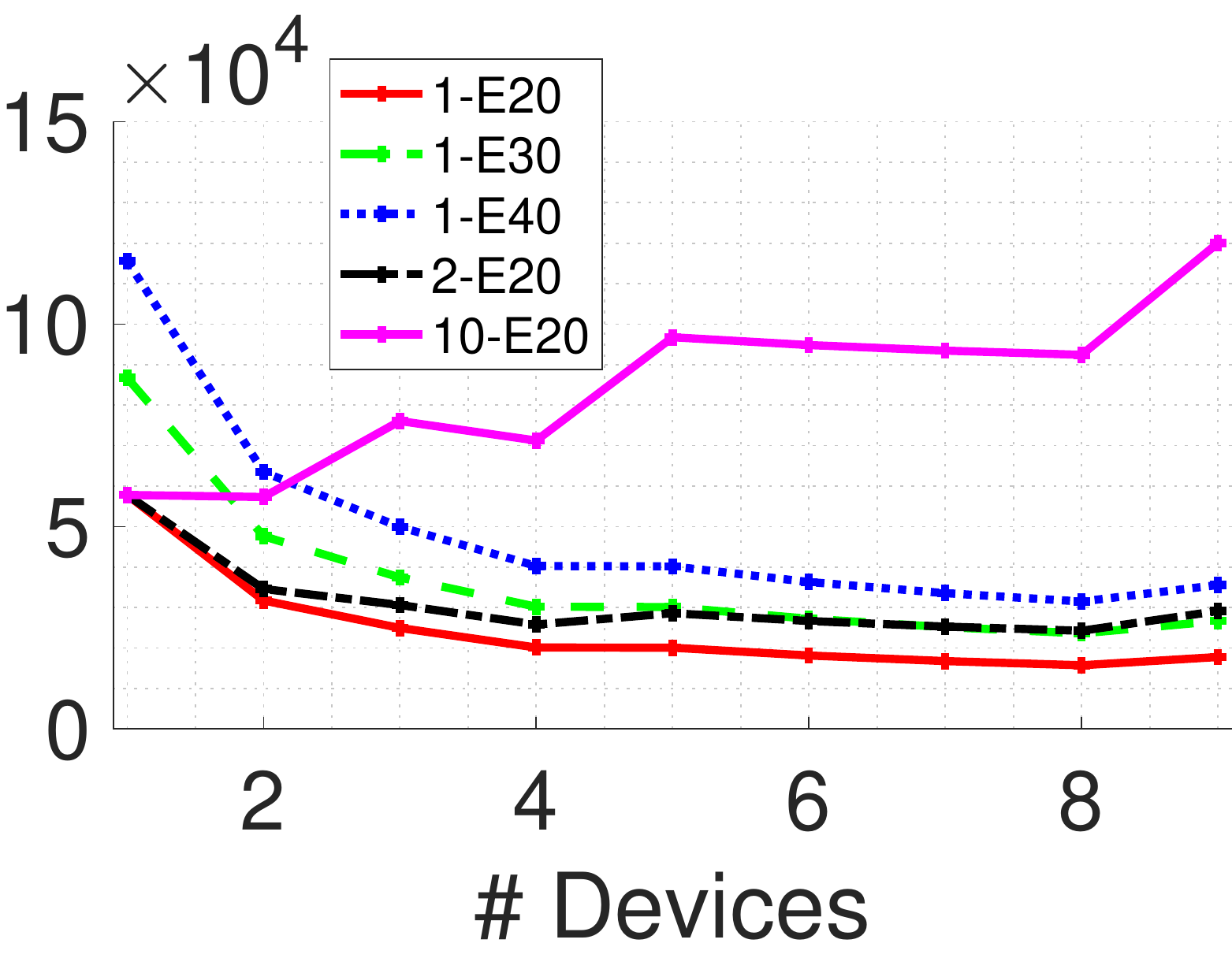}
    \label{fig:tradeoff_time_pama}
    } 
      \subfloat[MHEALTH]{
    \includegraphics[width=0.16\linewidth]{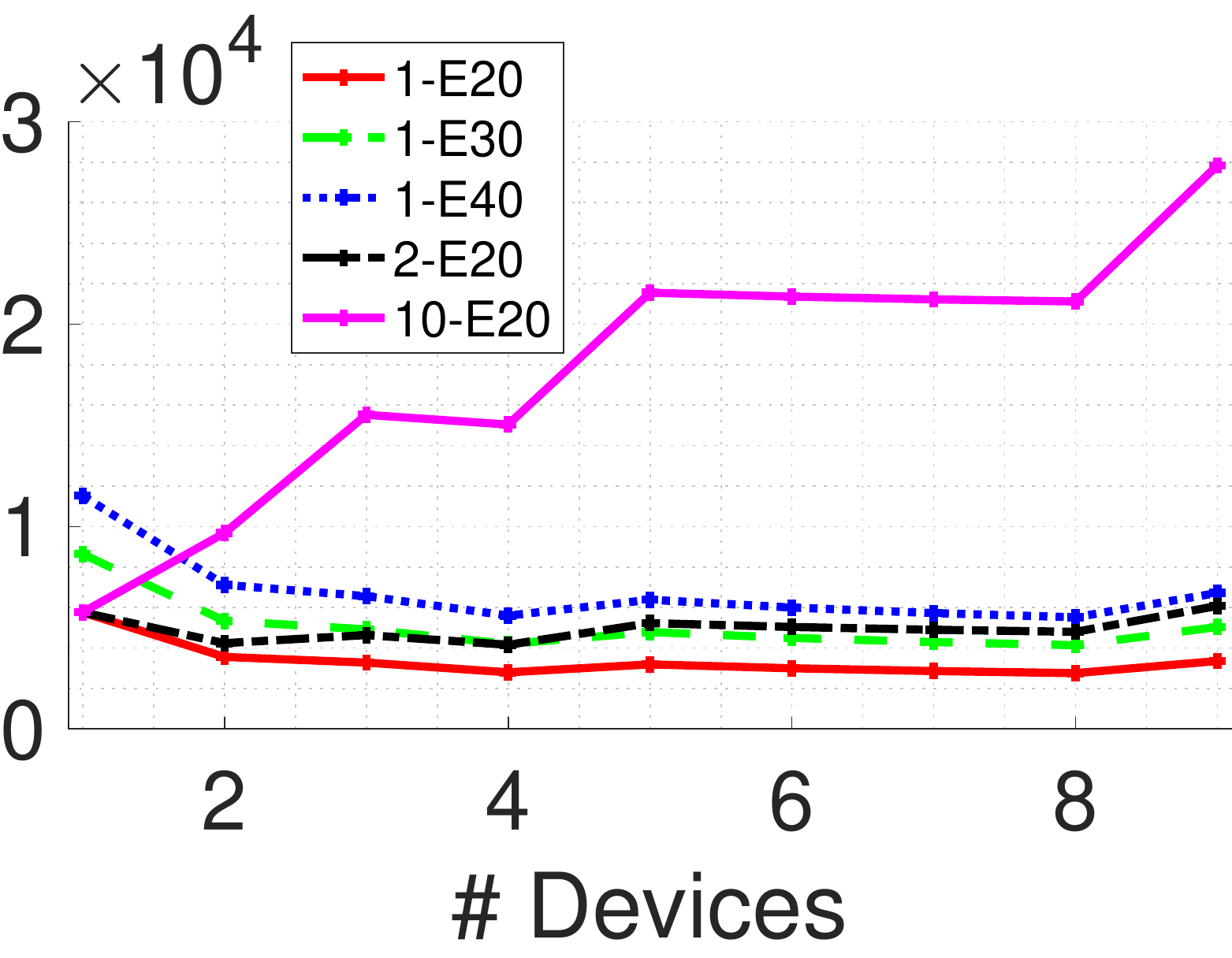}
    \label{fig:tradeoff_time_mhe}
    } 
      \subfloat[OPPORTUNITY]{
    \includegraphics[width=0.16\linewidth]{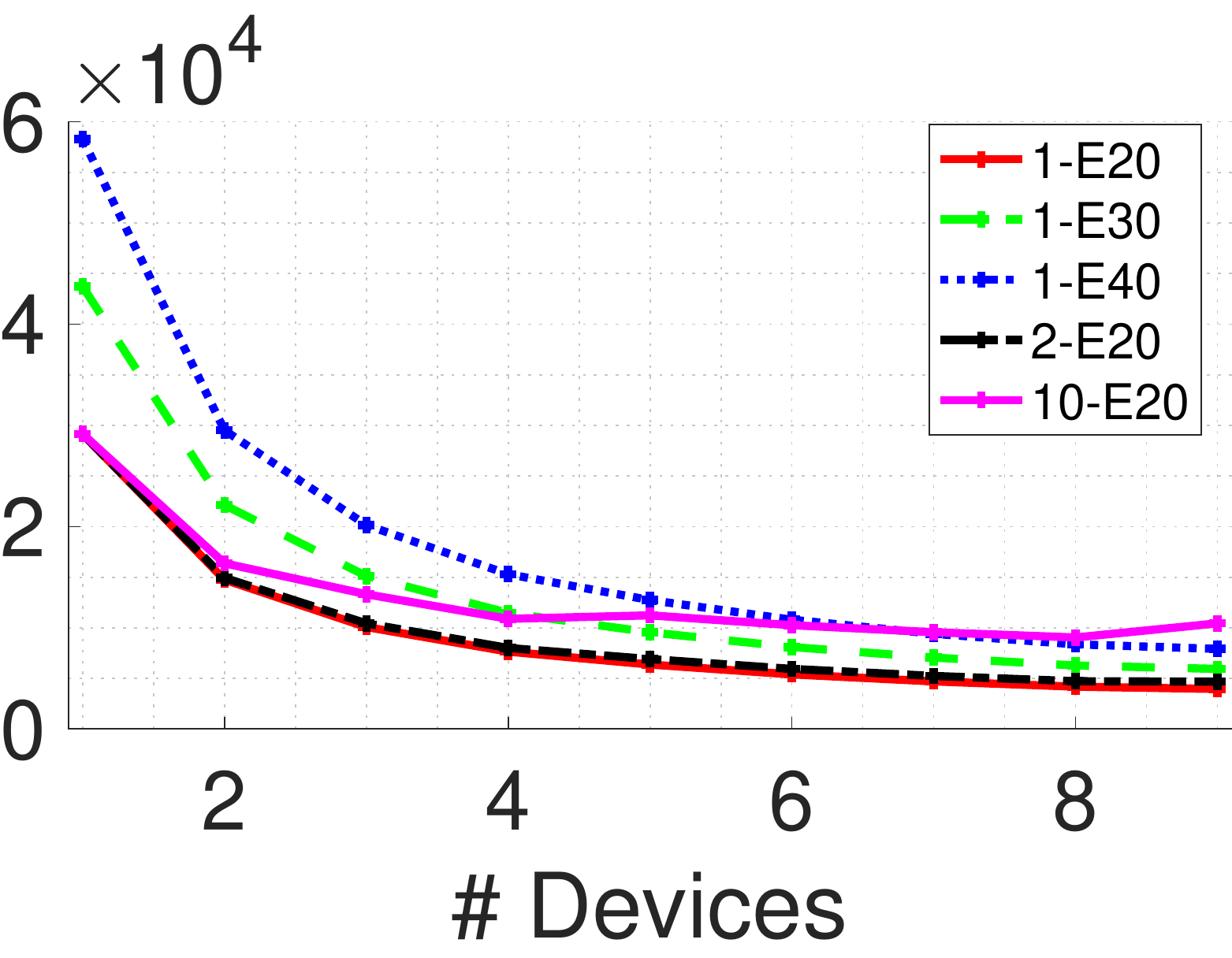}
    \label{fig:tradeoff_time_op}
    } 
     \caption{Training time by trade-off options}
 \label{fig:tradeoff_time}
\end{figure*}

\textbf{Scalability}
We observe that the training accuracy decreases after applying trade-offs, as shown in Figure \ref{fig:tradeoff_ac}. With the network size increases, the training data size for each device decreases, resulting that the learning convergence rate of a DL model becomes slow. However, Figure \ref{fig:tradeoff_ac} indicates the accuracy can be improved if we enlarge the \textit{Epoch} from 1-E20 to 1-E40, while the training time and battery cost increase as shown in Figure \ref{fig:tradeoff_time} and \ref{fig:tradeoff_battery}, respectively.
Therefore, the scalability of network may be limited by the trade-off between resource and accuracy.

\textbf{Battery Saving}
To explore the battery limitations by different trade-offs, we continually charge smartphones during the experiment. Specifically, the battery consumption can be reduced by 2x to 8.5x compared to single-device training, as shown in Figure \ref{fig:tradeoff_battery}. With the number of devices increases, the battery consumption can be effectively shared among multiple devices.

\textbf{Training Time Reduction}
Figure \ref{fig:tradeoff_time} shows that the training time of MDLdroid is reduced by 2x on average compared to single-device training.
Besides, the training time is significantly increased if the model aggregation frequency increases since sending model gradient parameters via BS is time-consuming depending on the model size of dataset.

\textbf{Training Speed Limitation}
Figure \ref{fig:read_speed} reveals that loading one batch data from file to memory has a huge latency on device using DL4J libraries. The same result can be found on the server. As DL4J requires to convert data into an \textit{INDArray}, the speed of this step depends on the performance of hardware (i.e., CPUs). Since DL4J uses only CPUs on Android, 
the latency is much larger than that on the server. Using accelerators available on the mobile device \cite{Chen:2018:ECM:3220192.3220460} may largely improve the training process.

In summary, MDLdroid accelerates training by 2x to 3.5x as shown in Figure \ref{fig:tradeoff_time} and reduces battery consumption by 2x to 8.5x, compared to single device as shown in Figure \ref{fig:tradeoff_battery}.
With the number of training \textit{Epoch} increases, the training accuracy is increased by 2.5\% on average as shown in Figure \ref{fig:tradeoff_ac}, but much more battery and time are consumed. While increasing model aggregation frequency can sightly improve training accuracy with a small impact on battery usage. However, the training time is increased by up to 2x due to heavy communication cost , as shown in Figure \ref{fig:tradeoff_time}. 
Therefore, we choose 1-E20 as the optimized resource-accuracy option to achieve the best performance. Finally, Table \ref{tab:data_accuracy} summarizes the comparison result of training accuracy, and MDLdroid achieves reliable state-of-the-art results. In the table, FL-B denotes the best accuracy on FL, Chain-B denotes the best accuracy on MDLdroid and Chain-Off denotes the accuracy on MDLdroid after trade-off.

\begin{table}
\caption{Training Accuracy Comparison}
\label{tab:data_accuracy}
\resizebox{\columnwidth}{!} {
\begin{tabular}{llllll}
\hline
\textbf{Datasets}        & \textbf{State-of-art} & \textbf{Server} & \textbf{FL-B} & \textbf{Chain-B}  & \textbf{Chain-Off} \\ \hline
\textbf{HAR}             & 96\%  \cite{Anguita:2013} &      93.8\%    &     92.7\%      &  \textbf{92.5\% }     &   \textbf{90.0\%}  \\
\textbf{PAMAP2}          & 90\%+    \cite{Reiss:2012:ISWC}&   97.6\%        &   94.7\%   &  \textbf{95.2\%}   &   \textbf{90.7\%} \\
\textbf{MHEALTH}         & 90\%    \cite{Banos:2014}     &    92.3\%       &    91.0\%       &   \textbf{90.2\%}       &  \textbf{85.4\%}  \\
\textbf{UniMiB}     & 85\%    \cite{Anguita:2013}      &      96.1\%     &      93.3\%     &   \textbf{93.6\%}    &    \textbf{91.5\%}   \\
\textbf{sEMG} & 88\%   \cite{Lobov:2018:EMG}       &     89.2\%      &    86.3\%       &   \textbf{85.8\% }    &   \textbf{84.6\%}  \\
\textbf{OPPO}     & 85\%   \cite{Roggen:2010}         &    88.6\%       &    87.5\%       &  \textbf{86.9\%}   &    \textbf{84.7\%}   \\ \hline
\end{tabular}
}
\end{table}


\section{Discussion and Limitation}

In this section, we discuss several limitations in our current prototype implementation.

\textbf{Training Time} 
The long training time in Figure \mbox{\ref{fig:tradeoff_time}} may not be very practical for end users in reality. However, the training speed of the prototype implementation depends on several factors: 1) the speed of reading training data from a CSV file on mobile device is low as shown in Figure \mbox{\ref{fig:read_speed}}. 2) the training performance on single device is strongly limited by DL4J libraries; 3) the used standard model structures are large for mobile training, and advanced lightweight models should be applied in practice, such as MobileNet \mbox{\cite{Andrew:2017}}. 
Further improving training time will leave for our future works.

\textbf{Bluetooth Limitation} 
As an initial prototype system, we use Bluetooth Low Energy for building a mesh network to simplify our implementation. However, when the model complexity increases, sending a large number of model parameters via BS may suffer long latency due to low bandwidth available in Bluetooth, and hence will affect the overall training performance. Since Wi-Fi Direct \mbox{\cite{Shahin:2014}} has been widely available on smartphones and it offers much higher bandwidth, a hybrid solution can be implemented to use Wi-Fi Direct in the model aggregation process to reduce the communication cost between devices. Furthermore, proper model compressing techniques \mbox{\cite{DBLP:journals/corr/abs-1712-01887}} can be applied for effective communication, which we leave for our future work.



\textbf{Future User Inference and Applications} 
MDLdroid primarily targets personal sensing applications which are privacy-sensitive with low-latency response requirement for continually model inference and update, but the framework can be applied to a wide range of DL-based low-latency applications with a moderate model size, such as real-time surveillance, image recognition, and natural language processing. The prototype user inference of MDLdroid is mainly used for experiments. Since end users may not need to manually set complex parameters for training, we plan to embed the MDLdroid into mobile OS to offer automatic background training, and develop a wider variety of applications to fully explore the capability of MDLdroid in our future work.


\section{RELATED WORK}

\textbf{Decentralized Deep Learning}
For decentralized framework, the existing work \cite{Jiang:2017:CDL:3295222.3295340} 
proposes a theoretical model based on a \textit{fixed directed} graph to offer a decentralized SGD algorithm to exchange model gradient parameters with its \textit{one-hop} neighbors.
However, if the relationship between the device and its \textit{one-hop} neighbor is one-to-many, the device still suffers huge resource overhead which is similar to the master device case. 
On the other hand, 
as the underlying topology is a fixed graph, it cannot properly be performed in a real-time condition with resource dynamicity. By contrast, MDLdroid presents a dynamic \textit{chain-directed} SGD algorithm based on a mesh network with a Chain-scheduler that enables a resource-aware model aggregation process to minimize training latency and reduce training resource overhead.

\textbf{Resource-aware Mobile Deep Learning}
Most of existing works about resource-aware mobile DL mainly focus on inference tasks. NestDNN \cite{Fang:2018:NRM:3241539.3241559} proposes a multi-tenant framework that can enable a resource-aware on-device to efficiently execute inference tasks for mobile vision applications.
Besides, MCDNN \cite{Fang:2018:NRM:3241539.3241559} presents a framework that can execute multiple mobile vision applications based on cloud-based inference solution. In MDLdroid, we fully implement and execute both DL training and inference tasks on devices.

\textbf{Resource-aware Task Scheduling}
The latest works \cite{ben:2017:RL} \cite{Thanh:2018} 
propose to use a MARL based approach to solve task scheduling based on distributed network, which achieves fair performance. However, due to on-device resource limitation, the MARL implementation cannot well perform with training task on device. 
In contrast, MDLdroid applies a single agent-based DQN approach to deal with resource-aware task scheduling.


\section{CONCLUSION}

Towards pushing DL on devices, in this paper, we present MDLdroid, a novel decentralized mobile DL framework to enable resource-aware on-device collaborative learning for personal mobile sensing applications.
MDLdroid achieves a reliable state-of-the-art model training accuracy on multiple off-the-shelf mobile devices. 
The key advantages of MDLdroid include on-device mobile DL, high training accuracy, low resource overhead, low latency for model inference and update, and fair scalability.


%
\begin{acks}
This work is supported by Australian Research Council (ARC) Discovery Project grants DP180103932 and DP190101888.
\end{acks}

%

\balance
\bibliographystyle{ACM-Reference-Format}


\bibliography{reference}


\begin{thebibliography}{48}


\ifx \showCODEN    \undefined \def \showCODEN     #1{\unskip}     \fi
\ifx \showDOI      \undefined \def \showDOI       #1{#1}\fi
\ifx \showISBNx    \undefined \def \showISBNx     #1{\unskip}     \fi
\ifx \showISBNxiii \undefined \def \showISBNxiii  #1{\unskip}     \fi
\ifx \showISSN     \undefined \def \showISSN      #1{\unskip}     \fi
\ifx \showLCCN     \undefined \def \showLCCN      #1{\unskip}     \fi
\ifx \shownote     \undefined \def \shownote      #1{#1}          \fi
\ifx \showarticletitle \undefined \def \showarticletitle #1{#1}   \fi
\ifx \showURL      \undefined \def \showURL       {\relax}        \fi
\providecommand\bibfield[2]{#2}
\providecommand\bibinfo[2]{#2}
\providecommand\natexlab[1]{#1}
\providecommand\showeprint[2][]{arXiv:#2}

\bibitem[\protect\citeauthoryear{A., G., O., P., and L~{Reyes-Ortiz}}{A.
  et~al\mbox{.}}{2013}]%
        {Anguita:2013}
\bibfield{author}{\bibinfo{person}{{Davide} A.}, \bibinfo{person}{{Alessandro}
  G.}, \bibinfo{person}{{Luca} O.}, \bibinfo{person}{{Xavier} P.}, {and}
  \bibinfo{person}{J L~{Reyes-Ortiz}}.} \bibinfo{year}{2013}\natexlab{}.
\newblock \showarticletitle{A Public Domain Dataset for Human Activity
  Recognition using Smartphones}. In \bibinfo{booktitle}{\emph{ESANN'13}}.
\newblock


\bibitem[\protect\citeauthoryear{{Abdallah} and {Kaisers}}{{Abdallah} and
  {Kaisers}}{2016}]%
        {JMLR:v17:14-037}
\bibfield{author}{\bibinfo{person}{S. {Abdallah}} {and} \bibinfo{person}{M.
  {Kaisers}}.} \bibinfo{year}{2016}\natexlab{}.
\newblock \showarticletitle{Addressing Environment Non-Stationarity by
  Repeating Q-learning Updates}.
\newblock \bibinfo{journal}{\emph{Journal of Machine Learning Research}}
  (\bibinfo{year}{2016}).
\newblock


\bibitem[\protect\citeauthoryear{B., G., H., D., P., R., S., and V.}{B.
  et~al\mbox{.}}{2014}]%
        {Banos:2014}
\bibfield{author}{\bibinfo{person}{{Oresti} B.}, \bibinfo{person}{{Rafael} G.},
  \bibinfo{person}{{Juan A.} H.}, \bibinfo{person}{{Miguel} D.},
  \bibinfo{person}{{Hector} P.}, \bibinfo{person}{{Ignacio} R.},
  \bibinfo{person}{{Alejandro} S.}, {and} \bibinfo{person}{{Claudia} V.}}
  \bibinfo{year}{2014}\natexlab{}.
\newblock \showarticletitle{mHealthDroid: A Novel Framework for Agile
  Development of Mobile Health Applications}. In
  \bibinfo{booktitle}{\emph{Ambient Assisted Living and Daily Activities}}.
\newblock


\bibitem[\protect\citeauthoryear{B.~Abkenar, Loke, Zaslavsky, and
  Rahayu}{B.~Abkenar et~al\mbox{.}}{2019}]%
        {Abkenar:19}
\bibfield{author}{\bibinfo{person}{Amin B.~Abkenar}, \bibinfo{person}{Seng
  Loke}, \bibinfo{person}{Arkady Zaslavsky}, {and} \bibinfo{person}{Wenny
  Rahayu}.} \bibinfo{year}{2019}\natexlab{}.
\newblock \showarticletitle{GARSAaaS: group activity recognition and situation
  analysis as a service}.
\newblock \bibinfo{journal}{\emph{JISA'19}} (\bibinfo{year}{2019}).
\newblock


\bibitem[\protect\citeauthoryear{b.~{noureddine}, G., and A.}{b.~{noureddine}
  et~al\mbox{.}}{2017}]%
        {ben:2017:RL}
\bibfield{author}{\bibinfo{person}{D. b. {noureddine}}, \bibinfo{person}{{Atef}
  G.}, {and} \bibinfo{person}{{Samir} A.}} \bibinfo{year}{2017}\natexlab{}.
\newblock \showarticletitle{Multi-agent Deep Reinforcement Learning for Task
  Allocation in Dynamic Environment}. In \bibinfo{booktitle}{\emph{ICSOFT'17}}.
\newblock


\bibitem[\protect\citeauthoryear{Bhagoji, Chakraborty, Mittal, and
  Calo}{Bhagoji et~al\mbox{.}}{2019}]%
        {pmlr-v97-bhagoji19a}
\bibfield{author}{\bibinfo{person}{Arjun~Nitin Bhagoji},
  \bibinfo{person}{Supriyo Chakraborty}, \bibinfo{person}{Prateek Mittal},
  {and} \bibinfo{person}{Seraphin Calo}.} \bibinfo{year}{2019}\natexlab{}.
\newblock \showarticletitle{Analyzing Federated Learning through an Adversarial
  Lens}. In \bibinfo{booktitle}{\emph{ICML'19}}.
\newblock


\bibitem[\protect\citeauthoryear{{Bonawitz}, {Eichner}, {Grieskamp}, {Huba},
  {Ingerman}, {Ivanov}, Kiddon, Konecn{\'{y}}, {Mazzocchi}, {McMahan},
  {Overveldt}, {Petrou}, {Ramage}, and {Roselander}}{{Bonawitz}
  et~al\mbox{.}}{2019}]%
        {DBLP:journals/corr/abs-1902-01046}
\bibfield{author}{\bibinfo{person}{K. {Bonawitz}}, \bibinfo{person}{H.
  {Eichner}}, \bibinfo{person}{W. {Grieskamp}}, \bibinfo{person}{D. {Huba}},
  \bibinfo{person}{A. {Ingerman}}, \bibinfo{person}{V. {Ivanov}},
  \bibinfo{person}{Chlo{\'{e}} Kiddon}, \bibinfo{person}{Jakub Konecn{\'{y}}},
  \bibinfo{person}{S. {Mazzocchi}}, \bibinfo{person}{H.~B. {McMahan}},
  \bibinfo{person}{T.~V. {Overveldt}}, \bibinfo{person}{D. {Petrou}},
  \bibinfo{person}{D. {Ramage}}, {and} \bibinfo{person}{J. {Roselander}}.}
  \bibinfo{year}{2019}\natexlab{}.
\newblock \showarticletitle{Towards Federated Learning at Scale: System
  Design}.
\newblock \bibinfo{journal}{\emph{CoRR}} (\bibinfo{year}{2019}).
\newblock


\bibitem[\protect\citeauthoryear{Chen, Biookaghazadeh, and Zhao}{Chen
  et~al\mbox{.}}{2018}]%
        {Chen:2018:ECM:3220192.3220460}
\bibfield{author}{\bibinfo{person}{Yitao Chen}, \bibinfo{person}{Saman
  Biookaghazadeh}, {and} \bibinfo{person}{Ming Zhao}.}
  \bibinfo{year}{2018}\natexlab{}.
\newblock \showarticletitle{Exploring the Capabilities of Mobile Devices
  Supporting Deep Learning}. In \bibinfo{booktitle}{\emph{HPDC '18}}.
\newblock


\bibitem[\protect\citeauthoryear{D., C., M., C., D., L., M., R., S., T., Y.,
  and N.}{D. et~al\mbox{.}}{2012}]%
        {Dean:2012:LSD:2999134.2999271}
\bibfield{author}{\bibinfo{person}{{Jeffrey} D.}, \bibinfo{person}{{Greg}~S.
  C.}, \bibinfo{person}{{Rajat} M.}, \bibinfo{person}{{Kai} C.},
  \bibinfo{person}{{Matthieu} D.}, \bibinfo{person}{{Quoc}~V. L.},
  \bibinfo{person}{{Mark}~Z. M.}, \bibinfo{person}{{Marc'Aurelio} R.},
  \bibinfo{person}{{Andrew} S.}, \bibinfo{person}{{Paul} T.},
  \bibinfo{person}{{Ke} Y.}, {and} \bibinfo{person}{{Andrew}~Y. N.}}
  \bibinfo{year}{2012}\natexlab{}.
\newblock \showarticletitle{Large Scale Distributed Deep Networks}. In
  \bibinfo{booktitle}{\emph{NIPS'12}}.
\newblock


\bibitem[\protect\citeauthoryear{deeplearning4j}{deeplearning4j}{2019}]%
        {dl4j:nd4j}
\bibfield{author}{\bibinfo{person}{deeplearning4j}.}
  \bibinfo{year}{2019}\natexlab{}.
\newblock \showarticletitle{NDArrays: How Are They Stored in Memory?}
\newblock \bibinfo{journal}{\emph{Deeplearning4j: Open-source, distributed deep
  learning for the JVM}} (\bibinfo{year}{2019}).
\newblock


\bibitem[\protect\citeauthoryear{Deng}{Deng}{2019}]%
        {Yunbin:2019}
\bibfield{author}{\bibinfo{person}{Yunbin Deng}.}
  \bibinfo{year}{2019}\natexlab{}.
\newblock \showarticletitle{Deep Learning on Mobile Devices - {A} Review}.
\newblock \bibinfo{journal}{\emph{CoRR}} (\bibinfo{year}{2019}).
\newblock


\bibitem[\protect\citeauthoryear{F., Z., and Z.}{F. et~al\mbox{.}}{2018}]%
        {Fang:2018:NRM:3241539.3241559}
\bibfield{author}{\bibinfo{person}{{Biyi} F.}, \bibinfo{person}{{Xiao} Z.},
  {and} \bibinfo{person}{{Mi} Z.}} \bibinfo{year}{2018}\natexlab{}.
\newblock \showarticletitle{NestDNN: Resource-Aware Multi-Tenant On-Device Deep
  Learning for Continuous Mobile Vision}. In \bibinfo{booktitle}{\emph{MobiCom
  '18}}.
\newblock


\bibitem[\protect\citeauthoryear{{Gupta}, {Zhang}, and {Wang}}{{Gupta}
  et~al\mbox{.}}{2016}]%
        {Gupta:2016}
\bibfield{author}{\bibinfo{person}{S. {Gupta}}, \bibinfo{person}{W. {Zhang}},
  {and} \bibinfo{person}{F. {Wang}}.} \bibinfo{year}{2016}\natexlab{}.
\newblock \showarticletitle{Model Accuracy and Runtime Tradeoff in Distributed
  Deep Learning: A Systematic Study}. In \bibinfo{booktitle}{\emph{ICDM'16}}.
\newblock


\bibitem[\protect\citeauthoryear{{Hoefler}, {Siebert}, and {Rehm}}{{Hoefler}
  et~al\mbox{.}}{2007}]%
        {Hoefler:2007}
\bibfield{author}{\bibinfo{person}{T. {Hoefler}}, \bibinfo{person}{C.
  {Siebert}}, {and} \bibinfo{person}{W. {Rehm}}.}
  \bibinfo{year}{2007}\natexlab{}.
\newblock \showarticletitle{A practically constant-time MPI Broadcast Algorithm
  for large-scale InfiniBand Clusters with Multicast}. In
  \bibinfo{booktitle}{\emph{IPDPS'07}}.
\newblock


\bibitem[\protect\citeauthoryear{Hoseini-Tabatabaei, Gluhak, and
  Tafazolli}{Hoseini-Tabatabaei et~al\mbox{.}}{2013}]%
        {Hoseini:13}
\bibfield{author}{\bibinfo{person}{Seyed~Amir Hoseini-Tabatabaei},
  \bibinfo{person}{Alexander Gluhak}, {and} \bibinfo{person}{Rahim Tafazolli}.}
  \bibinfo{year}{2013}\natexlab{}.
\newblock \showarticletitle{A Survey on Smartphone-Based Systems for
  Opportunistic User Context Recognition}.
\newblock \bibinfo{journal}{\emph{CSUR'13}} (\bibinfo{year}{2013}).
\newblock


\bibitem[\protect\citeauthoryear{Howard, Zhu, Chen, Kalenichenko, Wang, Weyand,
  Andreetto, and Adam}{Howard et~al\mbox{.}}{2017}]%
        {Andrew:2017}
\bibfield{author}{\bibinfo{person}{Andrew~G. Howard}, \bibinfo{person}{Menglong
  Zhu}, \bibinfo{person}{Bo Chen}, \bibinfo{person}{Dmitry Kalenichenko},
  \bibinfo{person}{Weijun Wang}, \bibinfo{person}{Tobias Weyand},
  \bibinfo{person}{Marco Andreetto}, {and} \bibinfo{person}{Hartwig Adam}.}
  \bibinfo{year}{2017}\natexlab{}.
\newblock \showarticletitle{MobileNets: Efficient Convolutional Neural Networks
  for Mobile Vision Applications}.
\newblock \bibinfo{journal}{\emph{CoRR}} (\bibinfo{year}{2017}).
\newblock


\bibitem[\protect\citeauthoryear{Jeyakumar, Lai, Suda, and
  Srivastava}{Jeyakumar et~al\mbox{.}}{2019}]%
        {Jeyakumar:19}
\bibfield{author}{\bibinfo{person}{Jeya~Vikranth Jeyakumar},
  \bibinfo{person}{Liangzhen Lai}, \bibinfo{person}{Naveen Suda}, {and}
  \bibinfo{person}{Mani Srivastava}.} \bibinfo{year}{2019}\natexlab{}.
\newblock \showarticletitle{SenseHAR: A Robust Virtual Activity Sensor for
  Smartphones and Wearables}. In \bibinfo{booktitle}{\emph{SenSys’19}}.
\newblock


\bibitem[\protect\citeauthoryear{Jiang, Balu, Hegde, and Sarkar}{Jiang
  et~al\mbox{.}}{2017}]%
        {Jiang:2017:CDL:3295222.3295340}
\bibfield{author}{\bibinfo{person}{Zhanhong Jiang}, \bibinfo{person}{Aditya
  Balu}, \bibinfo{person}{Chinmay Hegde}, {and} \bibinfo{person}{Soumik
  Sarkar}.} \bibinfo{year}{2017}\natexlab{}.
\newblock \showarticletitle{Collaborative Deep Learning in Fixed Topology
  Networks}. In \bibinfo{booktitle}{\emph{NIPS'17}}.
\newblock


\bibitem[\protect\citeauthoryear{Konecn{\'{y}}, McMahan, Ramage, and
  Richt{\'{a}}rik}{Konecn{\'{y}} et~al\mbox{.}}{2016}]%
        {Jakub:2016}
\bibfield{author}{\bibinfo{person}{Jakub Konecn{\'{y}}},
  \bibinfo{person}{H.~Brendan McMahan}, \bibinfo{person}{Daniel Ramage}, {and}
  \bibinfo{person}{Peter Richt{\'{a}}rik}.} \bibinfo{year}{2016}\natexlab{}.
\newblock \showarticletitle{Federated Optimization: Distributed Machine
  Learning for On-Device Intelligence}.
\newblock \bibinfo{journal}{\emph{CoRR}} (\bibinfo{year}{2016}).
\newblock


\bibitem[\protect\citeauthoryear{Krizhevsky, Sutskever, and Hinton}{Krizhevsky
  et~al\mbox{.}}{2012}]%
        {Krizhevsky:2012:ICD:2999134.2999257}
\bibfield{author}{\bibinfo{person}{Alex Krizhevsky}, \bibinfo{person}{Ilya
  Sutskever}, {and} \bibinfo{person}{Geoffrey~E. Hinton}.}
  \bibinfo{year}{2012}\natexlab{}.
\newblock \showarticletitle{ImageNet Classification with Deep Convolutional
  Neural Networks}. In \bibinfo{booktitle}{\emph{NIPS'12}}.
\newblock


\bibitem[\protect\citeauthoryear{Kr\"{o}ger, R., and B.}{Kr\"{o}ger
  et~al\mbox{.}}{2019}]%
        {Kroger:2019:PIA:3309074.3309076}
\bibfield{author}{\bibinfo{person}{J.~{Leon} Kr\"{o}ger},
  \bibinfo{person}{{Philip} R.}, {and} \bibinfo{person}{T.~{Rahman} B.}}
  \bibinfo{year}{2019}\natexlab{}.
\newblock \showarticletitle{Privacy Implications of Accelerometer Data: A
  Review of Possible Inferences}. In \bibinfo{booktitle}{\emph{ICCSP '19}}.
\newblock


\bibitem[\protect\citeauthoryear{L., K., K., K., and M.}{L.
  et~al\mbox{.}}{2018}]%
        {Lobov:2018:EMG}
\bibfield{author}{\bibinfo{person}{{Sergey} L.}, \bibinfo{person}{{Nadia} K.},
  \bibinfo{person}{{Innokentiy} K.}, \bibinfo{person}{{Victor} K.}, {and}
  \bibinfo{person}{{Valeri}~A. M.}} \bibinfo{year}{2018}\natexlab{}.
\newblock \showarticletitle{Latent Factors Limiting the Performance of
  sEMG-Interfaces}.
\newblock \bibinfo{journal}{\emph{Sensors}} (\bibinfo{year}{2018}).
\newblock


\bibitem[\protect\citeauthoryear{L., Z., Z., H., Z., and L.}{L.
  et~al\mbox{.}}{2017}]%
        {Lian:2017:DAO:3295222.3295285}
\bibfield{author}{\bibinfo{person}{{Xiangru} L.}, \bibinfo{person}{{Ce} Z.},
  \bibinfo{person}{{Huan} Z.}, \bibinfo{person}{{Cho-Jui} H.},
  \bibinfo{person}{{Wei} Z.}, {and} \bibinfo{person}{{Ji} L.}}
  \bibinfo{year}{2017}\natexlab{}.
\newblock \showarticletitle{Can Decentralized Algorithms Outperform Centralized
  Algorithms? A Case Study for Decentralized Parallel Stochastic Gradient
  Descent}. In \bibinfo{booktitle}{\emph{NIPS'17}}.
\newblock


\bibitem[\protect\citeauthoryear{Lane and Georgiev}{Lane and Georgiev}{2015}]%
        {Lane:2015:DLR:2699343.2699349}
\bibfield{author}{\bibinfo{person}{Nicholas~D. Lane} {and}
  \bibinfo{person}{Petko Georgiev}.} \bibinfo{year}{2015}\natexlab{}.
\newblock \showarticletitle{Can Deep Learning Revolutionize Mobile Sensing?}.
  In \bibinfo{booktitle}{\emph{HotMobile '15}}.
\newblock


\bibitem[\protect\citeauthoryear{Laport-L{\'o}pez, Serrano, Bajo, and
  Campbell}{Laport-L{\'o}pez et~al\mbox{.}}{2019}]%
        {Laport:2019}
\bibfield{author}{\bibinfo{person}{Francisco Laport-L{\'o}pez},
  \bibinfo{person}{Emilio Serrano}, \bibinfo{person}{Javier Bajo}, {and}
  \bibinfo{person}{Andrew~T. Campbell}.} \bibinfo{year}{2019}\natexlab{}.
\newblock \showarticletitle{A review of mobile sensing systems, applications,
  and opportunities}.
\newblock \bibinfo{journal}{\emph{KAIS'19}} (\bibinfo{year}{2019}).
\newblock


\bibitem[\protect\citeauthoryear{{Lecun}, {Bottou}, {Bengio}, and
  {Haffner}}{{Lecun} et~al\mbox{.}}{1998}]%
        {Lecun:1998}
\bibfield{author}{\bibinfo{person}{Y. {Lecun}}, \bibinfo{person}{L. {Bottou}},
  \bibinfo{person}{Y. {Bengio}}, {and} \bibinfo{person}{P. {Haffner}}.}
  \bibinfo{year}{1998}\natexlab{}.
\newblock \showarticletitle{Gradient-based learning applied to document
  recognition}.
\newblock \bibinfo{journal}{\emph{Proc. IEEE}} (\bibinfo{year}{1998}).
\newblock


\bibitem[\protect\citeauthoryear{Li, Zhou, and Chen}{Li et~al\mbox{.}}{2018}]%
        {Li:2018:EIO:3229556.3229562}
\bibfield{author}{\bibinfo{person}{En Li}, \bibinfo{person}{Zhi Zhou}, {and}
  \bibinfo{person}{Xu Chen}.} \bibinfo{year}{2018}\natexlab{}.
\newblock \showarticletitle{Edge Intelligence: On-Demand Deep Learning Model
  Co-Inference with Device-Edge Synergy}. In
  \bibinfo{booktitle}{\emph{MECOMM'18}}.
\newblock


\bibitem[\protect\citeauthoryear{Li, Andersen, Park, Smola, Ahmed, Josifovski,
  Long, Shekita, and Su}{Li et~al\mbox{.}}{2014}]%
        {Li:2014:SDM:2685048.2685095}
\bibfield{author}{\bibinfo{person}{Mu Li}, \bibinfo{person}{David~G. Andersen},
  \bibinfo{person}{Jun~Woo Park}, \bibinfo{person}{Alexander~J. Smola},
  \bibinfo{person}{Amr Ahmed}, \bibinfo{person}{Vanja Josifovski},
  \bibinfo{person}{James Long}, \bibinfo{person}{Eugene~J. Shekita}, {and}
  \bibinfo{person}{Bor-Yiing Su}.} \bibinfo{year}{2014}\natexlab{}.
\newblock \showarticletitle{Scaling Distributed Machine Learning with the
  Parameter Server}. In \bibinfo{booktitle}{\emph{OSDI'14}}.
\newblock


\bibitem[\protect\citeauthoryear{Lin, Han, Mao, Wang, and Dally}{Lin
  et~al\mbox{.}}{2017}]%
        {DBLP:journals/corr/abs-1712-01887}
\bibfield{author}{\bibinfo{person}{Yujun Lin}, \bibinfo{person}{Song Han},
  \bibinfo{person}{Huizi Mao}, \bibinfo{person}{Yu Wang}, {and}
  \bibinfo{person}{William~J. Dally}.} \bibinfo{year}{2017}\natexlab{}.
\newblock \showarticletitle{Deep Gradient Compression: Reducing the
  Communication Bandwidth for Distributed Training}.
\newblock \bibinfo{journal}{\emph{CoRR}} (\bibinfo{year}{2017}).
\newblock


\bibitem[\protect\citeauthoryear{M., M., and N.}{M. et~al\mbox{.}}{2017}]%
        {micucci2017SHAR}
\bibfield{author}{\bibinfo{person}{{Daniela} M.}, \bibinfo{person}{{Marco} M.},
  {and} \bibinfo{person}{{Paolo} N.}} \bibinfo{year}{2017}\natexlab{}.
\newblock \showarticletitle{UniMiB SHAR: a new dataset for human activity
  recognition using acceleration data from smartphones}.
\newblock \bibinfo{journal}{\emph{CoRR}} (\bibinfo{year}{2017}).
\newblock


\bibitem[\protect\citeauthoryear{M., A., M., K., K., C., G., and C.}{M.
  et~al\mbox{.}}{2016}]%
        {Murnane:2016:MMA:2935334.2935383}
\bibfield{author}{\bibinfo{person}{E.~L. M.}, \bibinfo{person}{{Saeed} A.},
  \bibinfo{person}{{Mark} M.}, \bibinfo{person}{{Matthew} K.},
  \bibinfo{person}{{Julie}~A. K.}, \bibinfo{person}{{Tanzeem} C.},
  \bibinfo{person}{{Geri} G.}, {and} \bibinfo{person}{{Dan} C.}}
  \bibinfo{year}{2016}\natexlab{}.
\newblock \showarticletitle{Mobile Manifestations of Alertness: Connecting
  Biological Rhythms with Patterns of Smartphone App Use}. In
  \bibinfo{booktitle}{\emph{MobileHCI '16}}.
\newblock


\bibitem[\protect\citeauthoryear{Miotto, Wang, Wang, and Jiang}{Miotto
  et~al\mbox{.}}{2017}]%
        {Miotto:2017:healthcare}
\bibfield{author}{\bibinfo{person}{Riccardo Miotto}, \bibinfo{person}{Fei
  Wang}, \bibinfo{person}{Shuang Wang}, {and} \bibinfo{person}{Xiaoqian
  Jiang}.} \bibinfo{year}{2017}\natexlab{}.
\newblock \showarticletitle{Deep learning for healthcare: review, opportunities
  and challenges}.
\newblock \bibinfo{journal}{\emph{Briefings in bioinformatics}}
  (\bibinfo{year}{2017}).
\newblock


\bibitem[\protect\citeauthoryear{{Nguyen}, {Nguyen}, and {Nahavandi}}{{Nguyen}
  et~al\mbox{.}}{2018}]%
        {Thanh:2018}
\bibfield{author}{\bibinfo{person}{T.~T. {Nguyen}}, \bibinfo{person}{N.~D.
  {Nguyen}}, {and} \bibinfo{person}{S. {Nahavandi}}.}
  \bibinfo{year}{2018}\natexlab{}.
\newblock \showarticletitle{Deep Reinforcement Learning for Multi-Agent
  Systems: {A} Review of Challenges, Solutions and Applications}.
\newblock \bibinfo{journal}{\emph{CoRR}} (\bibinfo{year}{2018}).
\newblock


\bibitem[\protect\citeauthoryear{{Reiss} and {Stricker}}{{Reiss} and
  {Stricker}}{2012}]%
        {Reiss:2012:ISWC}
\bibfield{author}{\bibinfo{person}{A. {Reiss}} {and} \bibinfo{person}{D.
  {Stricker}}.} \bibinfo{year}{2012}\natexlab{}.
\newblock \showarticletitle{Introducing a New Benchmarked Dataset for Activity
  Monitoring}. In \bibinfo{booktitle}{\emph{ISWC'12}}.
\newblock


\bibitem[\protect\citeauthoryear{{Roggen}, {Calatroni}, {Rossi}, {Holleczek},
  {Förster}, {Tröster}, {Lukowicz}, {Bannach}, {Pirkl}, {Ferscha}, {Doppler},
  {Holzmann}, {Kurz}, {Holl}, {Chavarriaga}, {Sagha}, {Bayati}, {Creatura}, and
  d.~R.~{Millàn}}{{Roggen} et~al\mbox{.}}{2010}]%
        {Roggen:2010}
\bibfield{author}{\bibinfo{person}{D. {Roggen}}, \bibinfo{person}{A.
  {Calatroni}}, \bibinfo{person}{M. {Rossi}}, \bibinfo{person}{T. {Holleczek}},
  \bibinfo{person}{K. {Förster}}, \bibinfo{person}{G. {Tröster}},
  \bibinfo{person}{P. {Lukowicz}}, \bibinfo{person}{D. {Bannach}},
  \bibinfo{person}{G. {Pirkl}}, \bibinfo{person}{A. {Ferscha}},
  \bibinfo{person}{J. {Doppler}}, \bibinfo{person}{C. {Holzmann}},
  \bibinfo{person}{M. {Kurz}}, \bibinfo{person}{G. {Holl}}, \bibinfo{person}{R.
  {Chavarriaga}}, \bibinfo{person}{H. {Sagha}}, \bibinfo{person}{H. {Bayati}},
  \bibinfo{person}{M. {Creatura}}, {and} \bibinfo{person}{J. d. R.~{Millàn}}.}
  \bibinfo{year}{2010}\natexlab{}.
\newblock \showarticletitle{Collecting complex activity datasets in highly rich
  networked sensor environments}. In \bibinfo{booktitle}{\emph{INSS'10}}.
\newblock


\bibitem[\protect\citeauthoryear{{Shahin} and {Younis}}{{Shahin} and
  {Younis}}{2014}]%
        {Shahin:2014}
\bibfield{author}{\bibinfo{person}{A.~A. {Shahin}} {and} \bibinfo{person}{M.
  {Younis}}.} \bibinfo{year}{2014}\natexlab{}.
\newblock \showarticletitle{A framework for P2P networking of smart devices
  using Wi-Fi direct}. In \bibinfo{booktitle}{\emph{PIMRC'14}}.
\newblock


\bibitem[\protect\citeauthoryear{Shi, Wang, Zhao, Tang, Wang, Huang, and
  Chu}{Shi et~al\mbox{.}}{2019}]%
        {Shaohuai:2019}
\bibfield{author}{\bibinfo{person}{Shaohuai Shi}, \bibinfo{person}{Qiang Wang},
  \bibinfo{person}{Kaiyong Zhao}, \bibinfo{person}{Zhenheng Tang},
  \bibinfo{person}{Yuxin Wang}, \bibinfo{person}{Xiang Huang}, {and}
  \bibinfo{person}{Xiaowen Chu}.} \bibinfo{year}{2019}\natexlab{}.
\newblock \showarticletitle{A Distributed Synchronous {SGD} Algorithm with
  Global Top-k Sparsification for Low Bandwidth Networks}.
\newblock \bibinfo{journal}{\emph{CoRR}} (\bibinfo{year}{2019}).
\newblock


\bibitem[\protect\citeauthoryear{Tokic}{Tokic}{2010}]%
        {Tokic:2010}
\bibfield{author}{\bibinfo{person}{Michel Tokic}.}
  \bibinfo{year}{2010}\natexlab{}.
\newblock \showarticletitle{Adaptive $\epsilon$-Greedy Exploration in
  Reinforcement Learning Based on Value Differences}. In
  \bibinfo{booktitle}{\emph{KI 2010: Advances in Artificial Intelligence}}.
\newblock


\bibitem[\protect\citeauthoryear{Wang, Lai, and Roy~Choudhury}{Wang
  et~al\mbox{.}}{2015}]%
        {Wang:2015:MML:2789168.2790121}
\bibfield{author}{\bibinfo{person}{He Wang}, \bibinfo{person}{Ted Tsung-Te
  Lai}, {and} \bibinfo{person}{Romit Roy~Choudhury}.}
  \bibinfo{year}{2015}\natexlab{}.
\newblock \showarticletitle{MoLe: Motion Leaks Through Smartwatch Sensors}. In
  \bibinfo{booktitle}{\emph{MobiCom '15}}.
\newblock


\bibitem[\protect\citeauthoryear{{Wang}, {Cao}, {Yu}, {Sun}, {Bao}, and
  {Zhu}}{{Wang} et~al\mbox{.}}{2018}]%
        {Wang:2018}
\bibfield{author}{\bibinfo{person}{J. {Wang}}, \bibinfo{person}{B. {Cao}},
  \bibinfo{person}{P. {Yu}}, \bibinfo{person}{L. {Sun}}, \bibinfo{person}{W.
  {Bao}}, {and} \bibinfo{person}{X. {Zhu}}.} \bibinfo{year}{2018}\natexlab{}.
\newblock \showarticletitle{Deep Learning towards Mobile Applications}. In
  \bibinfo{booktitle}{\emph{ICDCS'18}}.
\newblock


\bibitem[\protect\citeauthoryear{{Wang}, {Tuor}, {Salonidis}, {Leung},
  {Makaya}, {He}, and {Chan}}{{Wang} et~al\mbox{.}}{2019}]%
        {Wang:19}
\bibfield{author}{\bibinfo{person}{S. {Wang}}, \bibinfo{person}{T. {Tuor}},
  \bibinfo{person}{T. {Salonidis}}, \bibinfo{person}{K.~K. {Leung}},
  \bibinfo{person}{C. {Makaya}}, \bibinfo{person}{T. {He}}, {and}
  \bibinfo{person}{K. {Chan}}.} \bibinfo{year}{2019}\natexlab{}.
\newblock \showarticletitle{Adaptive Federated Learning in Resource Constrained
  Edge Computing Systems}.
\newblock \bibinfo{journal}{\emph{J-SAC'19}} (\bibinfo{year}{2019}).
\newblock


\bibitem[\protect\citeauthoryear{{Yang}, {Xu}, and {Jia}}{{Yang}
  et~al\mbox{.}}{2009}]%
        {Yang:2009}
\bibfield{author}{\bibinfo{person}{J. {Yang}}, \bibinfo{person}{H. {Xu}}, {and}
  \bibinfo{person}{P. {Jia}}.} \bibinfo{year}{2009}\natexlab{}.
\newblock \showarticletitle{Task Scheduling for Heterogeneous Computing Based
  on Bayesian Optimization Algorithm}. In \bibinfo{booktitle}{\emph{CIS'09}}.
\newblock


\bibitem[\protect\citeauthoryear{Yang, Nguyen, San, Li, and Krishnaswamy}{Yang
  et~al\mbox{.}}{2015}]%
        {Yang2015DeepCN}
\bibfield{author}{\bibinfo{person}{Jian~Bo Yang}, \bibinfo{person}{Minh~Nhut
  Nguyen}, \bibinfo{person}{Phyo~Phyo San}, \bibinfo{person}{Xiao~Li Li}, {and}
  \bibinfo{person}{Shonali Krishnaswamy}.} \bibinfo{year}{2015}\natexlab{}.
\newblock \showarticletitle{Deep Convolutional Neural Networks on Multichannel
  Time Series for Human Activity Recognition}. In
  \bibinfo{booktitle}{\emph{IJCAI'15}}.
\newblock


\bibitem[\protect\citeauthoryear{Yang, Liu, Chen, and Tong}{Yang
  et~al\mbox{.}}{2019}]%
        {Yang:19}
\bibfield{author}{\bibinfo{person}{Qiang Yang}, \bibinfo{person}{Yang Liu},
  \bibinfo{person}{Tianjian Chen}, {and} \bibinfo{person}{Yongxin Tong}.}
  \bibinfo{year}{2019}\natexlab{}.
\newblock \showarticletitle{Federated Machine Learning: Concept and
  Applications}.
\newblock \bibinfo{journal}{\emph{TIST'19}} (\bibinfo{year}{2019}).
\newblock


\bibitem[\protect\citeauthoryear{Yu, Flynn, Yoo, and D'Imperio}{Yu
  et~al\mbox{.}}{2019}]%
        {Kwangmin:2019}
\bibfield{author}{\bibinfo{person}{Kwangmin Yu}, \bibinfo{person}{Thomas
  Flynn}, \bibinfo{person}{Shinjae Yoo}, {and} \bibinfo{person}{Nicholas
  D'Imperio}.} \bibinfo{year}{2019}\natexlab{}.
\newblock \showarticletitle{Layered {SGD:} {A} Decentralized and Synchronous
  {SGD} Algorithm for Scalable Deep Neural Network Training}.
\newblock \bibinfo{journal}{\emph{CoRR}} (\bibinfo{year}{2019}).
\newblock


\bibitem[\protect\citeauthoryear{{Zhang}, {Chen}, {Wang}, and {Shi}}{{Zhang}
  et~al\mbox{.}}{2018}]%
        {Zhang:18:collaborative}
\bibfield{author}{\bibinfo{person}{D. {Zhang}}, \bibinfo{person}{X. {Chen}},
  \bibinfo{person}{D. {Wang}}, {and} \bibinfo{person}{J. {Shi}}.}
  \bibinfo{year}{2018}\natexlab{}.
\newblock \showarticletitle{A Survey on Collaborative Deep Learning and
  Privacy-Preserving}. In \bibinfo{booktitle}{\emph{DSC'18}}.
\newblock


\bibitem[\protect\citeauthoryear{Zhang, Yao, Huang, Wang, Tan, Long, and
  Wang}{Zhang et~al\mbox{.}}{2018}]%
        {Xiang:2018}
\bibfield{author}{\bibinfo{person}{Xiang Zhang}, \bibinfo{person}{Lina Yao},
  \bibinfo{person}{Chaoran Huang}, \bibinfo{person}{Sen Wang},
  \bibinfo{person}{Mingkui Tan}, \bibinfo{person}{Guodong Long}, {and}
  \bibinfo{person}{Can Wang}.} \bibinfo{year}{2018}\natexlab{}.
\newblock \showarticletitle{Multi-modality Sensor Data Classification with
  Selective Attention}. In \bibinfo{booktitle}{\emph{IJCAI-18}}.
\newblock


\bibitem[\protect\citeauthoryear{{Zyskind}, {Nathan}, and {Pentland}}{{Zyskind}
  et~al\mbox{.}}{2015}]%
        {Zyskind:15}
\bibfield{author}{\bibinfo{person}{G. {Zyskind}}, \bibinfo{person}{O.
  {Nathan}}, {and} \bibinfo{person}{A.~'. {Pentland}}.}
  \bibinfo{year}{2015}\natexlab{}.
\newblock \showarticletitle{Decentralizing Privacy: Using Blockchain to Protect
  Personal Data}. In \bibinfo{booktitle}{\emph{SPW'15}}.
\newblock


\end{thebibliography}

\end{document}